\documentclass{article}

\PassOptionsToPackage{numbers, compress}{natbib}

\usepackage[preprint]{neurips_2024}

\usepackage[utf8]{inputenc} %
\usepackage[T1]{fontenc}    %
\usepackage{hyperref}       %
\usepackage{url}            %
\usepackage{booktabs}       %
\usepackage{amsfonts}       %
\usepackage{nicefrac}       %
\usepackage{microtype}      %
\usepackage[table,xcdraw,dvipsnames]{xcolor}
\usepackage{graphicx}	
\usepackage{amsmath}
\usepackage{amssymb}	
\usepackage{booktabs}
\usepackage{times}
\usepackage{epsfig}
\usepackage{caption}
\usepackage{float}
\usepackage{placeins}
\usepackage{color, colortbl}
\usepackage{stfloats}
\usepackage{enumitem}
\usepackage{tabularx}
\usepackage{xstring}
\usepackage{multirow}
\usepackage{xspace}
\usepackage{subcaption}
\usepackage[hang,flushmargin]{footmisc}

\title{Evaluating Alternatives to SFM Point Cloud Initialization for Gaussian Splatting}

\newcommand*{\affaddr}[1]{#1}
\newcommand*{\affmark}[1][*]{\textsuperscript{#1}}

\author{
    Yalda Foroutan\affmark[1] \qquad
    Daniel Rebain\affmark[2] \qquad
    Kwang Moo Yi\affmark[2] \qquad
    Andrea Tagliasacchi\affmark[1]$^,$\affmark[3]$^,$\affmark[4]\\
    \affaddr{\affmark[1]Simon Fraser University},
    \affaddr{\affmark[2]University of British Columbia},
    \affaddr{\affmark[3]University of Toronto},\\
    \affaddr{\affmark[4]Google DeepMind}%
}

\usepackage{xcolor}
\usepackage{amsmath}
\usepackage{amssymb}
\usepackage{graphicx}
\usepackage{xspace}
\usepackage{multirow}
\usepackage{comment}
\usepackage{soul}
\usepackage{enumitem}

\DeclareMathOperator*{\argmin}{arg\,min}

\definecolor{turquoise}{cmyk}{0.65,0,0.1,0.3}
\definecolor{purple}{rgb}{0.65,0,0.65}
\definecolor{dark_green}{rgb}{0, 0.5, 0}
\definecolor{orange}{rgb}{0.9, 0.6, 0.1}
\definecolor{red}{rgb}{0.8, 0.2, 0.2}
\definecolor{darkred}{rgb}{0.6, 0.1, 0.05}
\definecolor{blueish}{rgb}{0.0, 0.3, .6}
\definecolor{light_gray}{rgb}{0.7, 0.7, .7}
\definecolor{pink}{rgb}{1, 0, 1}
\definecolor{greyblue}{rgb}{0.25, 0.25, 1}

\definecolor{color1}{rgb}{0.9, 0.65, 0.65}
\definecolor{color2}{rgb}{0.95, 0.8, 0.8}
\definecolor{color3}{rgb}{1.0, 0.9, 0.9}

\renewcommand{\paragraph}[1]{\vspace{.5em}\noindent\textbf{#1.}}

\newcommand{\losst}[1]{\mathcal{L}_\text{#1}}

\newcommand{\expect}{\mathbb{E}}

\newcommand{\ray}{\mathbf{r}}
\newcommand{\gaussian}{\mathcal{G}}
\newcommand{\opacity}{\alpha}
\newcommand{\covariance}{\Sigma}
\newcommand{\mean}{\mu}
\newcommand{\rgbsh}{h}
\newcommand{\surfaceness}{S}
\newcommand{\cdf}{W}
\newcommand{\depth}{D}

\newcolumntype{L}{>{\hspace*{-\tabcolsep}\kern\tabcolsep}l}
\newcolumntype{R}{c<{\hspace*{-\tabcolsep}\kern\tabcolsep}}

\begin{document}

\maketitle

\begin{abstract}
3D Gaussian Splatting has recently been embraced as a versatile and effective method for scene reconstruction and novel view synthesis, owing to its high-quality results and compatibility with hardware rasterization.
Despite its advantages, Gaussian Splatting's reliance on high-quality point cloud initialization by Structure-from-Motion (SFM) algorithms is 
a significant limitation to be overcome.
To this end, we investigate various initialization strategies for Gaussian Splatting and delve into how volumetric reconstructions from Neural Radiance Fields (NeRF) can be utilized to bypass the dependency on SFM data.
Our findings demonstrate that random initialization can perform much better if carefully designed and that by employing a combination of improved initialization strategies and structure distillation from low-cost NeRF models, it is possible to achieve equivalent results, or at times even superior, to those obtained from SFM initialization.
Source code is available at \href{https://theialab.github.io/nerf-3dgs}{https://theialab.github.io/nerf-3dgs}
\end{abstract}

\section{Introduction}
\label{sec:intro}

In the years since their introduction, Neural Radiance Fields (NeRF)~\cite{mildenhall2021nerf} have emerged as a leading technique for 3D reconstruction from images and novel view synthesis, capturing the interest of researchers and practitioners alike. 
Despite its success, the application of NeRF in practical scenarios has been limited by its high computational demands. 
Due in part to this limitation, 3D Gaussian Splatting~\cite{kerbl3Dgaussians} has gained traction as a more efficient alternative, as it achieves real-time inference speeds by leveraging rasterization.

However, the adoption of Gaussian Splatting is not without challenges. 
A significant hurdle is its dependence on careful initialization from point clouds, such as those from Structure-from-Motion~(SfM).
The reliance on slow and computationally expensive SFM implementations such as COLMAP~\cite{schoenberger2016sfm, schoenberger2016mvs}, and comparatively poor performance of random initialization is a critical limitation that impedes the application of Gaussian Splatting in domains where a full SFM solution would be too expensive.
This includes situations like SLAM sequences where 
camera poses are available alongside point clouds too sparse to be useful for initialization, such as ORB-SLAM~\cite{mur2015orb}, or autonomous vehicle applications where navigation and localization are performed with other sensors, \eg, aircraft.

Such applications could benefit significantly from a computationally efficient alternative to SFM for Gaussian Splatting initialization.
With this motivation, our purpose in this work is to perform a deeper analysis of the initialization phase of Gaussian Splatting, and identify how existing methods could be leveraged to improve the situation.

\begin{figure*}
    \centering
\includegraphics[width=\textwidth]{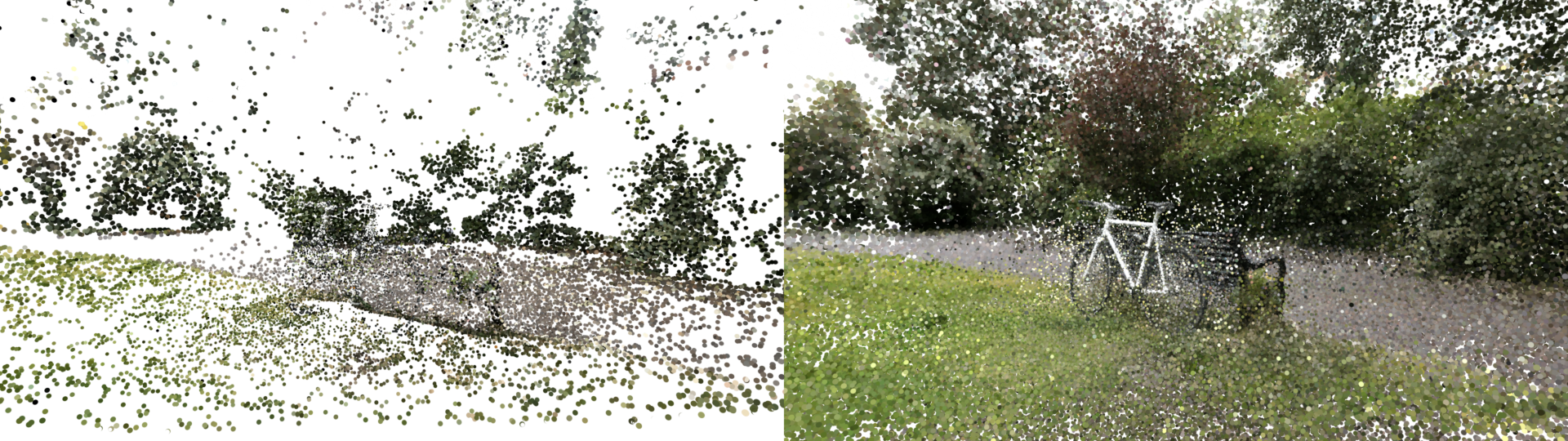}
\caption{
COLMAP~\cite{schoenberger2016sfm, schoenberger2016mvs} (Left) vs. NeRF-based (Right) initialization point clouds for Gaussian Splatting.
The NeRF-based initialization provides a much more complete model of the scene structure, while also being faster to construct with posed images.
}
\label{fig:teaser}
\end{figure*}

We hypothesize that the limitations of Gaussian Splatting stem from its discrete and localized pruning and growing operations, which struggle to capture the coarse structure of scenes effectively.
In particular, regions of missing geometry in the SFM solution, such as those shown in Figure~\ref{fig:teaser}, which can be caused by lack of texture, dynamic scene content, or strong view-dependent effects, can leave the optimization near pathological minima which the local optimization fails to escape.
In contrast, NeRF optimization handles such non-idealities more gracefully, tending to produce over-smoothed reconstructions rather than missing geometry.

Given the proficiency of NeRF and similar neural field-based volume-rendering methods in reconstructing scene geometry at a coarse level, our analysis centers on integrating these techniques into the Gaussian Splatting training pipeline.
Specifically, we investigate both point cloud initialization and depth supervision derived from a NeRF model trained for a short amount of time on the same data.

\paragraph{Contributions}
Through extensive experiments, we find that a combination of better random initialization and structure guidance from NeRF are sufficient to match or exceed the quality reached from a COLMAP initialization, even on very large-scale, challenging scenes.
We apply this scheme to scenes with cameras estimated using a state-of-the-art SLAM method~\cite{ORBSLAM3_TRO}, and show that the results are competitive with those derived from COLMAP initialization while taking far less time overall to go from images to the final model.

\begin{figure*}[h!]
    \centering
\includegraphics[width=\textwidth]{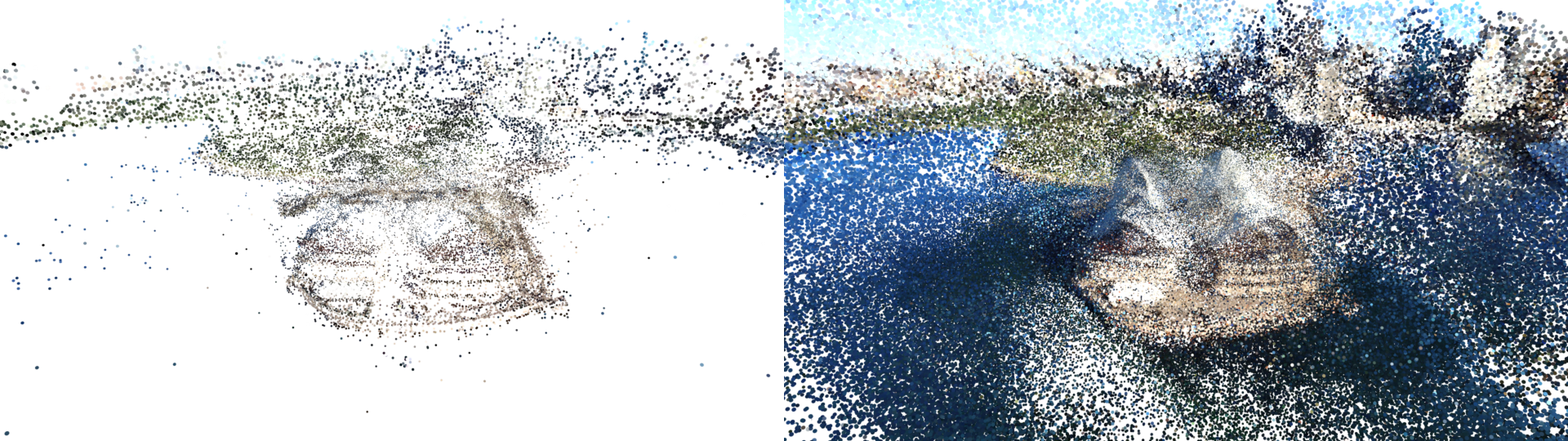}
\includegraphics[width=\textwidth]{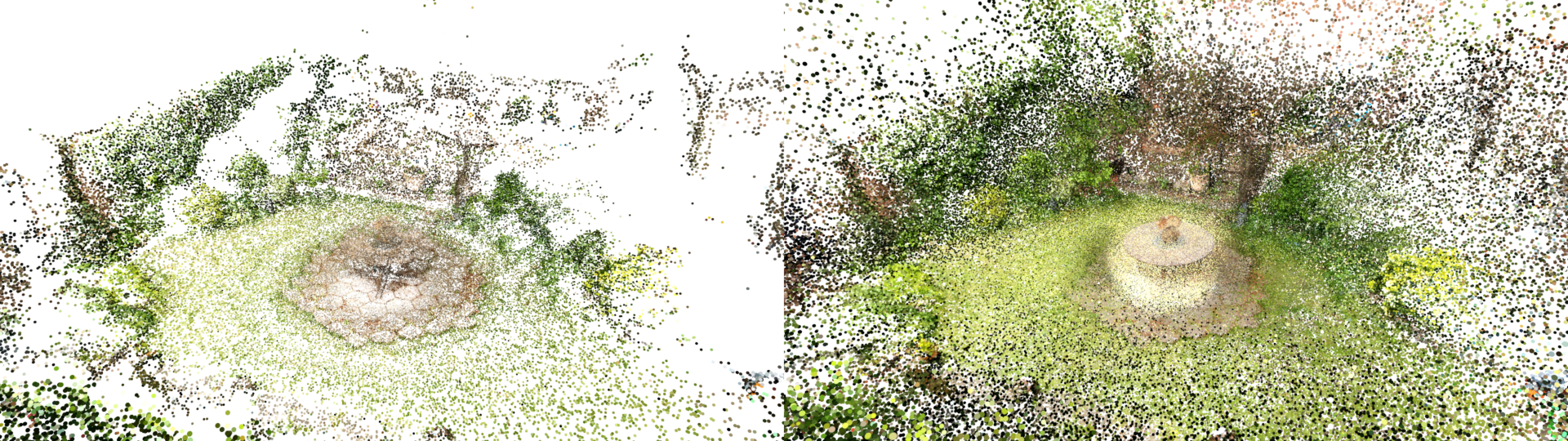}
\caption{COLMAP (Left) vs. NeRF-based (Right) initialization point clouds for Gaussian Splatting.
  In this scene from OMMO~\cite{lu2023large} (Top), the surface of the water is dominated by dynamic content (waves), and view-dependent effects (reflections), which causes the SFM point cloud to be nearly empty in these areas.
  In contrast, the NeRF initialization is able to place points near this surface despite the lack of view-consistent detail.
  Even for more static and Lambertian scenes, such as the garden from Mip-NeRF 360~\cite{barron2022mip} (Bottom), the NeRF point cloud is still significantly more complete.}
\label{fig:sidney_points}
\end{figure*}

\section{Related Work}
\label{sec:related}

\paragraph{Differentiable rendering architectures}
Since the introduction of neural fields~\cite{chen2019learning, mescheder2019occupancy, park2019deepsdf}, a variety of neural rendering methods have been proposed which use gradient descent optimization of image reconstruction objectives to construct scene representations for novel view synthesis.
Perhaps most notable is Neural Radiance Fields (NeRF)~\cite{mildenhall2021nerf}, which has become the basis of an ever-expanding family of volumetric scene models.
While NeRF originally used a purely neural model, further works explored methods based partially~\cite{liu2020neural} or entirely~\cite{fridovich2022plenoxels} on classical grids as a way of exploring the trade-off between compactness and efficiency.

\paragraph{Accelerated volumetric rendering}
As it became clear that much higher efficiency could be achieved through application of grid structures, works like Instant Neural Graphics Primitives (INGP)~\cite{mueller2022instant} and Tensorial Radiance Fields (TensoRF)~\cite{chen2022tensorf} appeared which leveraged not only grid structures but additional strategies of hashing and tensor factorization to compress the grid representation itself, leading to significant gains in both training and inference speed.

\paragraph{Gaussian Splatting}
While significant advances have been made in improving the efficiency of models based on volume rendering, it is still a challenge to reach truly interactive inference speeds with them, due largely to the Monte Carlo integration required for each ray.
An alternative approach to differentiable rendering that avoids this is \textit{rasterization}, in which individual primitives are projected into image space and written to a buffer in order of depth.
One such method is Pulsar~\cite{lassner2021pulsar}, which uses partially transparent spheres as a representation and demonstrates the expressive power of such primitive-based models.
More recently, Gaussians were proposed as a primitive~\cite{kerbl3Dgaussians}, and were shown to achieve both very high-quality image reconstruction and better than real-time inference speeds.
This result sparked a significant number of follow-up works, ranging from modelling dynamic content~\cite{yuan2023gavatar, yang2023deformable, wu20234d, kocabas2023hugs}, to generative modelling~\cite{zou2023triplane, charatan2023pixelsplat, tang2023dreamgaussian}.
There were also improvements made to the handling of primitive scale in rendering to avoid aliasing~\cite{yan2023multi, yu2023mip}.
A concurrent work further views the training process of Gaussian Splatting as a sampling process~\cite{kheradmand20243d}, and attempts to remove initialization dependence, similarly to ours.

\paragraph{Gaussian Splatting SLAM}
There have also been a number of methods proposed for using Gaussian Splatting as an internal representation for Simultaneous Localization And Mapping (SLAM)~\cite{yan2023gs, yugay2023gaussianslam}, which enables solving camera estimation and the construction of a representation for novel view synthesis at the same time given a stream of RGB and depth data.
One such recent work~\cite{monogs} also supports performing SLAM on RGB-only video -- we compare the results of this method with the strategies we present for COLMAP-free training.
Another similar work~\cite{fu2024colmapfree} supports RGB-only training, but has not yet made code available for comparison.

\paragraph{Depth guidance and regularization}
Other differentiable rendering works have also recognized the potential of depth regularization and supervision in improving novel view synthesis results.
Works such as RegNeRF~\cite{Niemeyer2021Regnerf}, LOLNeRF~\cite{rebain2022lolnerf}, and Mip-NeRF 360~\cite{barron2022mip} all proposed some form of regularization of the density sampled along a ray in order to improve the recovered 3D structure.
Other works~\cite{neff2021donerf, roessle2022dense, deng2022depth} have explored directly supervising depth to improve the final rendered results.
One concurrent work to ours~\cite{niemeyer2024radsplat} also explores initialization and supervision of Gaussian Splatting from NeRF models, with similar findings regarding the benefits of the improved initialization.
Another concurrent work~\cite{fan2024instantsplat} also explores replacing COLMAP initialization, specifically with pre-trained camera pose and scene structure estimation networks.

\section{Method}
\label{sec:method}

Gaussian Splatting~\cite{kerbl3Dgaussians} represents scenes as collections of 3D Gaussian primitives, each associated with a mean $\mean$, covariance matrix $\covariance$, opacity $\opacity$ and RGB spherical harmonics coefficients $\rgbsh$.
These primitives can be used to render the scene by sequentially projecting the Gaussians in  depth order into image space and rasterizing them with alpha blending to obtain a final color for each ray $\ray$:
\begin{equation}
    C_\textrm{GS}(\ray) = \sum_{i=1}^N c(\ray; \rgbsh_i) \: \opacity_i \: \gaussian(\ray; \covariance_i, \mean_i) \prod_{j=1}^{i-1} (1 - \opacity_j \: \gaussian(\ray; \covariance_j, \mean_j)),
\end{equation}
where $c(\ray; \rgbsh)$ denotes querying the spherical color function in the direction of $\ray$, and $\gaussian(\ray; \covariance, \mean)$ is the projection of the 3D Gaussian PDF into image space.
Such a model is differentiable with respect to all parameters, and can therefore be trained by image reconstruction objectives.
This, in theory, should allow continuous optimization starting from a random initialization, similar to NeRF.

However, unlike NeRF, 
because Gaussians are \emph{bounded locally} in practice despite their theoretically infinite support,
Gaussian Splatting models are less likely to converge from a random initialization far from the final distribution of primitives.
For this reason, the original Gaussian Splatting method employed discrete splitting, cloning, and pruning heuristics, which interrupt the continuous optimization to take larger steps toward an optimal configuration, as defined by tuned thresholds.
This approach is at least partially effective in closing the gap between Gaussian Splatting and NeRF in terms of scene structure recovery, but for large-scale, complex scenes, the original publication~\cite{kerbl3Dgaussians} still reported significant gains when initializing from an SFM point cloud versus a uniform random distribution.

In the remainder of this section, we will describe the potential strategies for closing this gap which we 
empirically explore.

\subsection{Strategies for random initialization}
The first and most obvious approach to avoiding the use of SFM initialization is to simply search for better random initializations which do not require knowledge of the scene structure.
The experiments with random initialization performed in the original paper are described as using a uniform distribution of Gaussian centers within a scaled version of the camera bounding box.
In our experiments, we attempt to reproduce the results of this bounding box strategy, as well as using an even simpler strategy of a very large uniform initialization sufficient to cover the entire scene.
We report the results of these experiments in Section~\ref{sec:rand_init_experiments}, as well as comparison to the results reported in the original paper.

\subsection{Initialization from a volumetric model}
Beyond seeking an ``improved'' random initialization, we also look for ways that the superior structure recovery of volumetric NeRF models can be leveraged to bootstrap the training of Gaussian Splatting.
We approach this by using a probabilistic interpretation of NeRF density in volume rendering which sees volume rendering weights as a distribution of ``surface'' locations, i.e. where a ray is likely to terminate~\cite{goli2023nerf2nerf}.
This differential\footnote{To define a continuous PDF, we use an unbounded differential value rather than integrating over a local interval as in~\cite{goli2023nerf2nerf}.} ``surfaceness'' value at a depth $t$ along some ray $\ray$ is defined as:
\begin{equation}
    \surfaceness(t; \ray) = \sigma(\ray(t)) \cdot \textrm{exp}\left( - \int_{t_\textrm{min}}^{t} \sigma(\ray(s)) ds \right),
\end{equation}
where $\sigma(\ray(t))$ denotes the NeRF density at depth $t$ along $\ray$.
This termination probability can then be used to derive a cumulative distribution function along each ray:
\begin{equation}
    \cdf(t; \ray) = \int_{t_\textrm{min}}^{t} \surfaceness(t; \ray) ds.
\end{equation}
Under this interpretation, the importance samples used in NeRF volume rendering can be repurposed as samples drawn from regions where the NeRF model~``expects'' the surface to be.
We draw a set of such samples from a trained NeRF model, and use the spatial positions of these samples, along with the predicted NeRF radiance, to initialize a Gaussian Splatting model:
\begin{align}
    \mean = \ray(\cdf^{-1}(u; \ray)), \quad 
    c = c_\textrm{NeRF}(\mean; \ray), \quad
    u \sim \mathcal{U}[0, 1],
\end{align}
where $\cdf^{-1}(u; \ray)$ is the inverse CDF which maps the random parameter $u$ to a depth along the ray $\ray$, and $c_\textrm{NeRF}(\mean; \ray)$ is the color value queried from the radiance field at the sample location $\mean$ in the direction of $\ray$.
Given the large number of rays in each scene, we uniformly sample a subset of all training rays and draw a single sample from each ray.

\paragraph{Implementation}
To keep our analysis as general as possible, we choose an efficient, off-the-shelf NeRF implementation, specifically NerfAcc~\cite{li2023NerfAcc}, to base our experiments on.
We use the proposal network-based INGP~\cite{mueller2022instant} model in NerfAcc, which supports large scale 360 degree scenes by implementing the scene contraction and proposal components of Mip-NeRF 360~\cite{barron2022mip}.
We found this model to provide an excellent balance of flexibility and efficiency.

\subsection{Volumetric structure distillation}
\label{sec:depth_supervision}

In addition to simply initializing from the NeRF, we also investigated using depth prediction from the NeRF to guide training of the Gaussian Splatting model with an additional loss term.
To do this, the rendering process for Gaussian Splatting must be modified to also predict depth values in a differentiable way.
We use a similar formulation to the alpha compositing model for color rendering:
\begin{equation}
    \depth_\textrm{GS}(\ray) = \sum_{i=1}^N d_i \: \opacity_i \: \gaussian(\ray; \covariance_i, \mean_i) \prod_j^{i-1} (1 - \opacity_j \: \gaussian(\ray; \covariance_j, \mean_j)),
\end{equation}
where $d_i$ is the mean of the 1D Gaussian conditional distribution formed by the intersection of $\ray$ with the 3D Gaussian (see the Supplementary Material for derivation).
Depth estimates can similarly be extracted from the NeRF model by accumulating sample depth:
\begin{equation}
    \depth_\textrm{NeRF}(\ray) = \int_{t_\textrm{min}}^{t_\textrm{max}} t \sigma(\ray(t)) \textrm{exp}\left( - \int_{t_\textrm{min}}^{t} \sigma(\ray(s)) ds \right) dt.
\end{equation}
We precompute depth estimates from the NeRF model for all training rays, and incorporate them into the Gaussian Splatting term using the following loss:
\[
\losst{GS+Depth} = \losst{GS} + \lambda \losst{Depth},
\]
\vspace{-2em}
\begin{align}
\losst{Depth} = \expect_{\ray} \:\: \left[|\depth_\textrm{NeRF}(\ray) - \depth_\textrm{GS}(\ray)|\right],
\label{eq:depth_loss}
\end{align}
where $\losst{GS}$ is the original Gaussian Splatting loss function, and $\lambda$ is a weight which we schedule over training to control the strength of the structure guidance; see the Supplementary Material.

\subsection{Hyper-parameters}

The original Gaussian Splatting implementation depends on a number of parameters which are tuned based on the original photometric loss formulation.
As such, it is necessary to take care in setting parameters when changing the loss formulation.
We found through experimentation that including a depth term in the loss function alters the training dynamics of pruning, splitting and cloning, which can lead to an excessive number of Gaussians being instantiated if the wrong parameters are used.
For all real data scenes, we use an exponential decay schedule for the depth loss weight $\lambda$ with an initial value of 0.9, and a decay rate of 0.9 every 100 iterations.

\section{Experiments}
\label{sec:experiments}
\begin{figure*}[h!]
  \centering
  \begin{subfigure}{.495\textwidth}
      \centering
      \includegraphics[width=\textwidth]{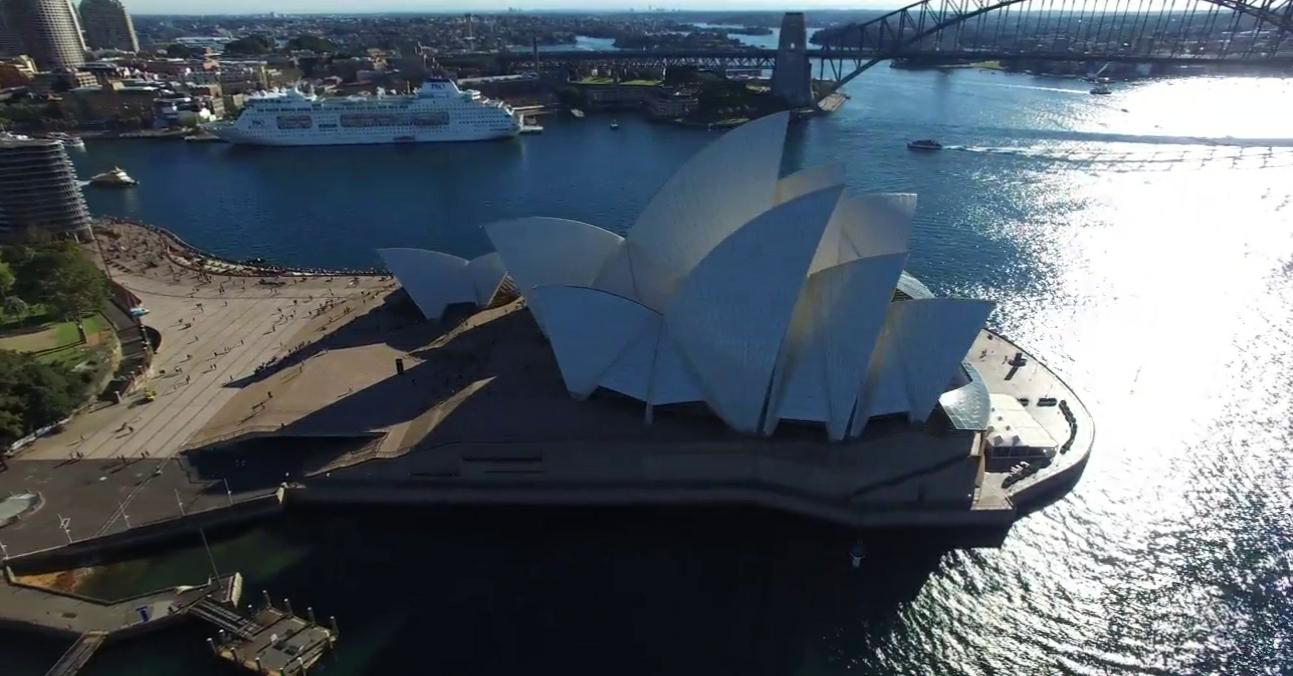}
      \caption*{Ground truth}
  \end{subfigure}
  \begin{subfigure}{.495\textwidth}
      \centering
      \includegraphics[width=\textwidth]{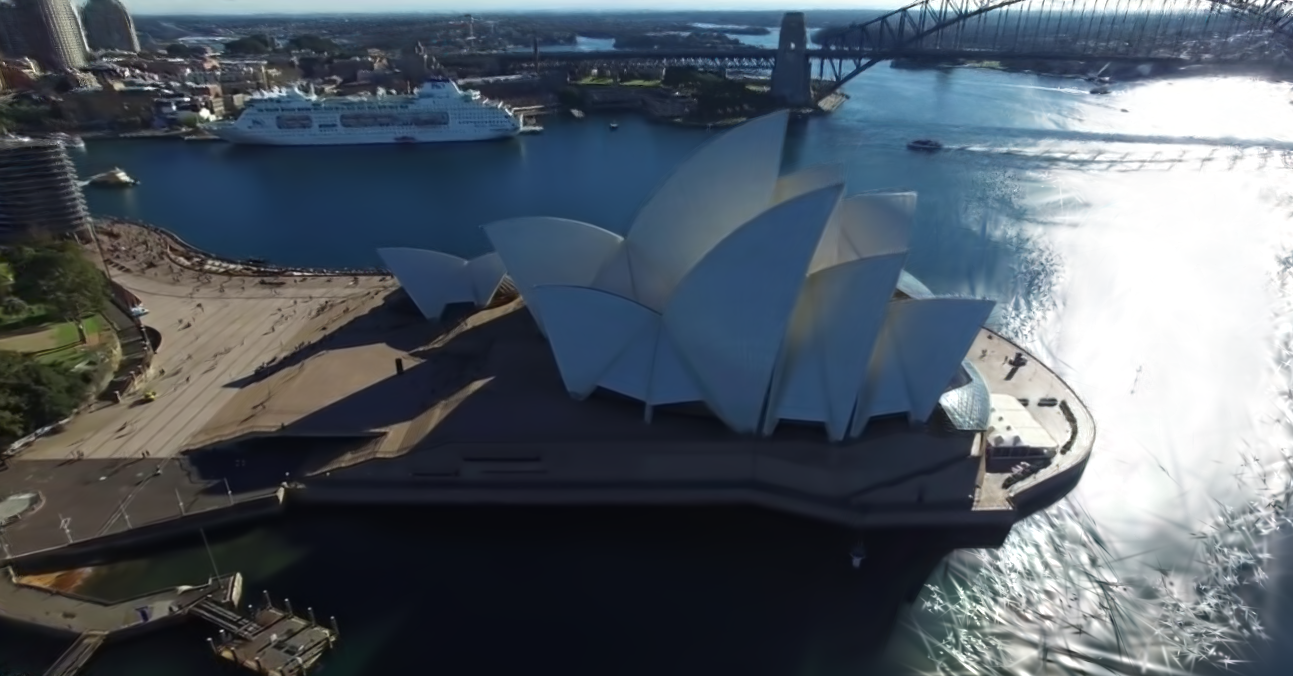}
      \caption*{COLMAP}
  \end{subfigure}
  \begin{subfigure}{.495\textwidth}
      \centering
      \includegraphics[width=\textwidth]{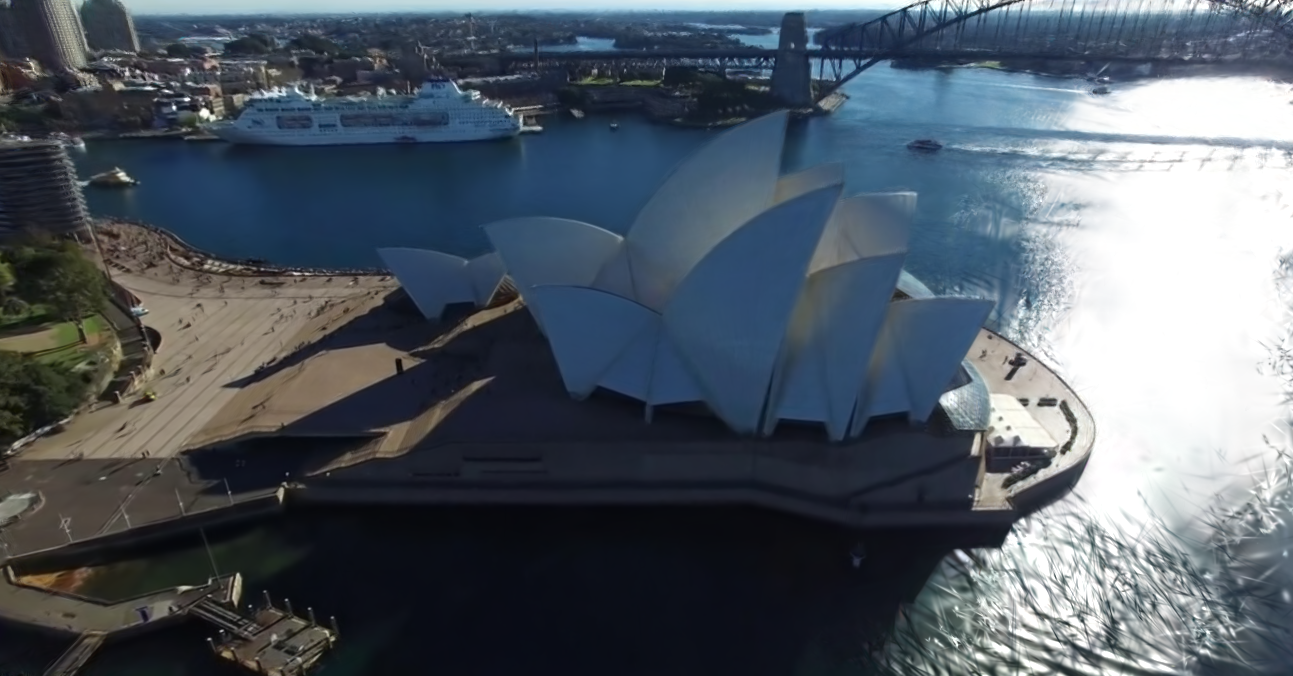}
      \caption*{NerfAcc}
  \end{subfigure}
    \begin{subfigure}{.495\textwidth}
      \centering
      \includegraphics[width=\textwidth]{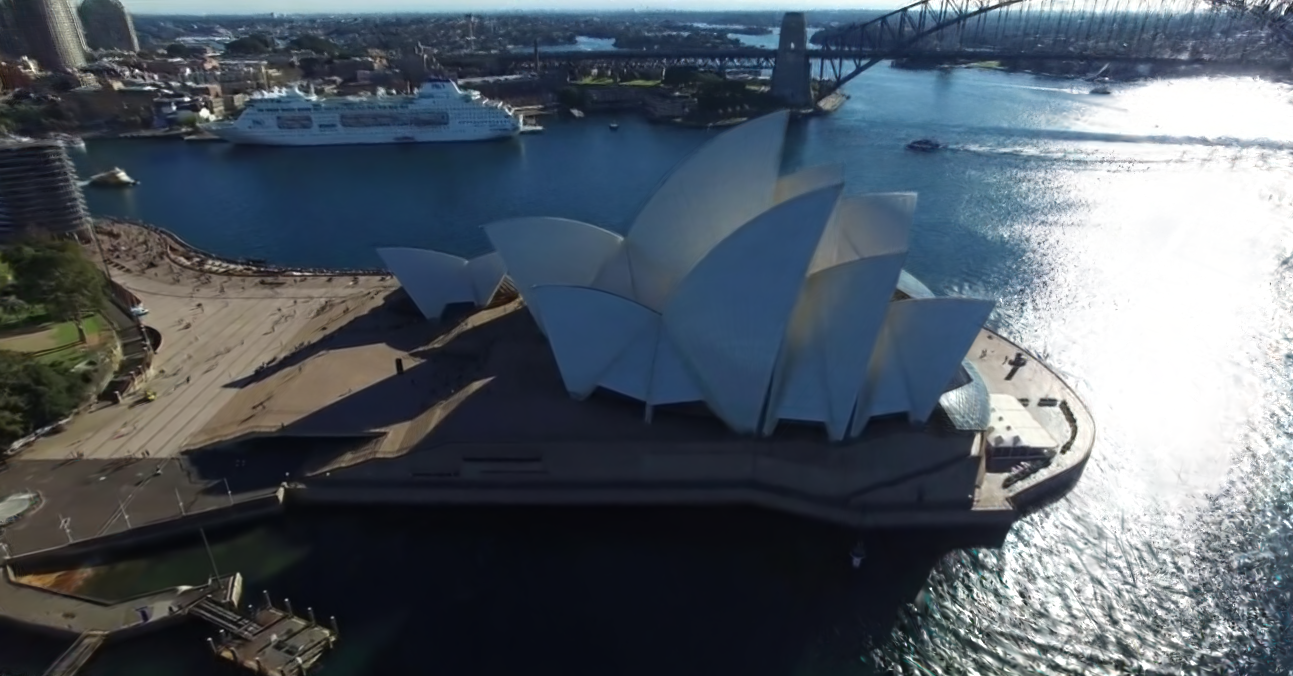}
      \caption*{NerfAcc + depth}
  \end{subfigure}
  \begin{subfigure}{.495\textwidth}
      \centering
      \includegraphics[width=\textwidth]{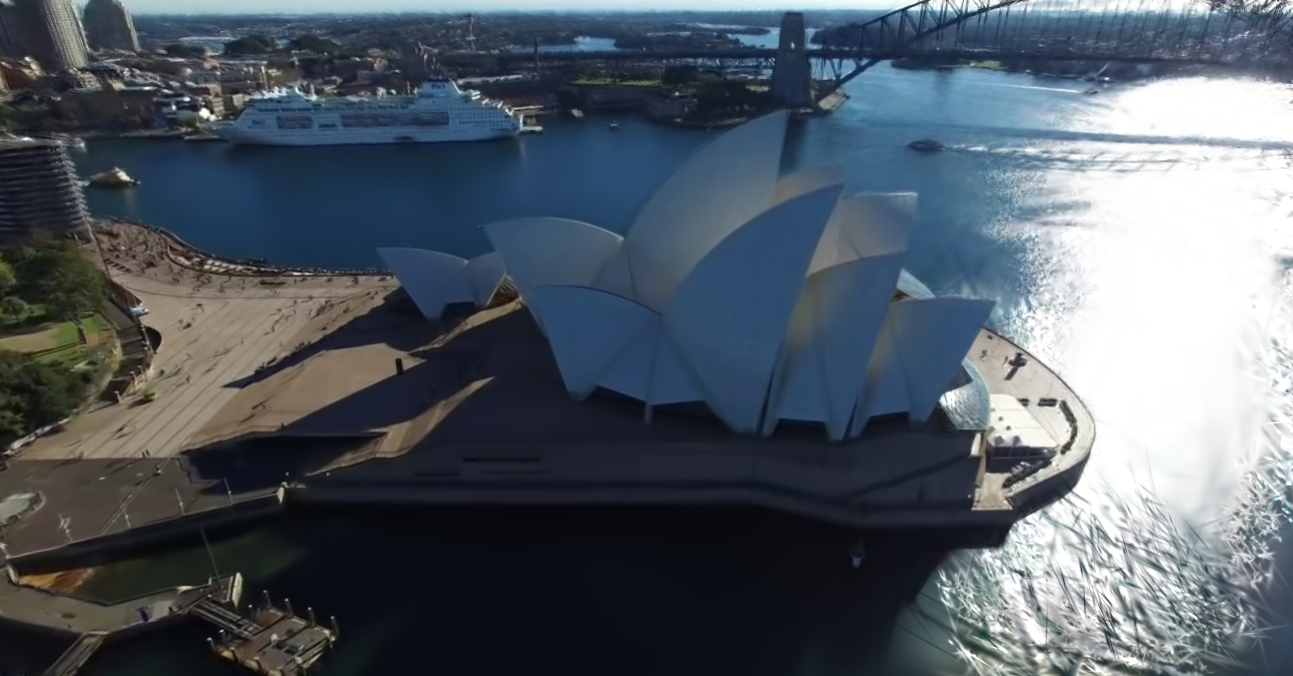}
      \caption*{Random}
  \end{subfigure}
    \begin{subfigure}{.495\textwidth}
      \centering
      \includegraphics[width=\textwidth]{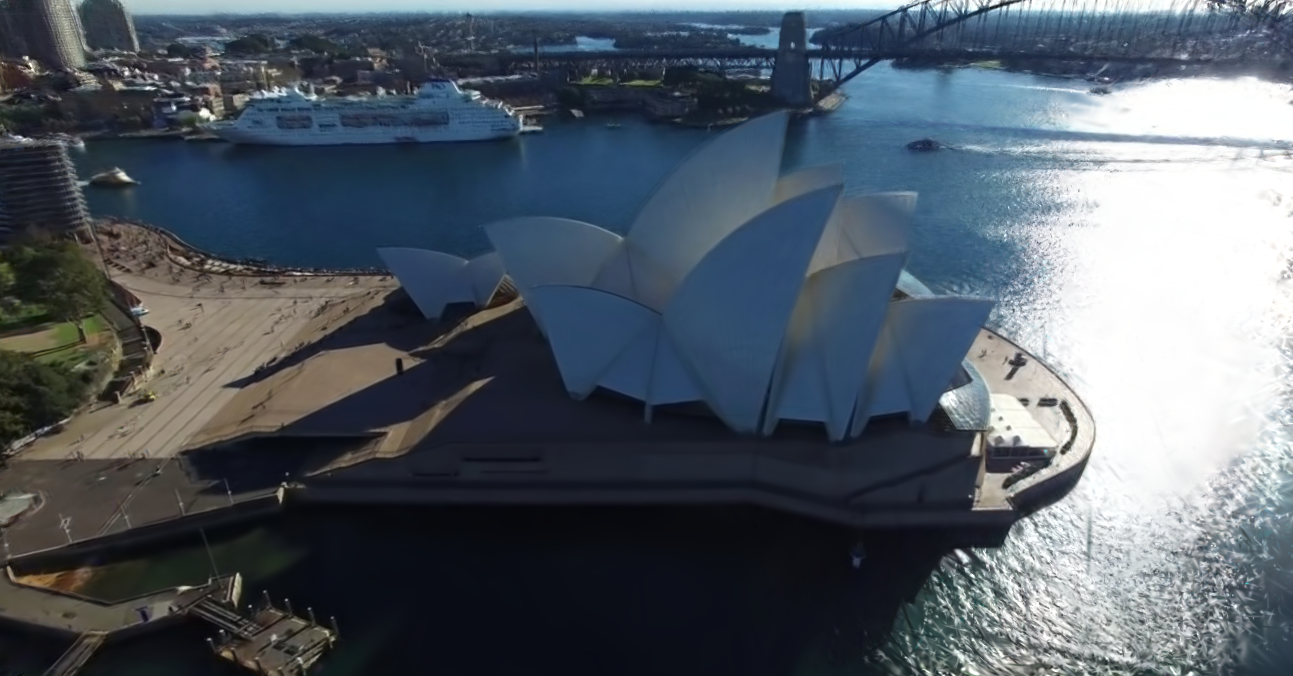}
      \caption*{Random + depth}
  \end{subfigure}
  \caption{Qualitative results on a scene from the OMMO dataset~\cite{lu2023large}. We find that the COLMAP and random initializations allocate fewer points to the surface of the water, which leads to over-smoothed reconstruction compared to the NeRF-based initialization and models with depth supervision.
  Please zoom in to see details. Please also refer to the interactive comparison in the supplementary material.}
  \label{fig:sydney_new}
\end{figure*}

\subsection{Datasets}

To best evaluate how the strategies under consideration perform in real-world scenarios, we focus our experiments on large scale, outdoor datasets.
The first dataset we employ is the Mip-NeRF 360 dataset~\cite{barron2022mip}, which includes several challenging indoor and outdoor scenes, and is a common benchmark for inverse rendering methods.
We also evaluate on the OMMO dataset~\cite{lu2023large}, which contains extremely large-scale scenes captured by drone videos.
This is representative of an application area where eliminating reliance on SFM initialization would be advantageous, as autonomous vehicles often have accurate navigation systems that could be used as an alternative to purely visual tracking.
These scenes also contain dynamic objects and strong view-dependence which degrade SFM point cloud coverage, as shown in Figure~\ref{fig:sidney_points}.

For these initial experiments, we continue to use camera parameters estimated by COLMAP to avoid the confounding effect of different camera parameters on our analysis.
To further demonstrate how our findings translate when SfM solutions are not available,
we also test a fully COLMAP-free pipeline on two SLAM datasets: Replica~\cite{replica19arxiv} and TUM~\cite{sturm12iros}.
For these datasets, we apply the volumetric pre-training and guidance strategies using cameras estimated by ORB-SLAM3~\cite{ORBSLAM3_TRO}, and compare the results to those using COLMAP cameras/initializations, as well as the results of a recent Gaussian Splatting SLAM method~\cite{monogs}.

\subsection{Analysis of random initialization}
\label{sec:rand_init_experiments}
\begin{table*}
    \caption{%
    \textbf{Random initializations.}
    We compare test PSNR values for the Mip-NeRF 360 dataset~\cite{barron2022mip} 
    resulting from different random initialization schemes to the PSNR values obtained with COLMAP initialization.
    We both retrieve metrics from the original paper (From \cite{kerbl3Dgaussians}), and re-run the official code release (re-run).
    We find that using a large constant $50\times 50\times 50$ bounding box as the initialization gives the best results.
    All models are trained to the maximum 30k iterations used in the original paper, and all random initializations start with 50k Gaussians, which we found to perform well across scenes and random distributions.
    }%
    \label{tab:rand_init}
    \begin{center}
    \resizebox{\linewidth}{!}{
    \setlength{\tabcolsep}{4pt}
    \begin{tabular}{@{}lcccccccR} \toprule
    & \multicolumn{8}{c}{Scene} \\
Initialization & Garden & Bicycle & Stump & Counter & Kitchen & Bonsai & Room & Average \\ 
\midrule
COLMAP (From~\cite{kerbl3Dgaussians}) & \cellcolor{color1}27.41 & \cellcolor{color1}25.25 & \cellcolor{color2}26.55 & \cellcolor{color2}28.70 & 30.32 & \cellcolor{color2}31.98 & \cellcolor{color2}30.63 & \cellcolor{color2}28.69 \\
COLMAP (Re-run) & \cellcolor{color2}27.29 & \cellcolor{color2}25.16 & \cellcolor{color1}26.82 & \cellcolor{color1}29.19 & \cellcolor{color1}31.50 & \cellcolor{color1}32.25 & \cellcolor{color1}31.43 & \cellcolor{color1}29.09 \\
\midrule
$3\times$ BBox (From~\cite{kerbl3Dgaussians}) & 22.19 & 21.05 & - & - & - & - & - & - \\
$3\times$ BBox (Re-run) & \cellcolor{color3}26.51 & 23.60 & 23.00 & 27.48 & \cellcolor{color3}30.33 & 29.91 & \cellcolor{color3}30.16 & 27.28 \\
$1.5\times$ BBox & 22.03 & 21.28 & 20.71 & 25.04 & 26.15 & 17.15 & 29.69 & 23.15 \\
$50^3$ BBox & 26.31 & \cellcolor{color3}24.12 & \cellcolor{color3}24.63 & \cellcolor{color3}28.04 & \cellcolor{color2}30.54 & \cellcolor{color3}30.74 & 29.90 & \cellcolor{color3}27.75 \\
    \bottomrule
    \end{tabular}
    }
    \end{center}
\end{table*}

We start our analysis by taking a deeper look at the conclusions drawn about random initialization in the original Gaussian Splatting publication.
Specifically, we try to \textit{reproduce} the reported results for the original proposed method for uniform random initialization.

According to the original text, the authors ``\textit{uniformly sample a cube with a size equal to three times the extent of the input camera’s bounding box}''~\cite{kerbl3Dgaussians}.
As shown in Table~\ref{tab:rand_init}, we find this strategy to give higher PSNR values than originally reported, with the original values more closely matching those of a bounding box with \textit{1.5$\times$} the extent of the cameras.
However, we find an even simpler strategy of using a very large
box centered at the origin without dependence on the camera distribution, covering a $50\times 50\times 50$ extent in COLMAP coordinates (effectively covering the entire space) to give even better results in average, so we use this as our baseline random initialization going forward.

As also shown in Table~\ref{tab:rand_init}, we also find that re-running the experiments with COLMAP initialization on the current official public code release results in \textit{slightly better} results than originally reported.
For all relevant experiments we provide both the original reported values as well as what we obtained re-running the code.

\subsection{Volumetric vs. SFM initialization}
\label{sec:nerfacc_init_experiments}

\begin{table*}[t]
    \caption{%
    \textbf{Volumetric initialization (Mip-NeRF 360).}
    We evaluate test PSNR on the Mip-NeRF 360 dataset~\cite{barron2022mip} when initializing the Gaussian Splatting model from INGP models trained for different amounts of time.
    }%
    \label{tab:nerfacc_init_360}
    \begin{center}    
    \resizebox{\linewidth}{!}{
    \setlength{\tabcolsep}{4pt}
    \begin{tabular}{@{}lcccccccR} \toprule
    & \multicolumn{8}{c}{Scene} \\
Initialization & Garden & Bicycle & Stump & Counter & Kitchen & Bonsai & Room & Average \\ 
\midrule
COLMAP (From~\cite{kerbl3Dgaussians}) & \cellcolor{color1}27.41 & \cellcolor{color1}25.25 & 26.55 & \cellcolor{color3}28.70 & 30.32 & \cellcolor{color3}31.98 & 30.63 & 28.69 \\
COLMAP (Re-run) & \cellcolor{color2}27.29 & \cellcolor{color2}25.16 & 26.82 & \cellcolor{color1}29.19 & 31.50 & \cellcolor{color2}32.25 & 31.43 & \cellcolor{color2}29.09 \\
\midrule
$50^3$ Random & 26.31 & 24.12 & 24.63 & 28.04 & 30.54 & 30.74 & 29.90 & 27.75 \\
\midrule
NerfAcc @ 5k & 27.21 & 23.90 & \cellcolor{color3}27.35 & 28.47 & \cellcolor{color2}31.82 & 31.88 & \cellcolor{color1}32.11 & \cellcolor{color3}28.96 \\
NerfAcc @ 10k & \cellcolor{color2}27.29 & \cellcolor{color3}24.14 & \cellcolor{color2}27.37 & \cellcolor{color2}28.77 & \cellcolor{color1}31.97 & \cellcolor{color1}32.27 & \cellcolor{color2}32.01 & \cellcolor{color1}29.12 \\
NerfAcc @ 30k & \cellcolor{color3}27.28 & 24.11 & \cellcolor{color1}27.38 & 28.50 & \cellcolor{color3}31.71 & 31.78 & \cellcolor{color3}31.93 & \cellcolor{color3}28.96 \\
    \bottomrule
    \end{tabular}
    }
    \end{center}
\end{table*}

\begin{table}[t]
    \caption{%
    \textbf{Volumetric initialization (OMMO).}
    We evaluate test PSNR on the OMMO dataset~\cite{lu2023large} when initializing the Gaussian Splatting model from INGP models trained for different lengths.
    }%
    \label{tab:nerfacc_init_ommo}
    \begin{center}
    \resizebox{\linewidth}{!}{
    \setlength{\tabcolsep}{12pt}
    \begin{tabular}{@{}lcccccccR} \toprule
    & \multicolumn{8}{c}{Scene} \\
Initialization & 03 & 05 & 06 & 10 & 13 & 14 & 15 & Average \\ 
\midrule
COLMAP & \cellcolor{color1}26.07 & \cellcolor{color2}28.36 & 26.76 & \cellcolor{color1}29.72 & \cellcolor{color1}32.55 & \cellcolor{color1}31.11 & \cellcolor{color3}30.46 & \cellcolor{color1}29.29 \\
\midrule
$50^3$ Random & 25.41 & 28.06 & 27.12 & \cellcolor{color2}29.49 & 31.08 & 30.26 & 29.25 & 28.67 \\
\midrule
NerfAcc @ 5k & \cellcolor{color2}25.57 & \cellcolor{color2}28.36 & \cellcolor{color2}27.70 & \cellcolor{color3}29.36 & \cellcolor{color2}32.24 & 30.83 & 30.45 & \cellcolor{color2}29.22 \\
NerfAcc @ 10k & 25.20 & \cellcolor{color1}28.37 & \cellcolor{color1}27.73 & 29.21 & \cellcolor{color3}32.21 & \cellcolor{color3}30.91 & \cellcolor{color2}30.52 & \cellcolor{color3}29.16 \\
NerfAcc @ 30k & \cellcolor{color3}25.52 & \cellcolor{color3}28.22 & \cellcolor{color3}27.51 & 29.23 & 31.83 & \cellcolor{color2}30.96 & \cellcolor{color1}30.58 & 29.12 \\
    \bottomrule
    \end{tabular}
    }
    \end{center}
\end{table}

With our baseline for random initialization selected, we move on to testing initialization based on a trained NeRF model.
As described in Section~\ref{sec:depth_supervision}, for each scene, we sample point cloud initializations consisting of 500,000 points in total from a NerfAcc-based INGP model trained to 5000, 10000, and 30000 iterations.
These initializations are loaded in exactly the same way as COLMAP point clouds, and training proceeds with the same standard settings.

As reported in Tables~\ref{tab:nerfacc_init_360} and~\ref{tab:nerfacc_init_ommo}, we find that these initializations near-universally outperform purely random initialization, and for some scenes outperform COLMAP initialization, even when using NeRF models trained for only 5000 iterations (about 30 seconds on our RTX A6000 GPU).

Based on these results, which show inconsistent improvement from training the initial NeRF model longer, we choose to proceed to the final round of experiments with the NeRF initializations from the 5000-iteration model weights, as this minimizes the additional compute added to the pipeline from NeRF pre-training.

\subsection{Depth distillation from NeRF}

\label{sec:depth_loss_experiments}

Next, we experiment with integrating direct depth guidance from the NeRF model into the Gaussian Splatting training pipeline (see Figure~\ref{fig:bicycle_depth} for examples).
Using the loss formulation in (\ref{eq:depth_loss}), we directly supervise predicted depth values from Gaussian Splatting with the pre-trained NeRF model, which enables more fine-grained transfer of structure.
It also helps avoid erroneous pruning of geometry as shown in Figure~\ref{fig:stump}.

\begin{table*}
    \caption{%
    \textbf{Structure distillation (Mip-NeRF 360).}
    We evaluate test PSNR on the Mip-NeRF 360 dataset~\cite{barron2022mip} when applying our depth loss for structure distillation from INGP models trained for 5k iterations.
    }%
    \label{tab:depth_loss_360}
    \begin{center}
    \resizebox{\linewidth}{!}{
    \setlength{\tabcolsep}{4pt}
    \begin{tabular}{@{}lcccccccR} \toprule
    & \multicolumn{8}{c}{Scene} \\
Initialization & Garden & Bicycle & Stump & Counter & Kitchen & Bonsai & Room & Average \\ 
\midrule
COLMAP (From~\cite{kerbl3Dgaussians}) & \cellcolor{color1}27.41 & \cellcolor{color1}25.25 & 26.55 & \cellcolor{color3}28.70 & 30.32 & \cellcolor{color2}31.98 & 30.63 & 28.69 \\
COLMAP (Re-run) & \cellcolor{color3}27.29 & \cellcolor{color3}25.16 & \cellcolor{color3}26.82 & \cellcolor{color1}29.19 & 31.50 & \cellcolor{color1}32.25 & \cellcolor{color3}31.43 & \cellcolor{color2}29.09 \\
\midrule
NerfAcc @ 5k & 27.21 & 23.90 & \cellcolor{color1}27.35 & 28.47 & \cellcolor{color2}31.82 & \cellcolor{color3}31.88 & \cellcolor{color1}32.11 & \cellcolor{color3}28.96 \\
\midrule
$50^3$ Random + Depth Loss & 26.75 & 24.05 & 25.63 & 28.56 & \cellcolor{color3}31.60 & 31.29 & 31.19 & 28.44 \\
NerfAcc @ 5k + Depth Loss & \cellcolor{color2}27.37 & \cellcolor{color2}25.22 & \cellcolor{color2}27.21 & \cellcolor{color2}28.85 & \cellcolor{color1}31.87 & 31.54 & \cellcolor{color2}31.63 & \cellcolor{color1}29.10 \\
    \bottomrule
    \end{tabular}
    }
    \end{center}
\end{table*}

\begin{table*}[t]
    \caption{%
    \textbf{Structure distillation (OMMO).}
    We evaluate test PSNR on the OMMO dataset~\cite{lu2023large} when applying our depth loss for structure distillation from INGP models trained for 5k iterations.
    }%
    \label{tab:depth_loss_ommo}
    \begin{center}
    \resizebox{\linewidth}{!}{
    \setlength{\tabcolsep}{12pt}
    \begin{tabular}{@{}lcccccccR} \toprule
    & \multicolumn{8}{c}{Scene} \\
Initialization & 03 & 05 & 06 & 10 & 13 & 14 & 15 & Average \\ 
\midrule
COLMAP & \cellcolor{color3}26.07 & \cellcolor{color2}28.36 & 26.76 & \cellcolor{color3}29.72 & \cellcolor{color2}32.55 & \cellcolor{color1}31.11 & \cellcolor{color2}30.46 & \cellcolor{color3}29.29 \\
\midrule
NerfAcc @ 5k & 25.57 & \cellcolor{color2}28.36 & \cellcolor{color3}27.70 & 29.36 & 32.24 & \cellcolor{color3}30.83 & \cellcolor{color3}30.45 & 29.22 \\
\midrule
$50^3$ Random + Depth Loss & \cellcolor{color1}26.78 & \cellcolor{color3}28.31 & \cellcolor{color2}28.42 & \cellcolor{color1}30.67 & \cellcolor{color3}32.50 & 30.54 & 30.29 & \cellcolor{color2}29.64 \\
NerfAcc @ 5k + Depth Loss & \cellcolor{color2}26.48 & \cellcolor{color1}28.50 & \cellcolor{color1}28.89 & \cellcolor{color2}30.54 & \cellcolor{color1}32.74 & \cellcolor{color2}31.07 & \cellcolor{color1}30.76 & \cellcolor{color1}29.85 \\
    \bottomrule
    \end{tabular}
    }
    \end{center}
\end{table*}

We report the results of these experiments in Tables~\ref{tab:depth_loss_360} and~\ref{tab:depth_loss_ommo}, which show that the best final results are achieved with a combination of NeRF-based initialization and NeRF depth supervision.

\subsection{COLMAP-free training}

\begin{table*}[t]
    \caption{
    \textbf{SLAM camera estimates (TUM).}
    We compare the performance of NerfAcc initialization and depth loss, using cameras estimated by ORB-SLAM3~\cite{ORBSLAM3_TRO}, to that of standard Gaussian Splatting with COLMAP cameras and a recent Gaussian Splatting SLAM method on the TUM dataset~\cite{sturm12iros}.
    Along with test PSNR, we report the total run-time of each pipeline from images to the trained model.
    }
    \label{tab:tum}
    \begin{center}
    \resizebox{\linewidth}{!}{
    \setlength{\tabcolsep}{4pt}
    \begin{tabular}{@{}lcccc@{}} \toprule
& \multicolumn{3}{c}{Scene} & \\
Model & Desk & XYZ & Long & Average \\ 
\midrule
      COLMAP & \textbf{27.51} | 0:15:58 & \textbf{30.04} | 7:38:42 & 28.19 | 1:14:06 & \textbf{28.58} | 3:02:55 \\
\midrule
      MonoGS~\cite{monogs} & 20.90 | \textbf{0:06:46} & 22.38 | 0:19:58 & 22.41 | 0:16:01 & 21.90 | 0:14:15 \\
\midrule
NerfAcc @ 5k & 24.91 | 0:07:05 & 27.01 | \textbf{0:09:47} & 28.15 | \textbf{0:07:36} & 26.69 | \textbf{0:08:09}\\
NerfAcc @ 5k + Depth Loss & 25.01 | 0:08:22 & 27.24 | 0:18:30 & \textbf{28.27} | 0:12:10 & 26.84 | 0:13:01 \\
    \bottomrule
    \end{tabular}
    }
    \end{center}
\end{table*}

\begin{table*}[t!]
    \caption{
    \textbf{SLAM camera estimates (Replica).}
    We compare the performance of NerfAcc initialization and depth loss, using cameras estimated by ORB-SLAM3~\cite{ORBSLAM3_TRO}, to that of standard Gaussian Splatting with COLMAP cameras and a recent Gaussian Splatting SLAM method on the Replica dataset~\cite{replica19arxiv}.
    Along with test PSNR, we report the total run-time of each pipeline from images to the trained model.
    }%
    \label{tab:replica}
    \begin{center}
    \resizebox{\linewidth}{!}{
    \setlength{\tabcolsep}{4pt}
    \begin{tabular}{@{}lccccccccc@{}} \toprule
& \multicolumn{8}{c}{Scene} & \\
Model & Office0 & Office1 & Office2 & Office3 & Office4 & Room0 & Room1 & Room2 & Average \\ 
\midrule
      COLMAP & \textbf{44.65} | 0:38:33 & \textbf{44.20} | 0:38:51 & \textbf{40.54} | 1:01:20 & \textbf{40.61} | 0:52:18 & \textbf{38.46} | 0:35:05 & \textbf{39.40} | 1:13:51 & 32.59 | 2:55:00 & \textbf{41.66} | 0:49:44 & \textbf{40.26} | 1:05:35\\
\midrule
      MonoGS~\cite{monogs} & 34.42 | 0:22:11 & 35.14 | 0:21:51 & 28.11 | 0:24:30 & 29.45 | 0:24:14 & 27.49 | 0:23:38 & 28.07 | 0:26:19 & 26.58 | 0:24:10 & 28.11 | 0:24:13 & 29.67 | 0:23:53	\\
\midrule
NerfAcc @ 5k & 42.94 | \textbf{0:09:47} & 32.75 | \textbf{0:07:47} & 35.70 | \textbf{0:13:52} & 35.87 | \textbf{0:11:52} & 35.09 | \textbf{0:10:50} & 34.86 | \textbf{0:18:00} & \textbf{36.54}	| \textbf{0:14:52} & 38.33 | \textbf{0:12:56} & 36.51 | \textbf{0:12:29} \\
NerfAcc @ 5k + Depth Loss & 43.03 | 0:19:00 & 33.17 | 0:17:51 & 35.76 | 0:22:52 & 35.80 | 0:21:51 & 35.06 | 0:20:52 & 34.88 | 0:25:58 & 36.49 | 0:22:55 & 38.32 | 0:21:49 & 36.56 | 0:21:38 \\
    \bottomrule
    \end{tabular}
    }
    \end{center}
\end{table*}

Finally, we combine the best-performing strategies from the previous experiments with cameras estimated by a monocular SLAM method (ORB-SLAM3~\cite{ORBSLAM3_TRO}) which is much faster than COLMAP.
We also compare to a recent method that combines monocular SLAM and Gaussian Splatting (MonoGS) to estimate cameras during training~\cite{monogs}.
As shown in Tables~\ref{tab:tum} and~\ref{tab:replica}, we find that the approaches we tested result in higher PSNR on average than MonoGS, and run 
significantly
faster than the COLMAP-based pipeline, albeit with a somewhat lower PSNR, likely due to less accurate pose estimation.
We include video renders of sample sequences in the Supplementary Material for comparison.
The run times for the ORB-SLAM3 experiments include the SLAM tracking times themselves, about 40ms per frame, as well as the NerfAcc training and rendering time, about 300ms per frame if including depth supervision, or about 45ms per frame if only sampling a point cloud.

To be as fair as possible, we run COLMAP for these scenes with sequential matching, which only works for video input but results in much faster camera estimation.
Despite this, we found COLMAP to still be extremely slow (20$\times$ on average on TUM compared to the NerfAcc init pipeline and up to $45\times$ slower) for long sequences.
We also reduce the size of the NerfAcc-generated point cloud to 50,000 points to more closely match the SFM point cloud, which is much smaller for these indoor non-360 scenes.
All timed experiments are run on the same machine with an RTX A6000 GPU.

\section{Conclusion}
\label{sec:conclusion}

We performed an analysis of several approaches for avoiding SFM initialization in training of Gaussian Splatting.
Through our experiments, we found that random initialization can perform better than expected based on the results of the original Gaussian Splatting paper. We also found that use of a small amount of time spent pre-training a NeRF model can enable results on par with or better than those achieved by COLMAP initialization by providing both a dense initial point cloud and depth supervision throughout training.
We also tested these strategies with SLAM-estimated cameras and found that they can result in significant time savings across the entire pipeline.

\paragraph{Limitations}
The most immediate limitation in applying the strategies we have described to avoid requiring SFM solutions for Gaussian Splatting is that, as our SLAM experiments show, SFM is still the most reliable way of obtaining the camera estimates necessary to train for many use cases.
As such, careful consideration of the target application is needed to determine whether alternatives to SFM are available which can provide sufficiently accurate camera poses.

Another potential complication to including these strategies in a robust pipeline is the NeRF training. 
While we found un-modified NerfAcc to perform well on Mip-NeRF 360 and OMMO, the default settings were not able to reconstruct the Tanks and Temples~\cite{knapitsch2017tanks} scenes well enough to be useful for supervision.
This suggests that more work might be needed to automate the NeRF configuration process to avoid adding to the tuning workload of the entire pipeline.

\section{Acknowledgements} %
This work was supported in part by the Natural Sciences and Engineering Research Council of Canada (NSERC) Discovery Grant [2023-05617], NSERC Collaborative Research and Development Grant, the SFU Visual Computing Research Chair, Google, Digital Research Alliance of Canada, and Advanced Research Computing at the University of British Columbia.

{\small
\bibliographystyle{ieeenat_fullname}
\bibliography{references}
}

\newpage
\appendix
\section{Appendix}
\label{sec:appendix_section}

\begin{figure*}[!]
  \centering
\begin{subfigure}{.245\textwidth}
      \centering
      \includegraphics[width=\textwidth]{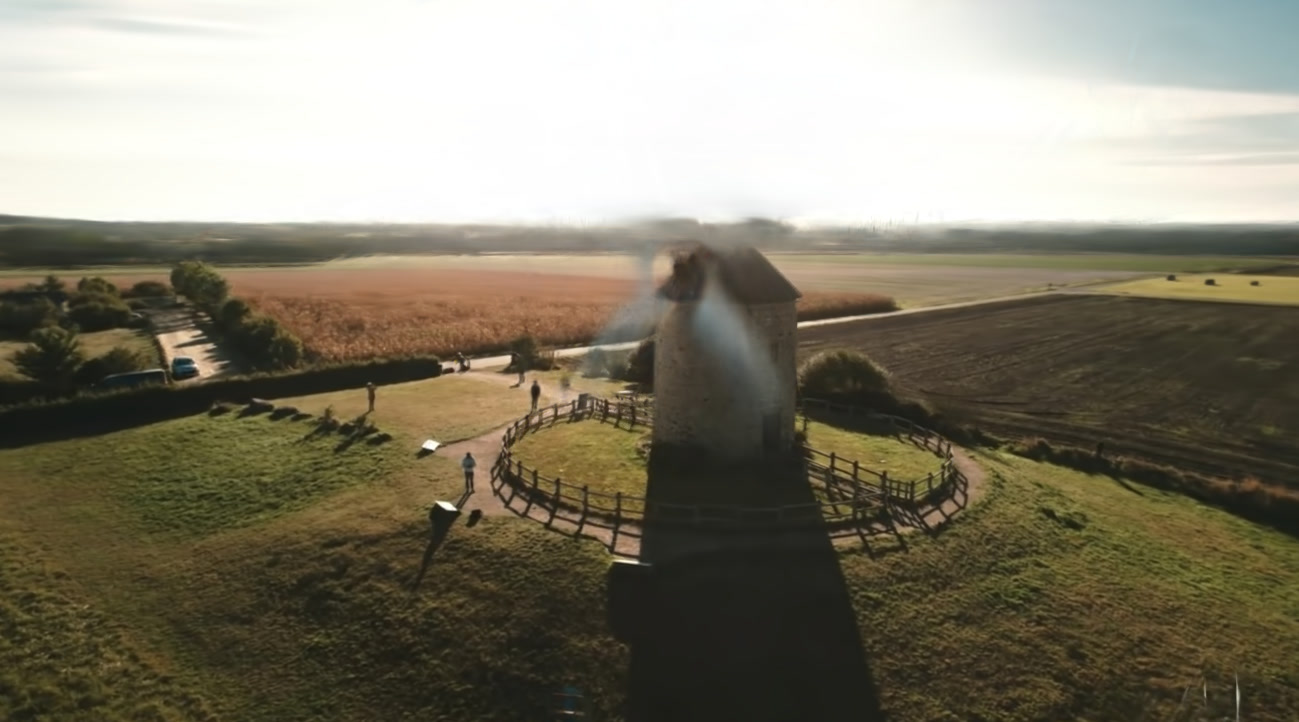}
  \end{subfigure}
\begin{subfigure}{.245\textwidth}
      \centering
      \includegraphics[width=\textwidth]{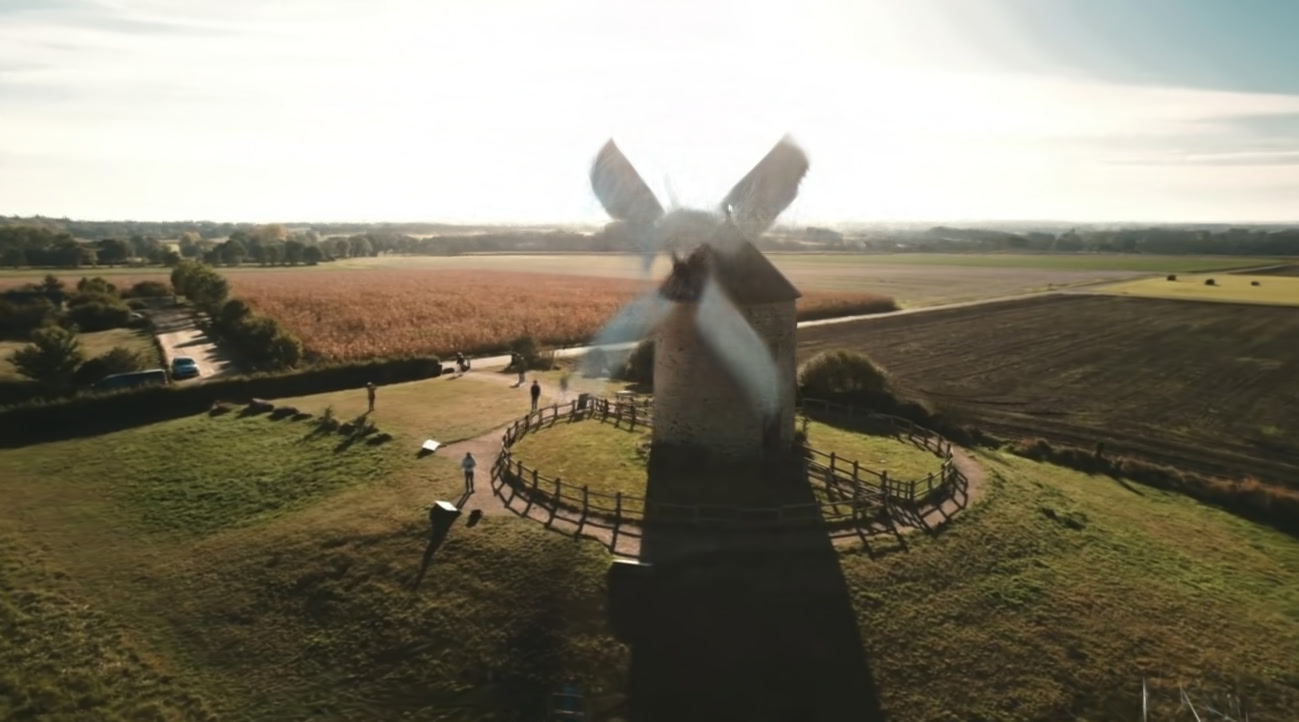}
  \end{subfigure}
\begin{subfigure}{.245\textwidth}
      \centering
      \includegraphics[width=\textwidth]{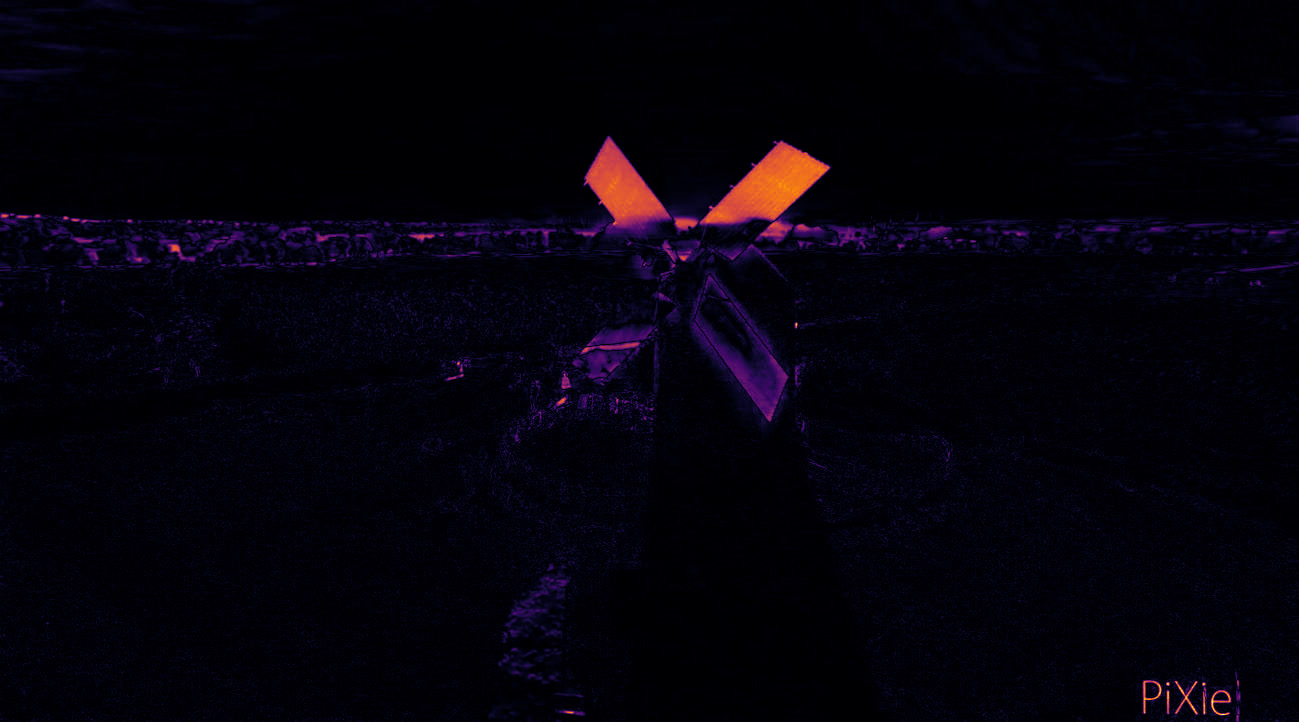}
  \end{subfigure}
\begin{subfigure}{.245\textwidth}
      \centering
      \includegraphics[width=\textwidth]{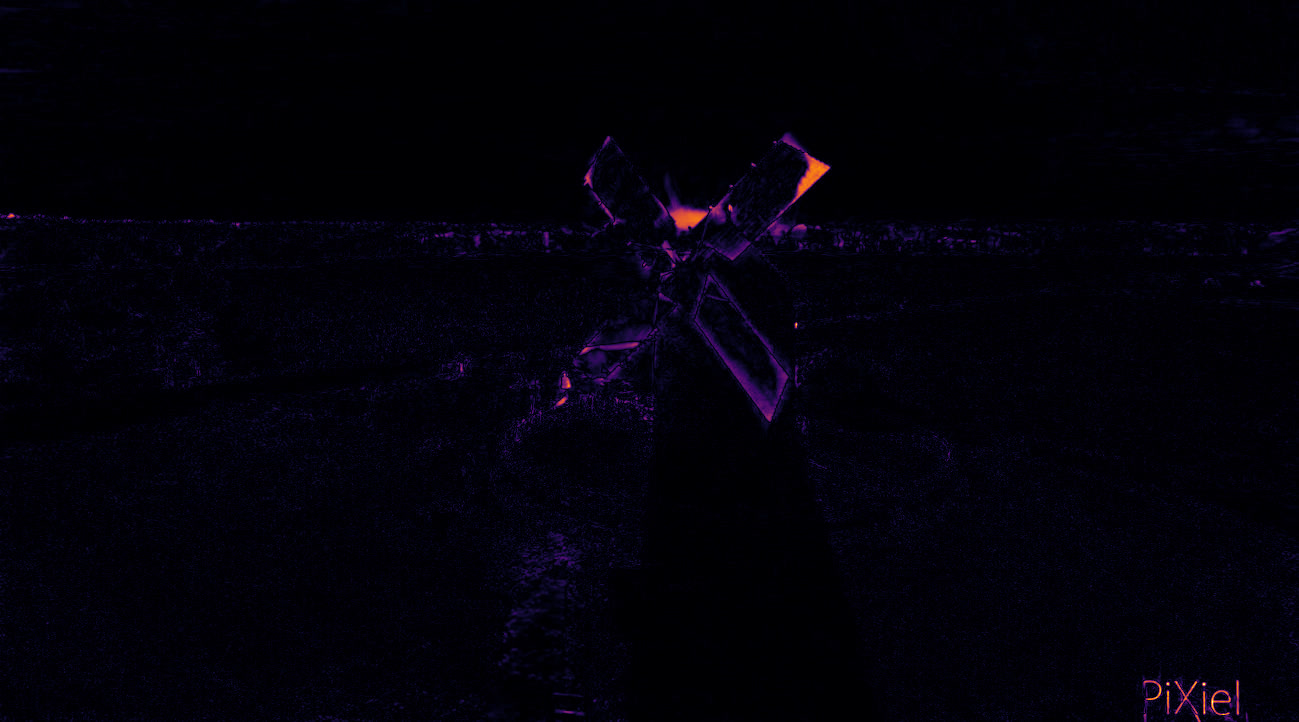}
\end{subfigure}
  
\begin{subfigure}{.245\textwidth}
      \centering
      \includegraphics[width=\textwidth]{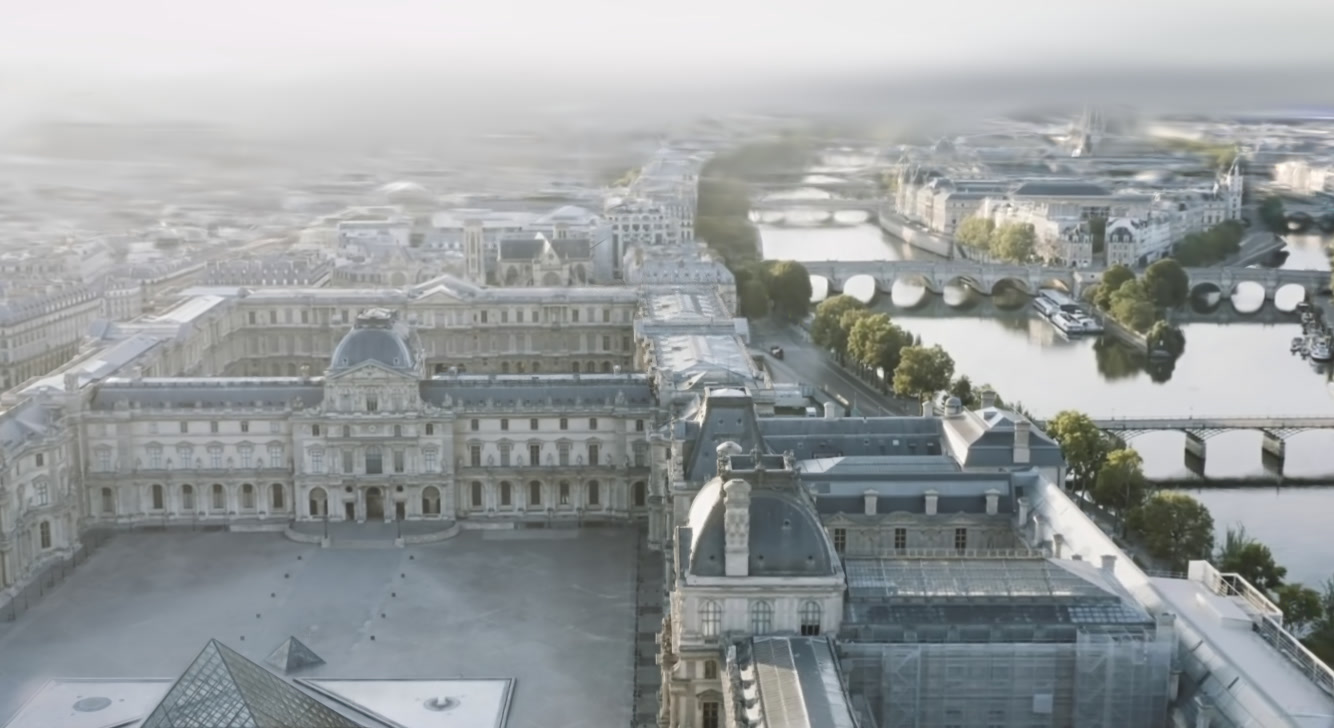}
  \end{subfigure}
\begin{subfigure}{.245\textwidth}
      \centering
      \includegraphics[width=\textwidth]{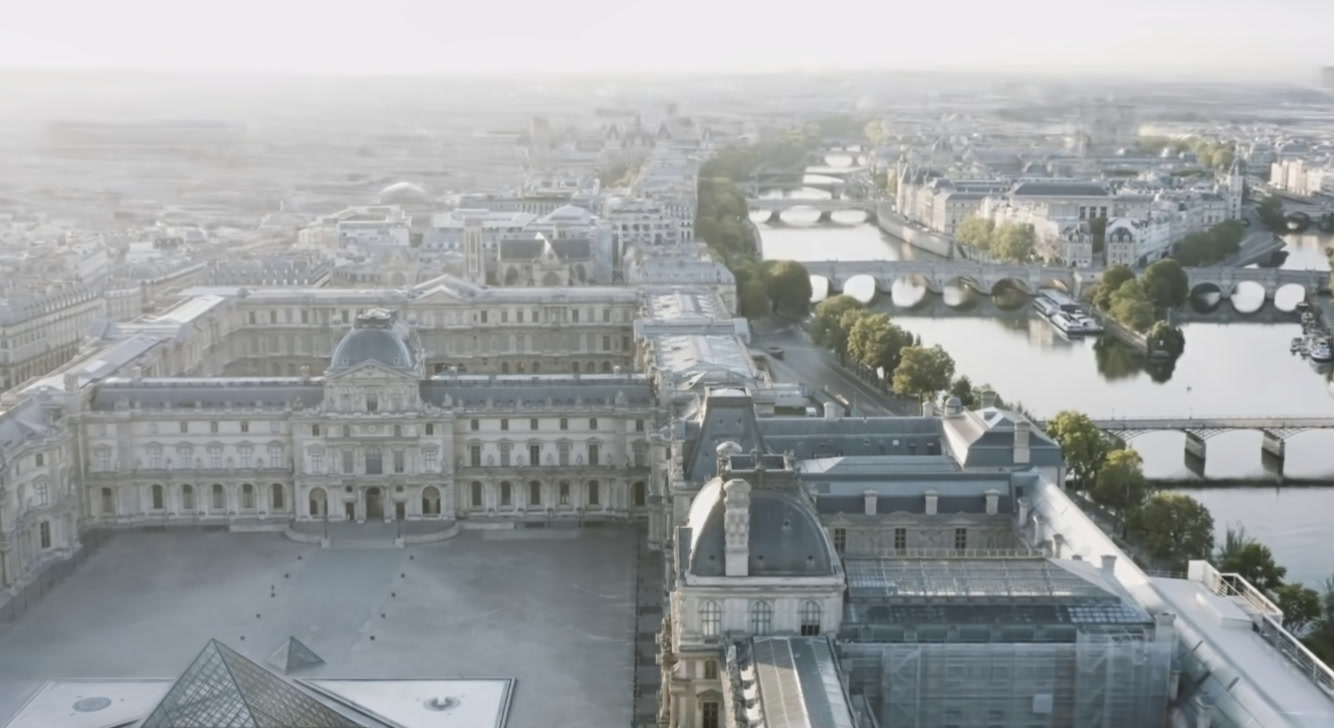}
  \end{subfigure}
\begin{subfigure}{.245\textwidth}
      \centering
      \includegraphics[width=\textwidth]{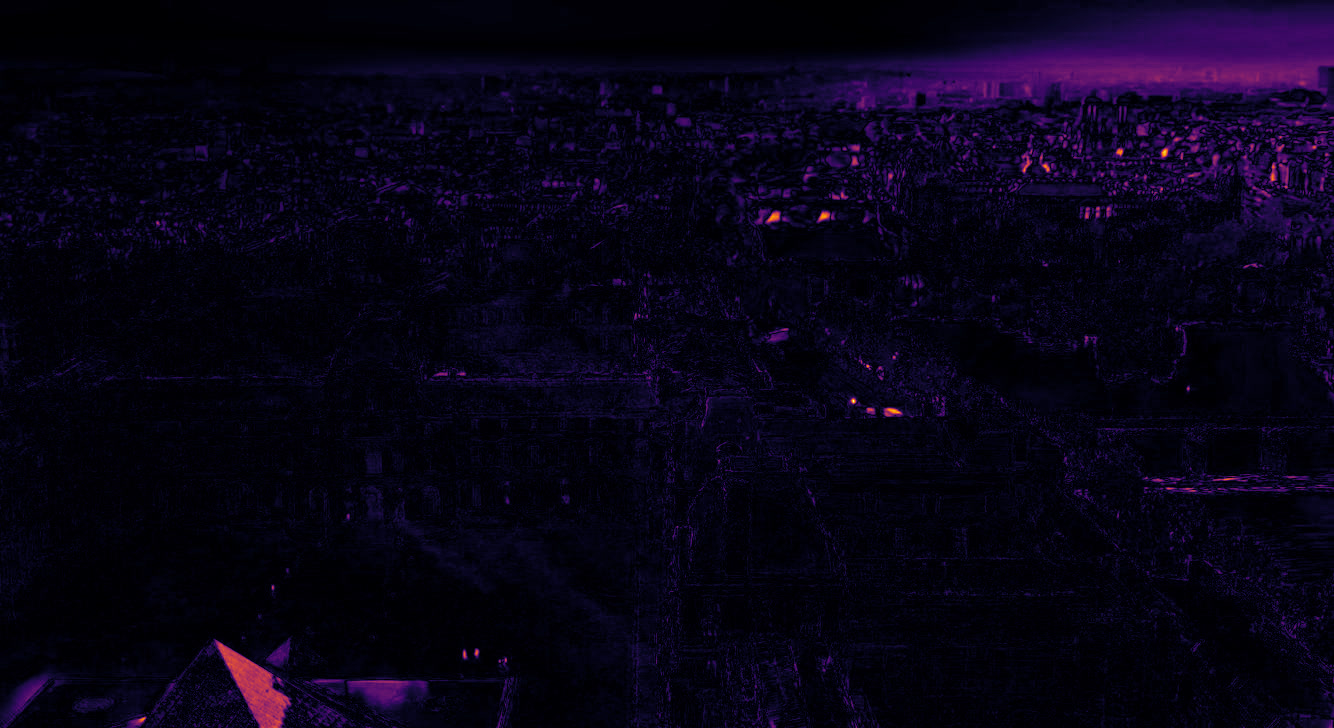}
  \end{subfigure}
\begin{subfigure}{.245\textwidth}
      \centering
      \includegraphics[width=\textwidth]{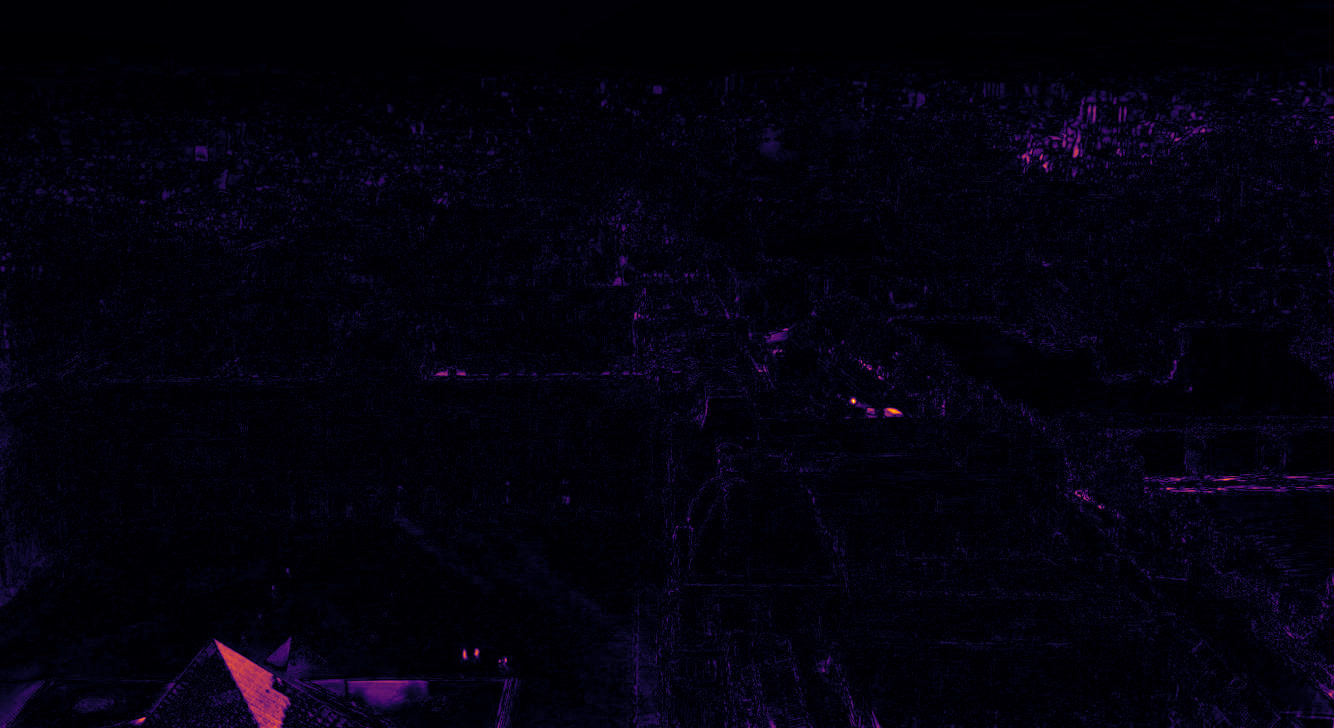}
\end{subfigure}

\begin{subfigure}{.245\textwidth}
      \centering
      \includegraphics[width=\textwidth]{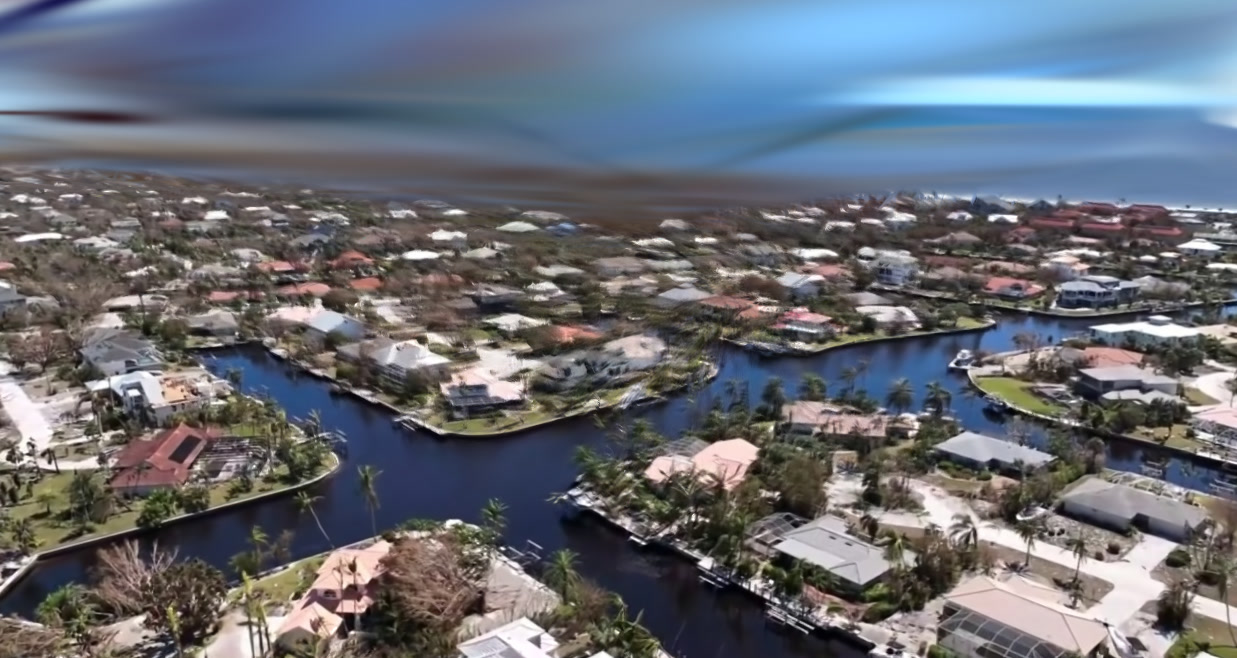}
  \end{subfigure}
\begin{subfigure}{.245\textwidth}
      \centering
      \includegraphics[width=\textwidth]{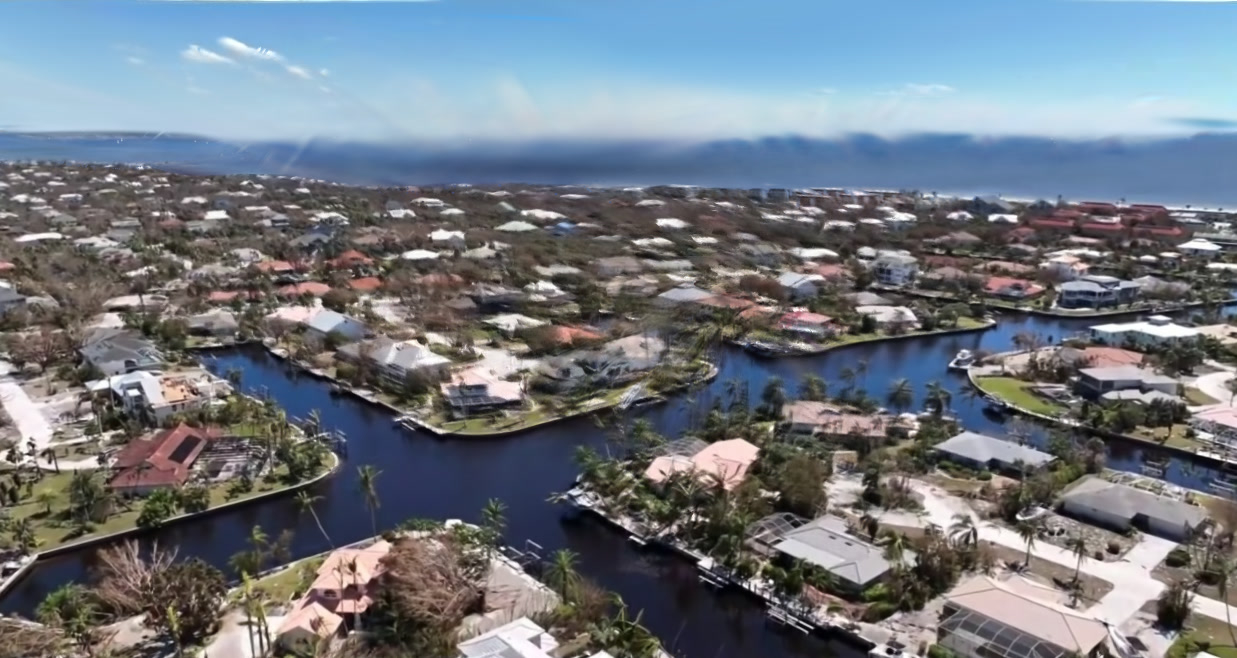}
  \end{subfigure}
\begin{subfigure}{.245\textwidth}
      \centering
      \includegraphics[width=\textwidth]{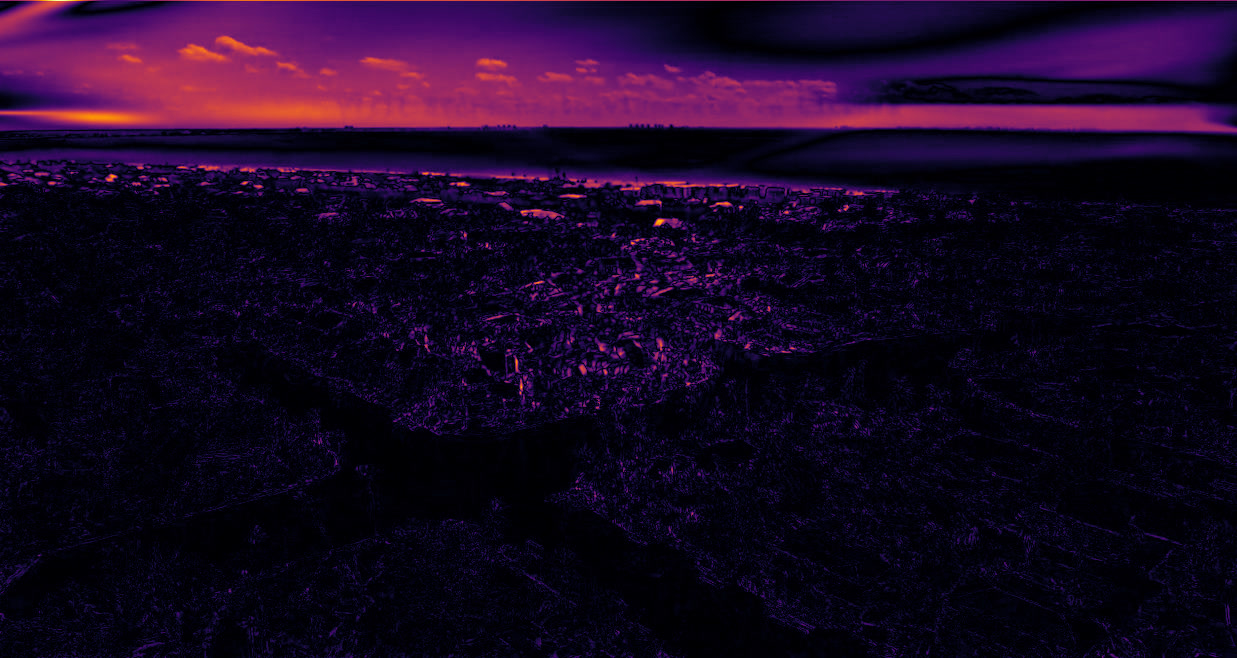}
  \end{subfigure}
\begin{subfigure}{.245\textwidth}
      \centering
      \includegraphics[width=\textwidth]{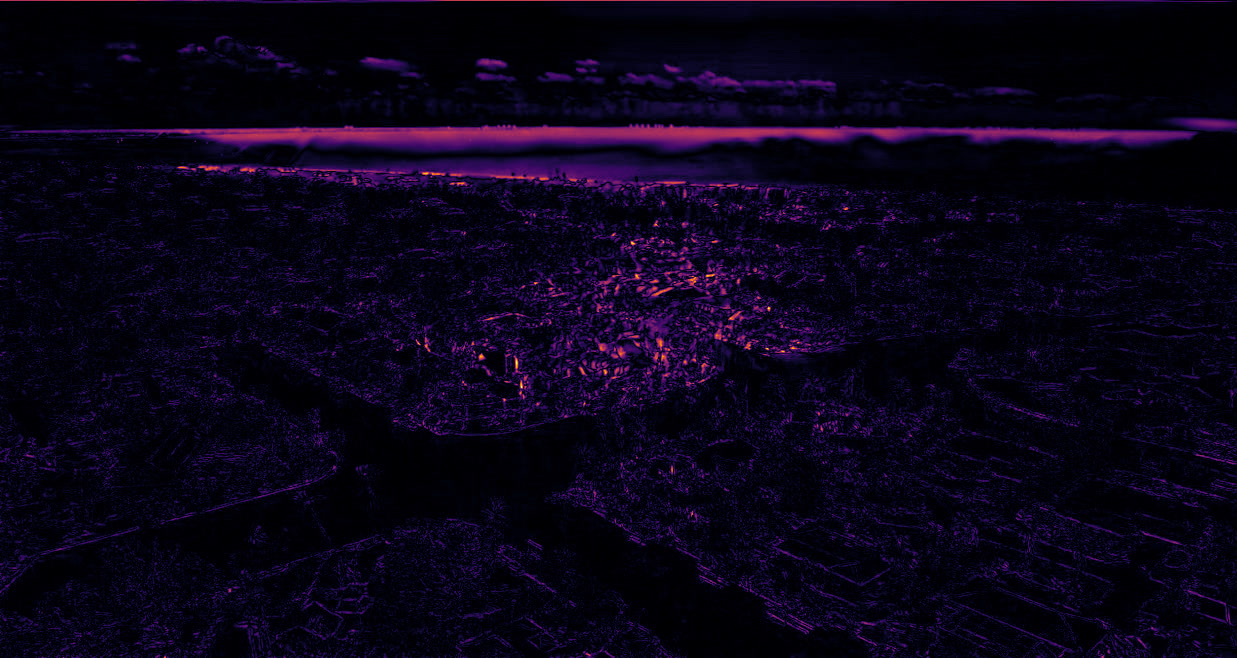}
\end{subfigure}

\begin{subfigure}{.245\textwidth}
      \centering
      \includegraphics[width=\textwidth]{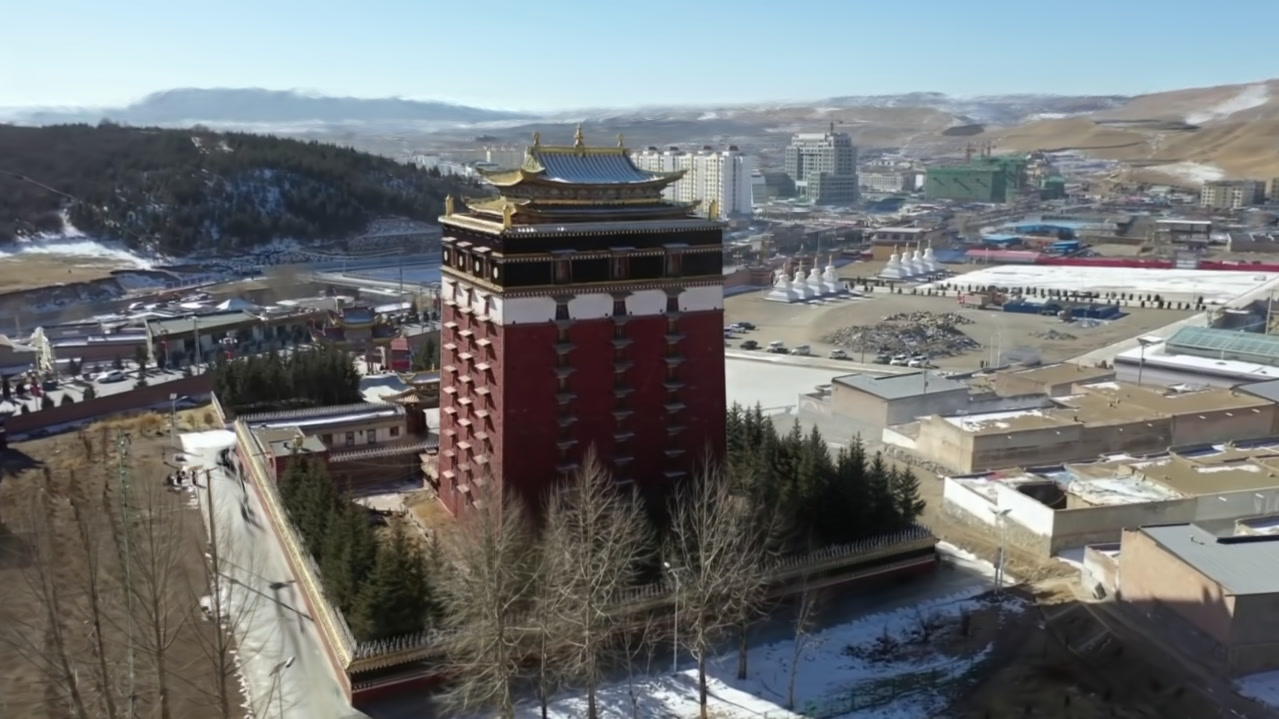}
  \end{subfigure}
\begin{subfigure}{.245\textwidth}
      \centering
      \includegraphics[width=\textwidth]{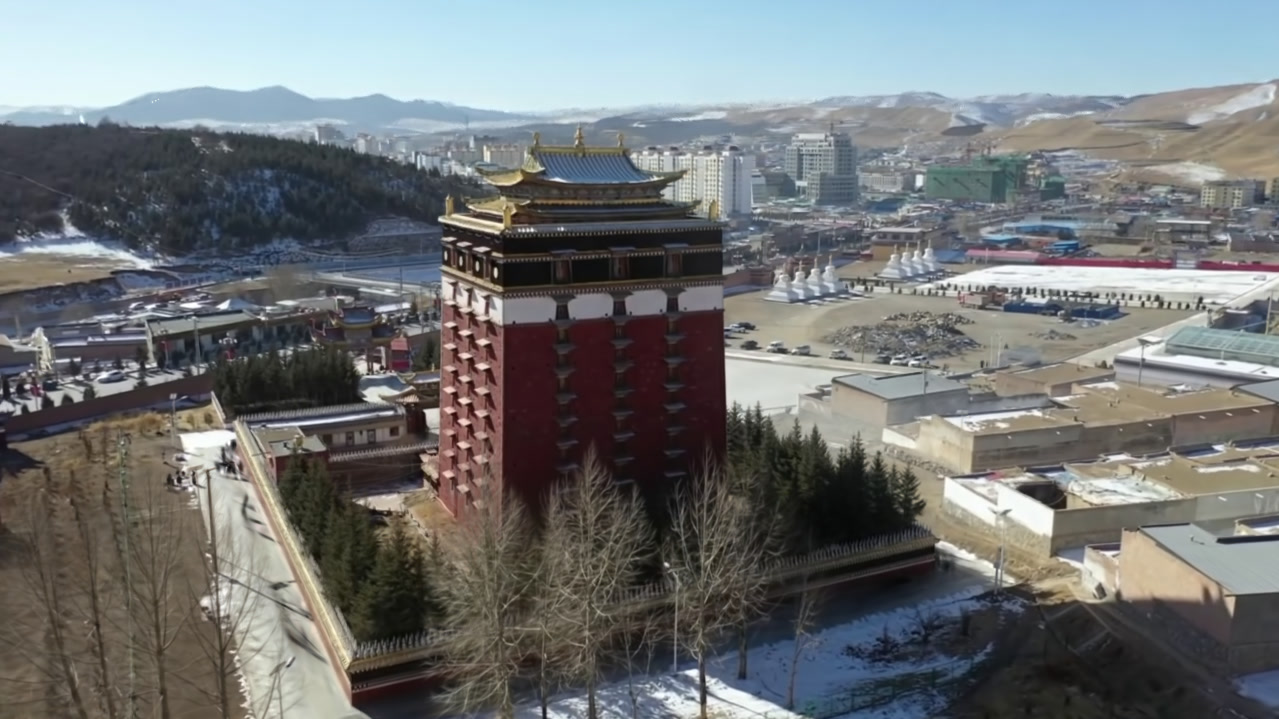}
  \end{subfigure}
\begin{subfigure}{.245\textwidth}
      \centering
      \includegraphics[width=\textwidth]{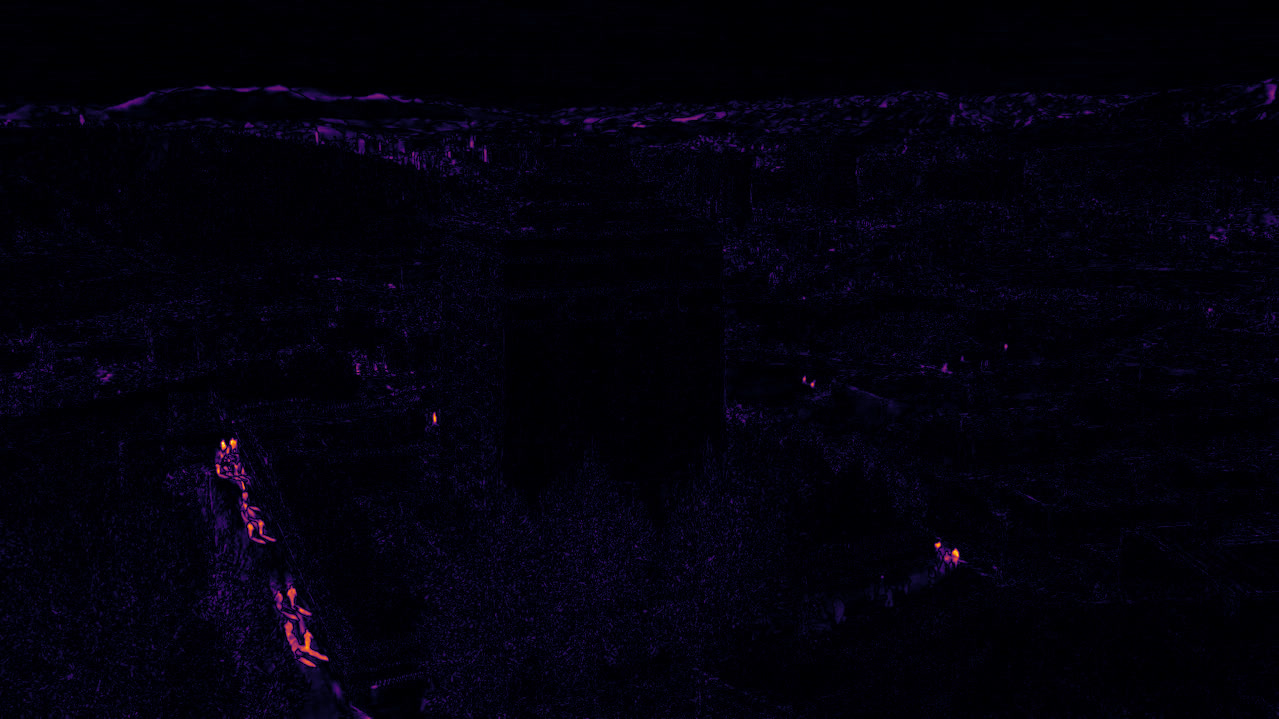}
  \end{subfigure}
\begin{subfigure}{.245\textwidth}
      \centering
      \includegraphics[width=\textwidth]{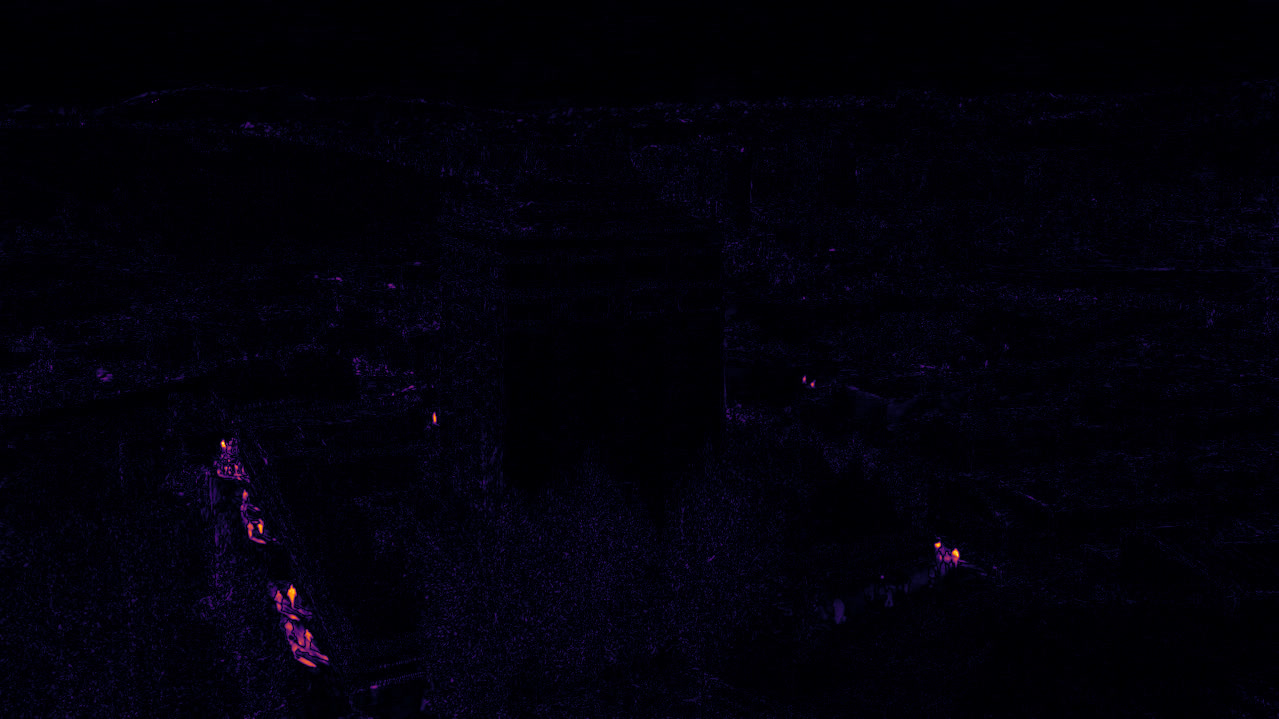}
\end{subfigure}

\begin{subfigure}{.245\textwidth}
      \centering
      \includegraphics[width=\textwidth]{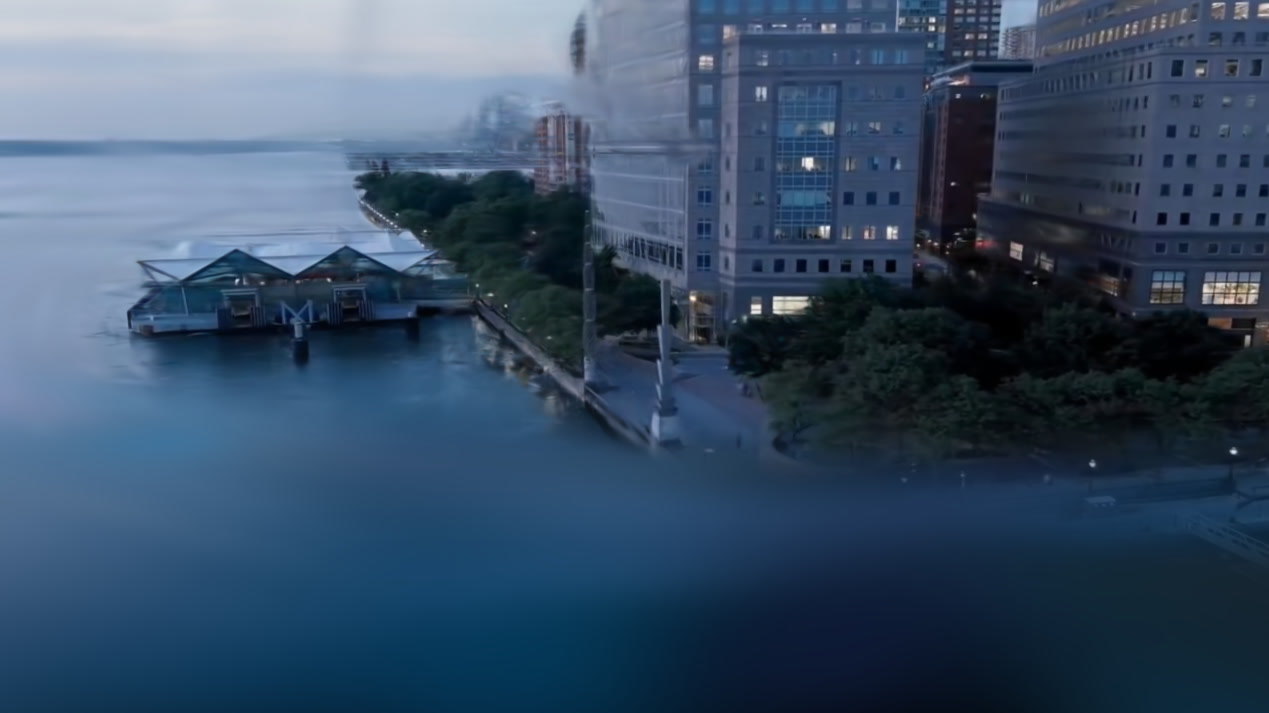}
  \end{subfigure}
\begin{subfigure}{.245\textwidth}
      \centering
      \includegraphics[width=\textwidth]{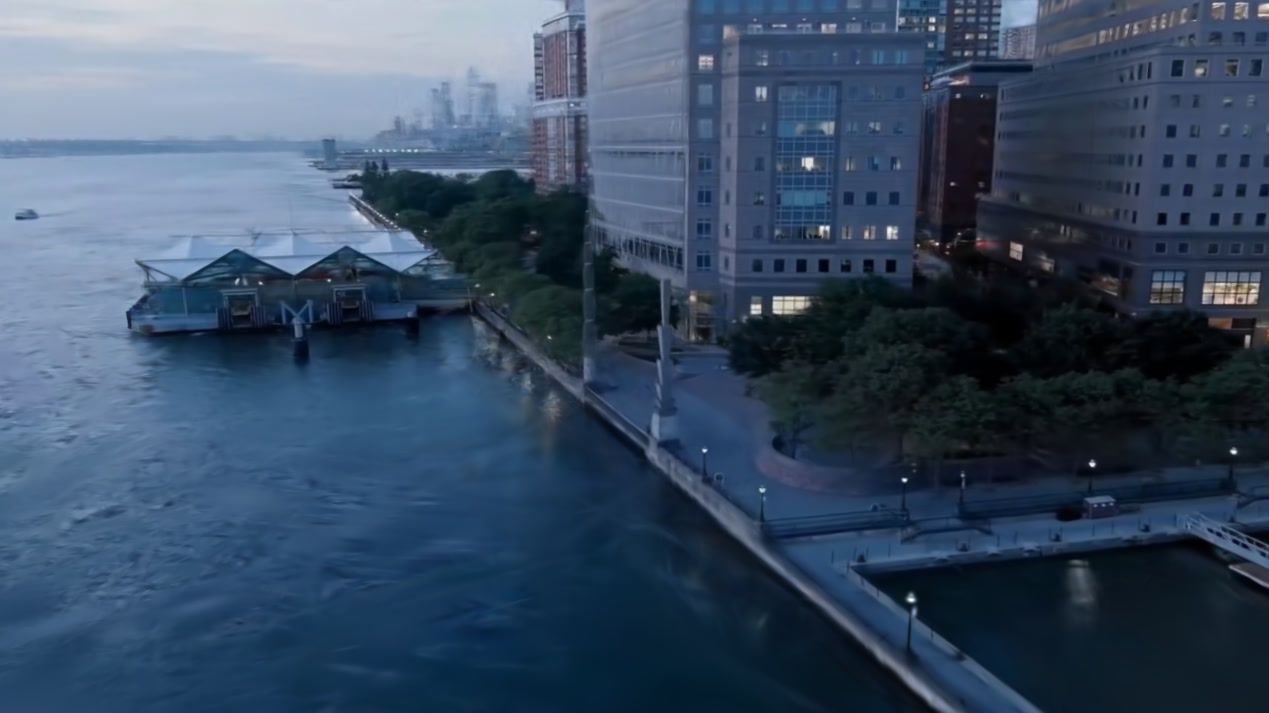}
  \end{subfigure}
\begin{subfigure}{.245\textwidth}
      \centering
      \includegraphics[width=\textwidth]{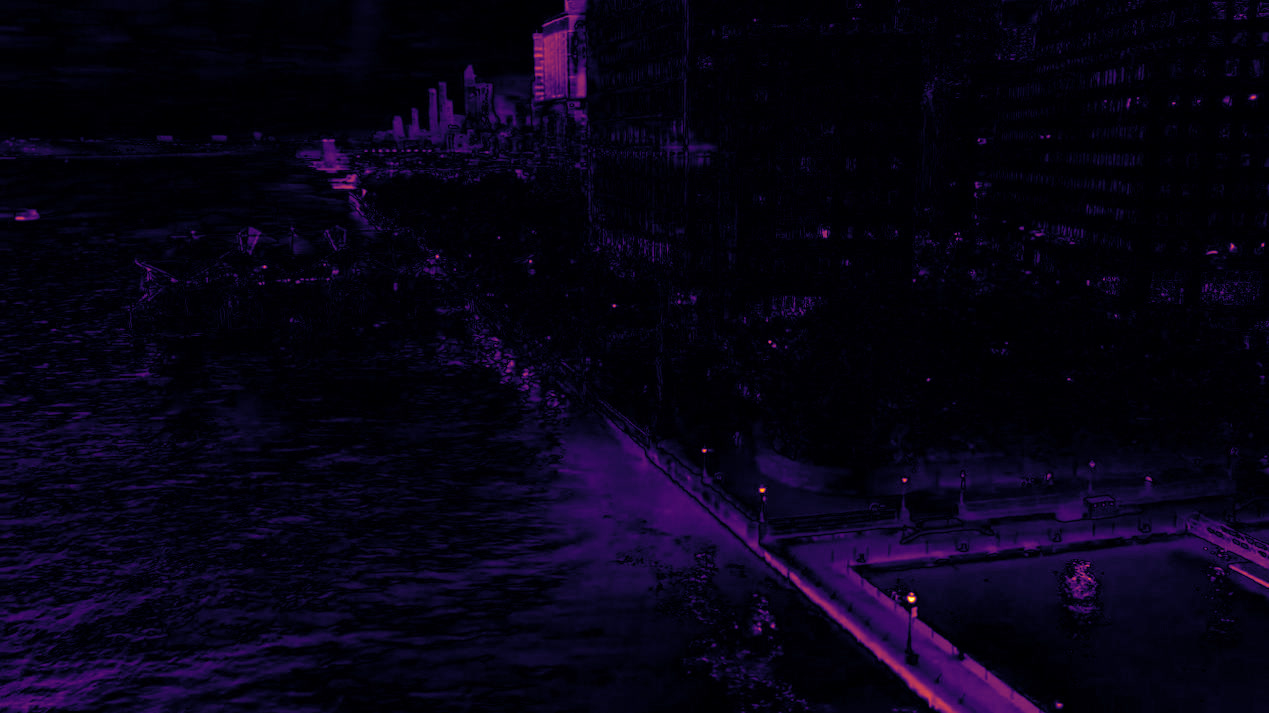}
  \end{subfigure}
\begin{subfigure}{.245\textwidth}
      \centering
      \includegraphics[width=\textwidth]{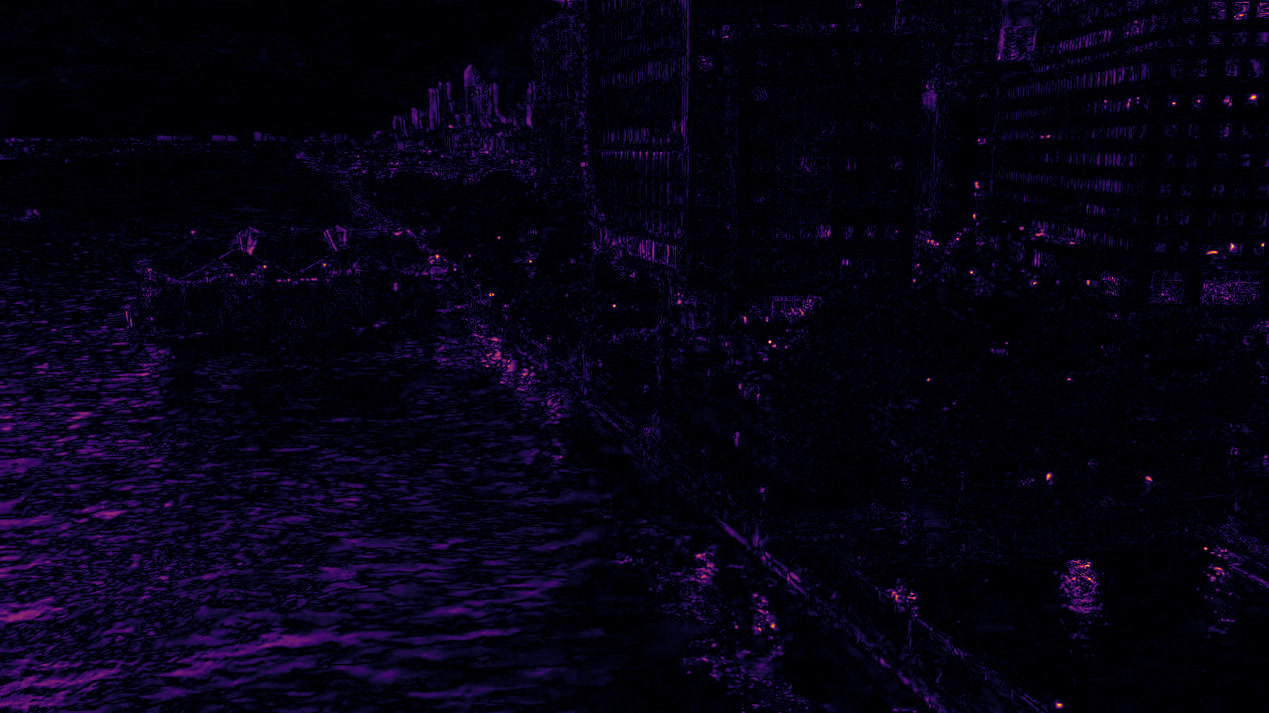}
\end{subfigure}

\begin{subfigure}{.245\textwidth}
      \centering
      \includegraphics[width=\textwidth]{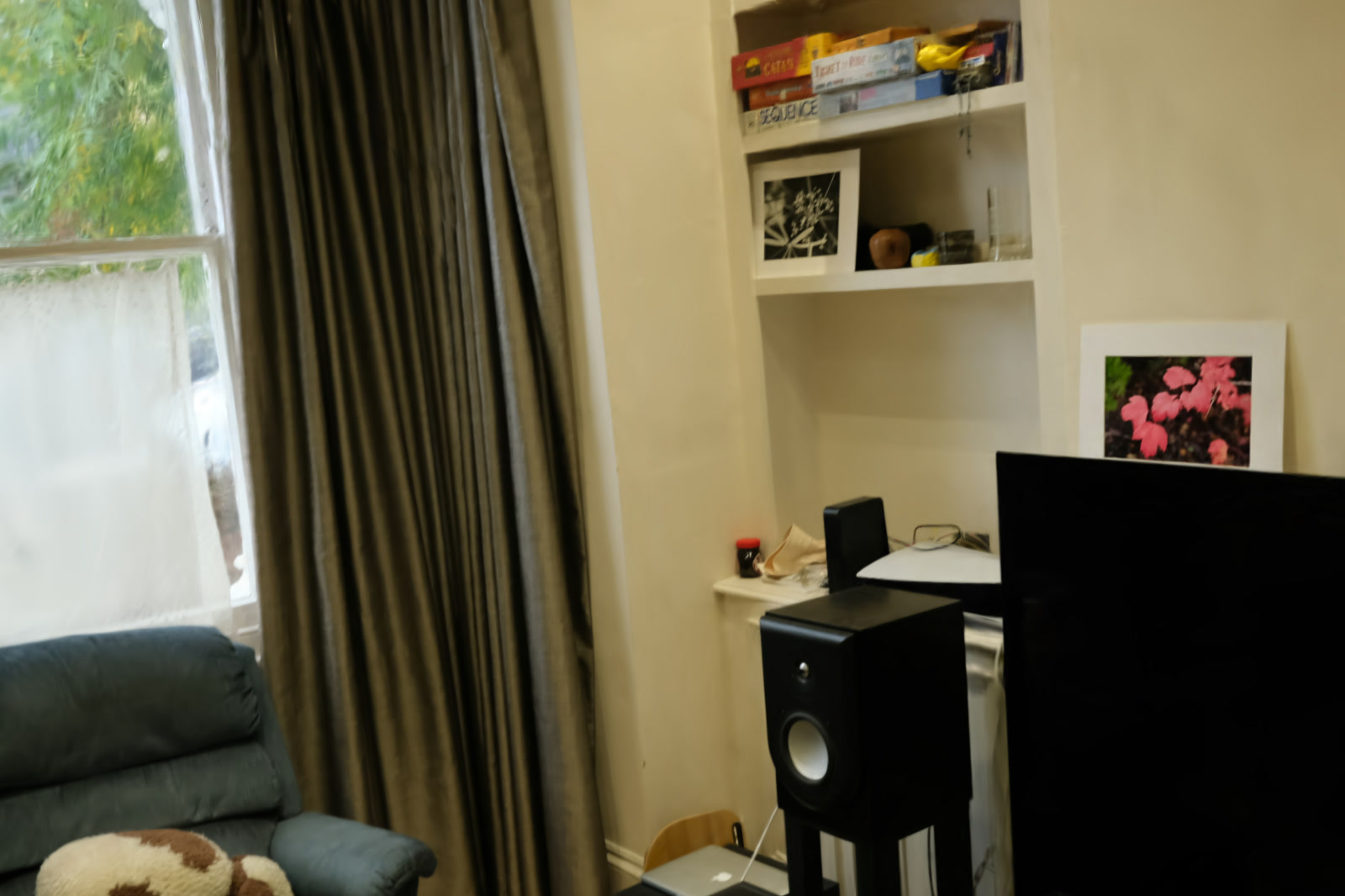}
  \end{subfigure}
\begin{subfigure}{.245\textwidth}
      \centering
      \includegraphics[width=\textwidth]{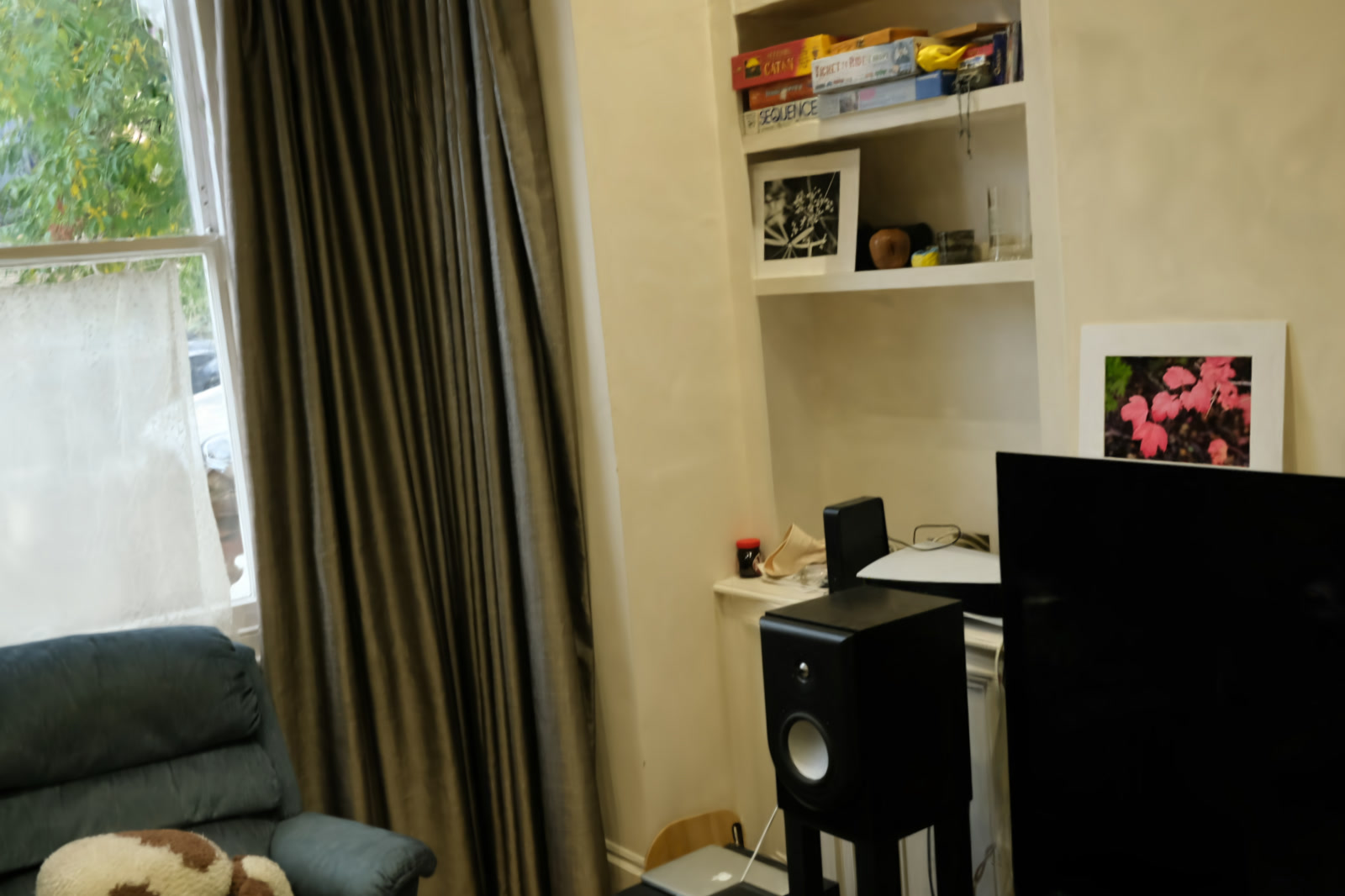}
  \end{subfigure}
\begin{subfigure}{.245\textwidth}
      \centering
      \includegraphics[width=\textwidth]{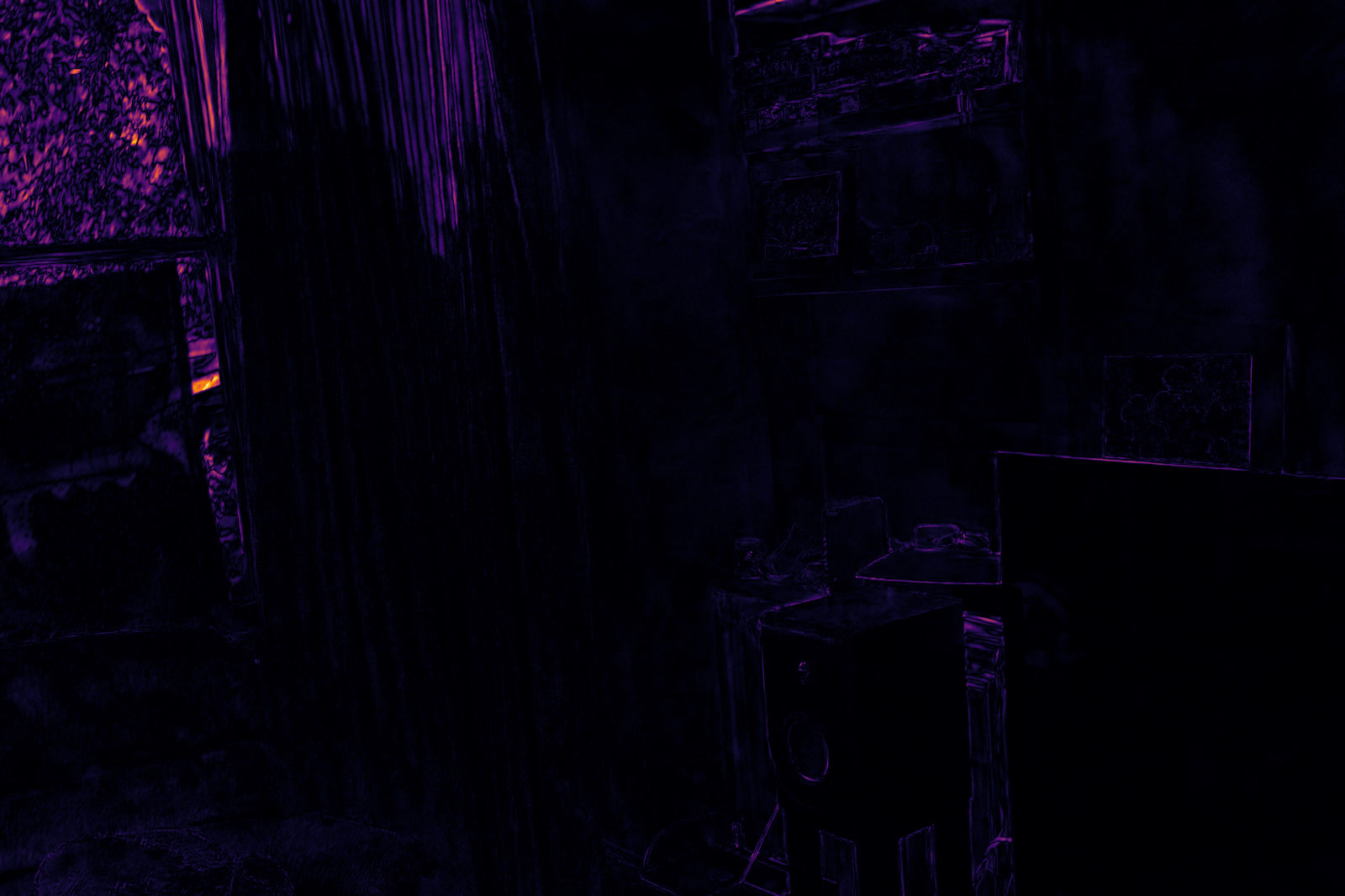}
  \end{subfigure}
\begin{subfigure}{.245\textwidth}
      \centering
      \includegraphics[width=\textwidth]{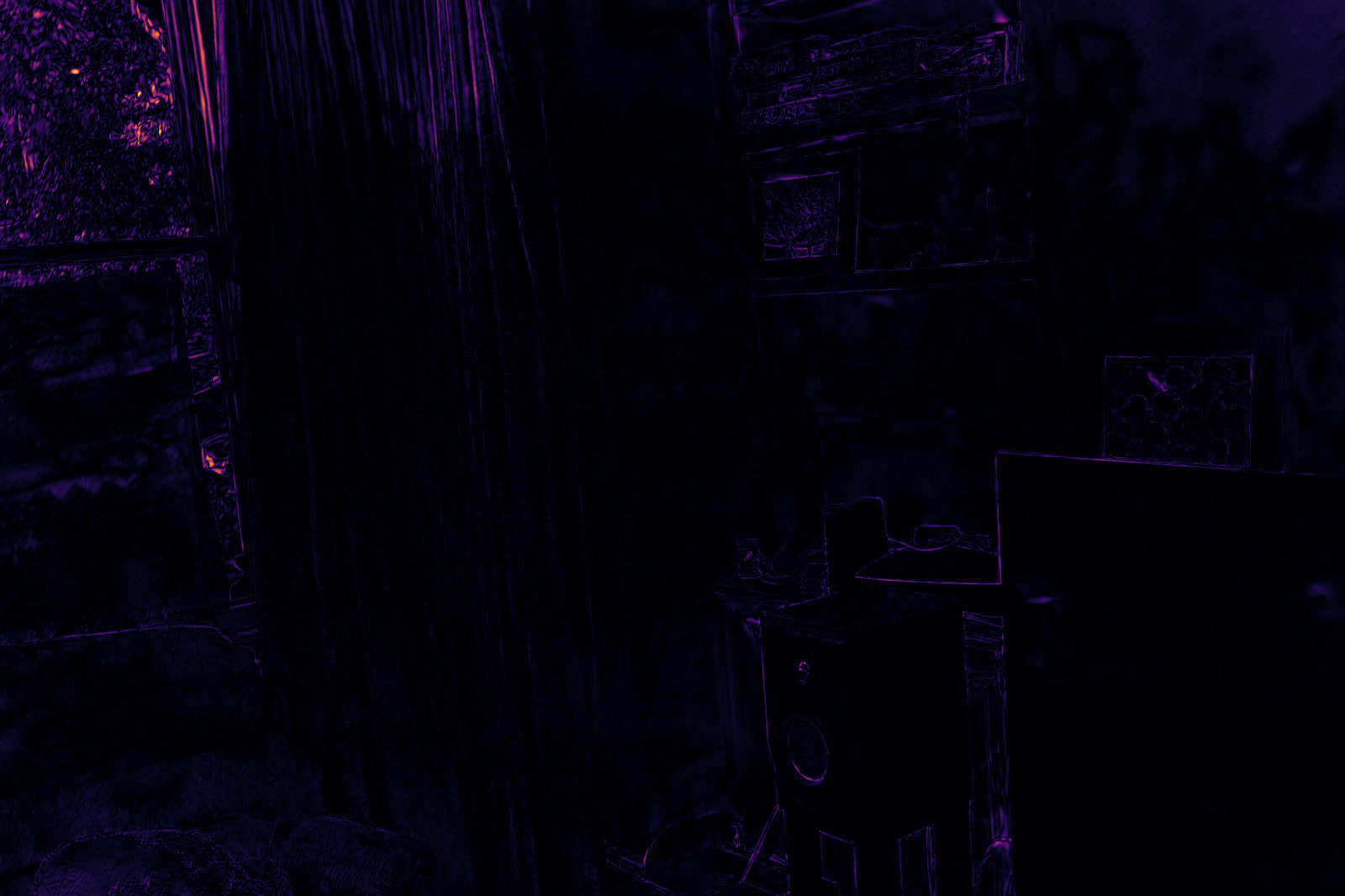}
\end{subfigure}

\begin{subfigure}{.245\textwidth}
      \centering
      \includegraphics[width=\textwidth]{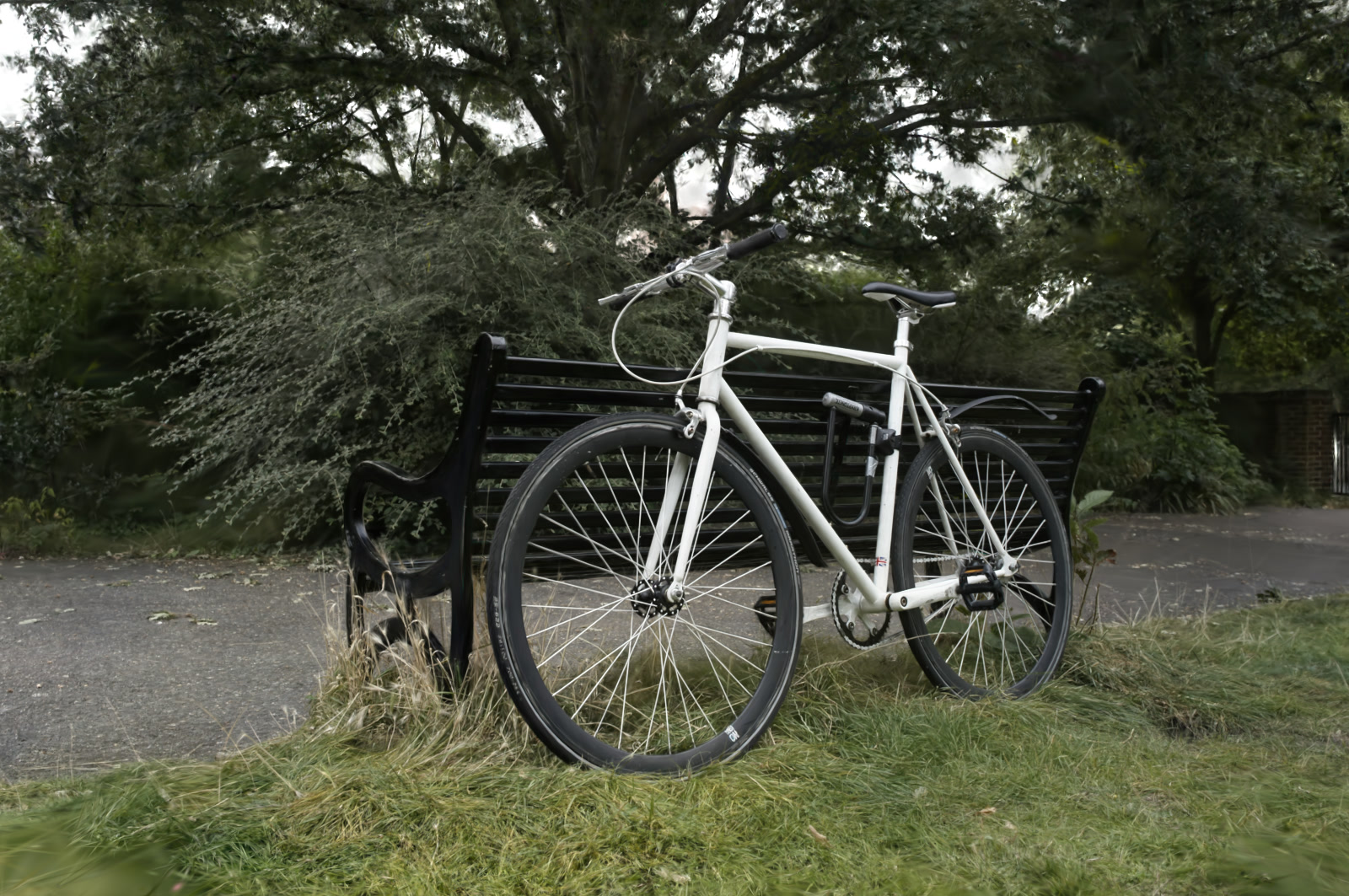}
  \end{subfigure}
\begin{subfigure}{.245\textwidth}
      \centering
      \includegraphics[width=\textwidth]{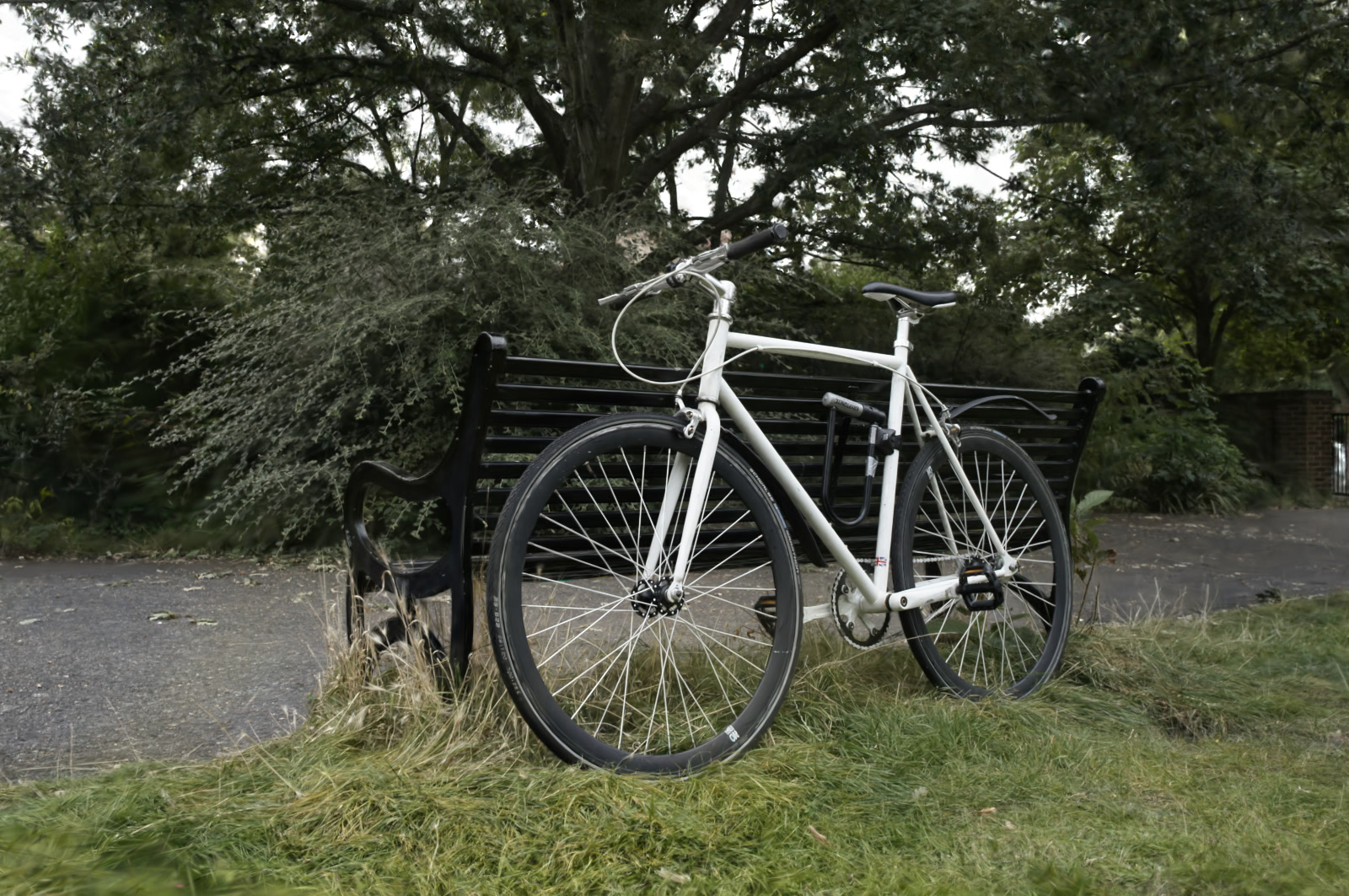}
  \end{subfigure}
\begin{subfigure}{.245\textwidth}
      \centering
      \includegraphics[width=\textwidth]{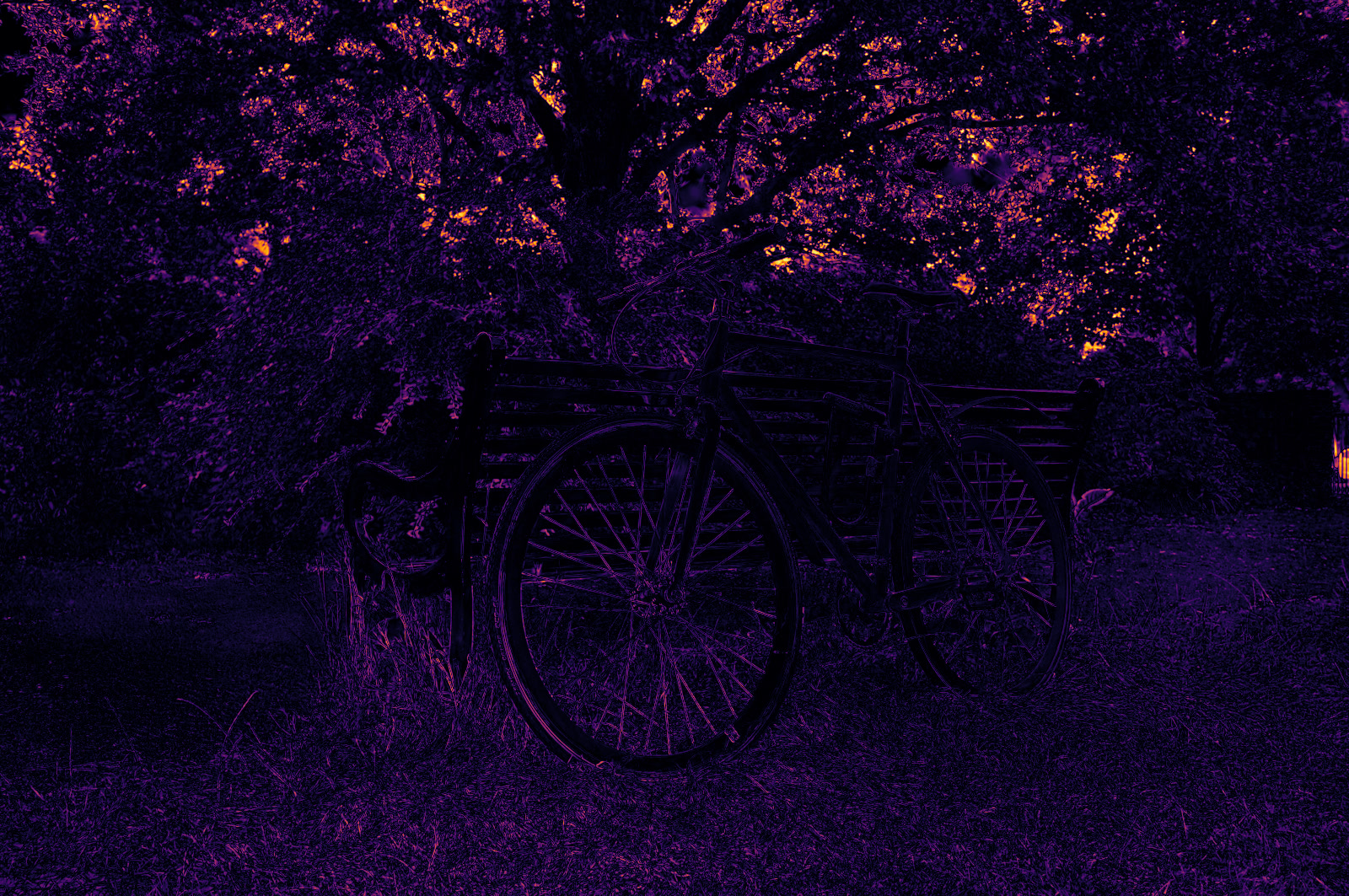}
  \end{subfigure}
\begin{subfigure}{.245\textwidth}
      \centering
      \includegraphics[width=\textwidth]{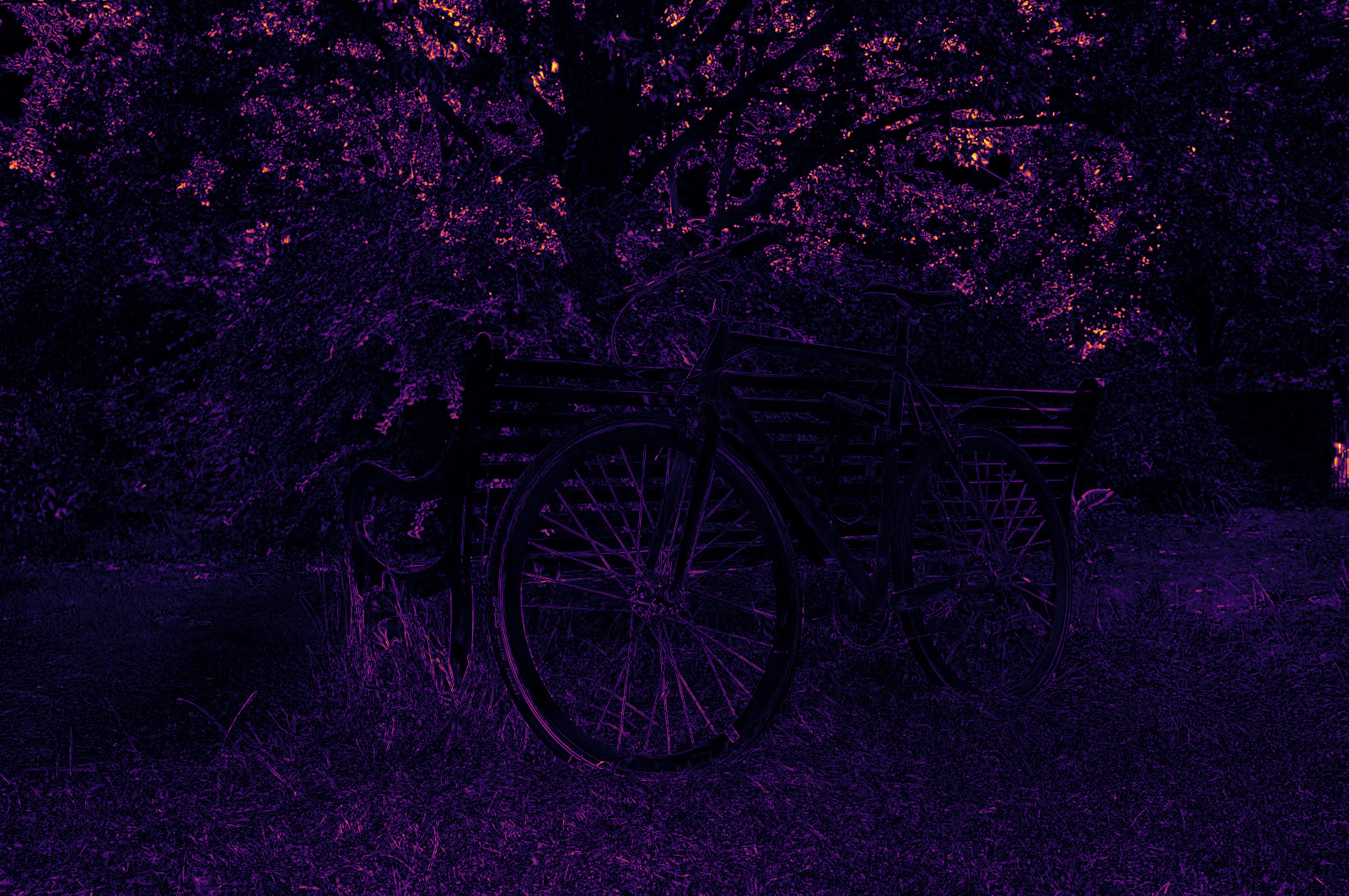}
\end{subfigure}

\begin{subfigure}{.245\textwidth}
      \centering
      \includegraphics[width=\textwidth]{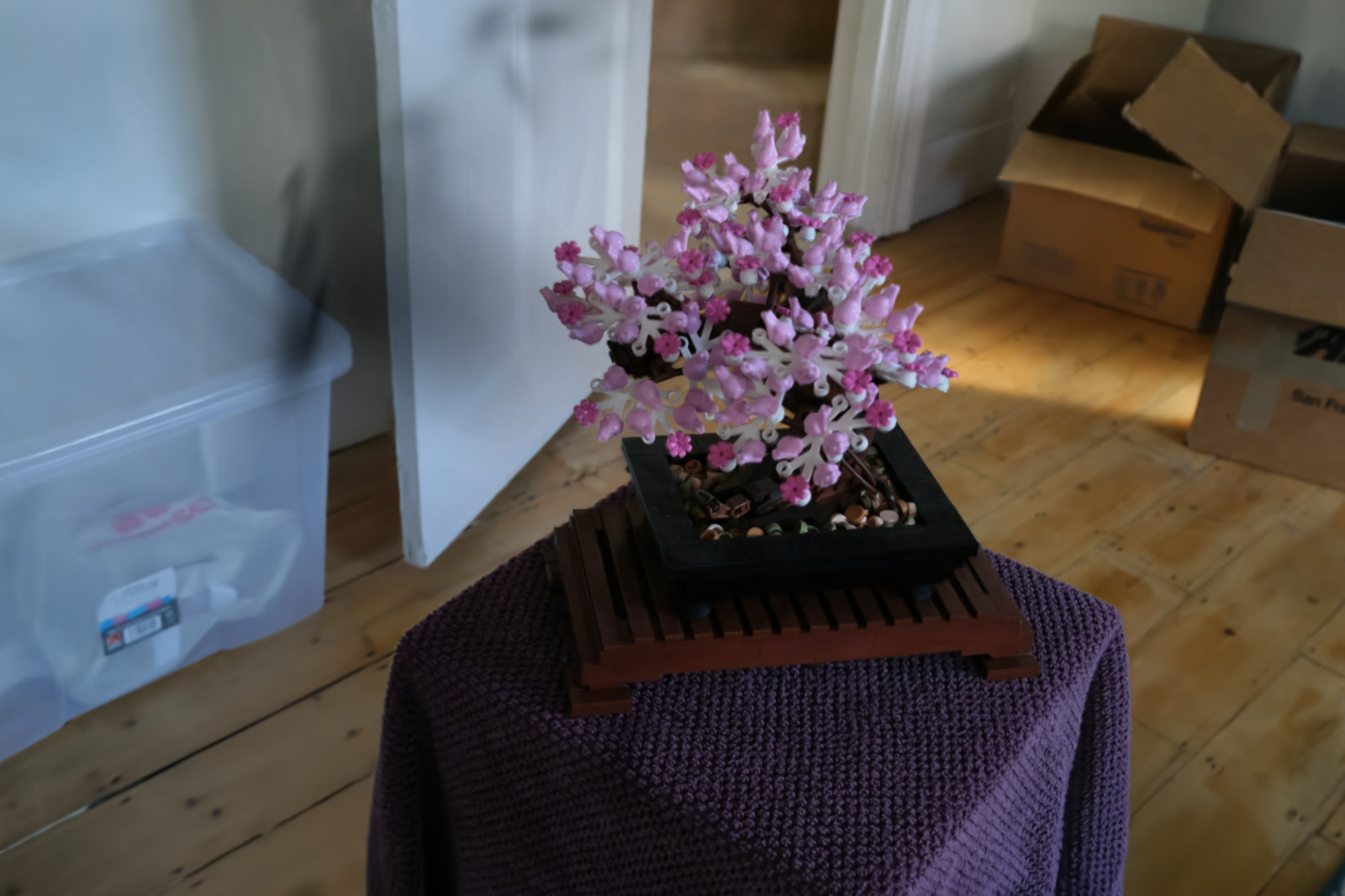}
  \end{subfigure}
\begin{subfigure}{.245\textwidth}
      \centering
      \includegraphics[width=\textwidth]{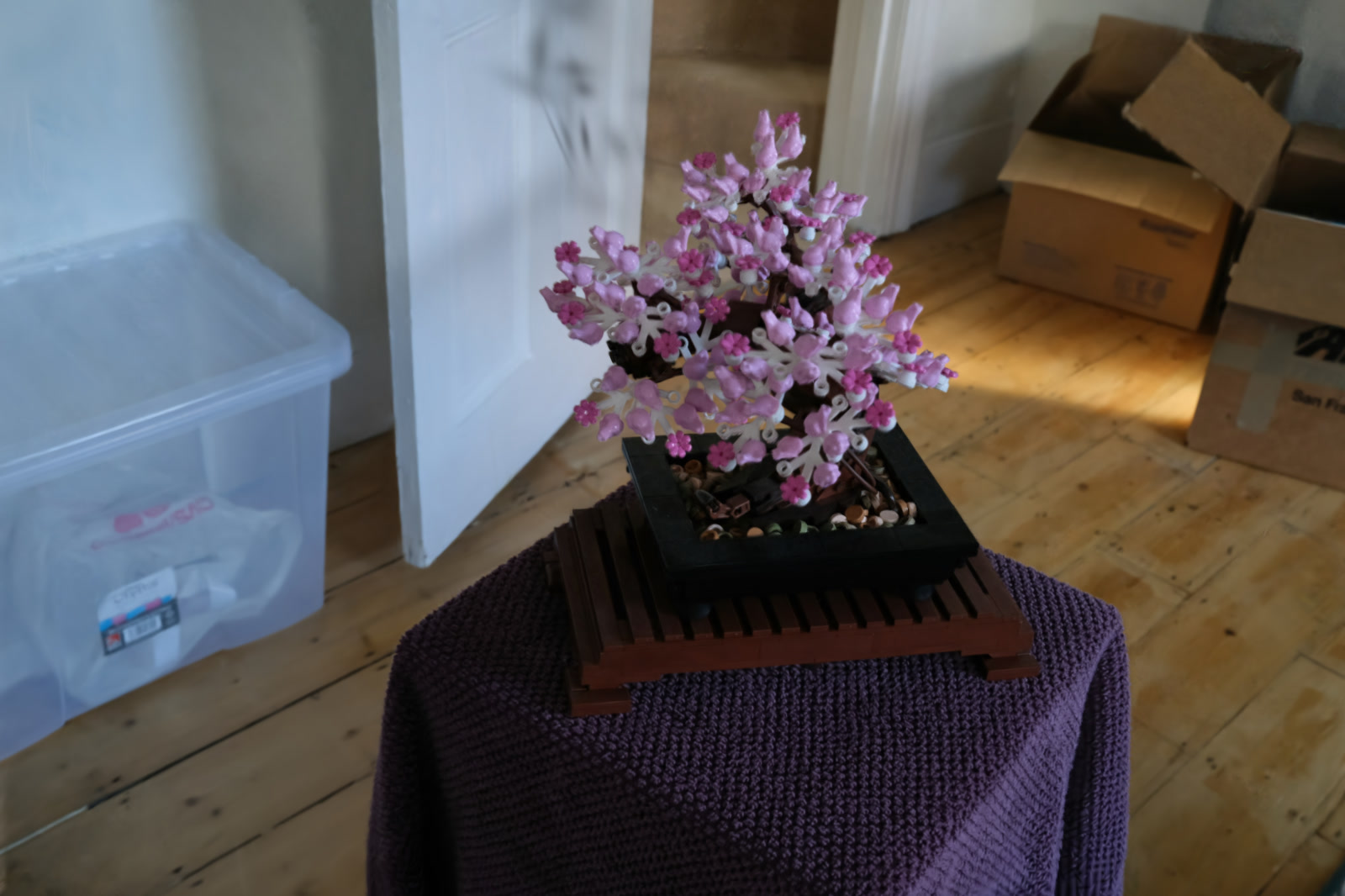}
  \end{subfigure}
\begin{subfigure}{.245\textwidth}
      \centering
      \includegraphics[width=\textwidth]{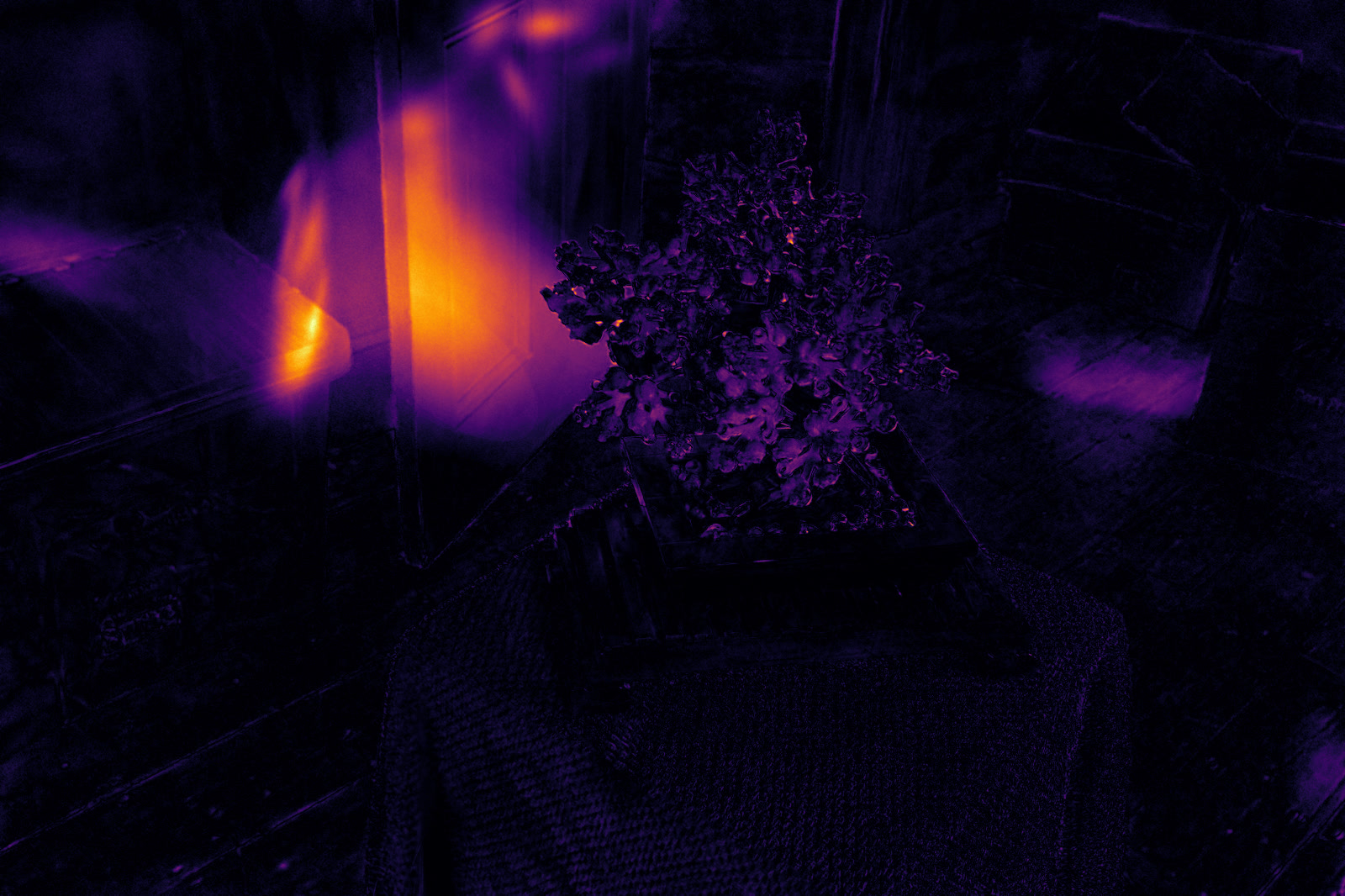}
  \end{subfigure}
\begin{subfigure}{.245\textwidth}
      \centering
      \includegraphics[width=\textwidth]{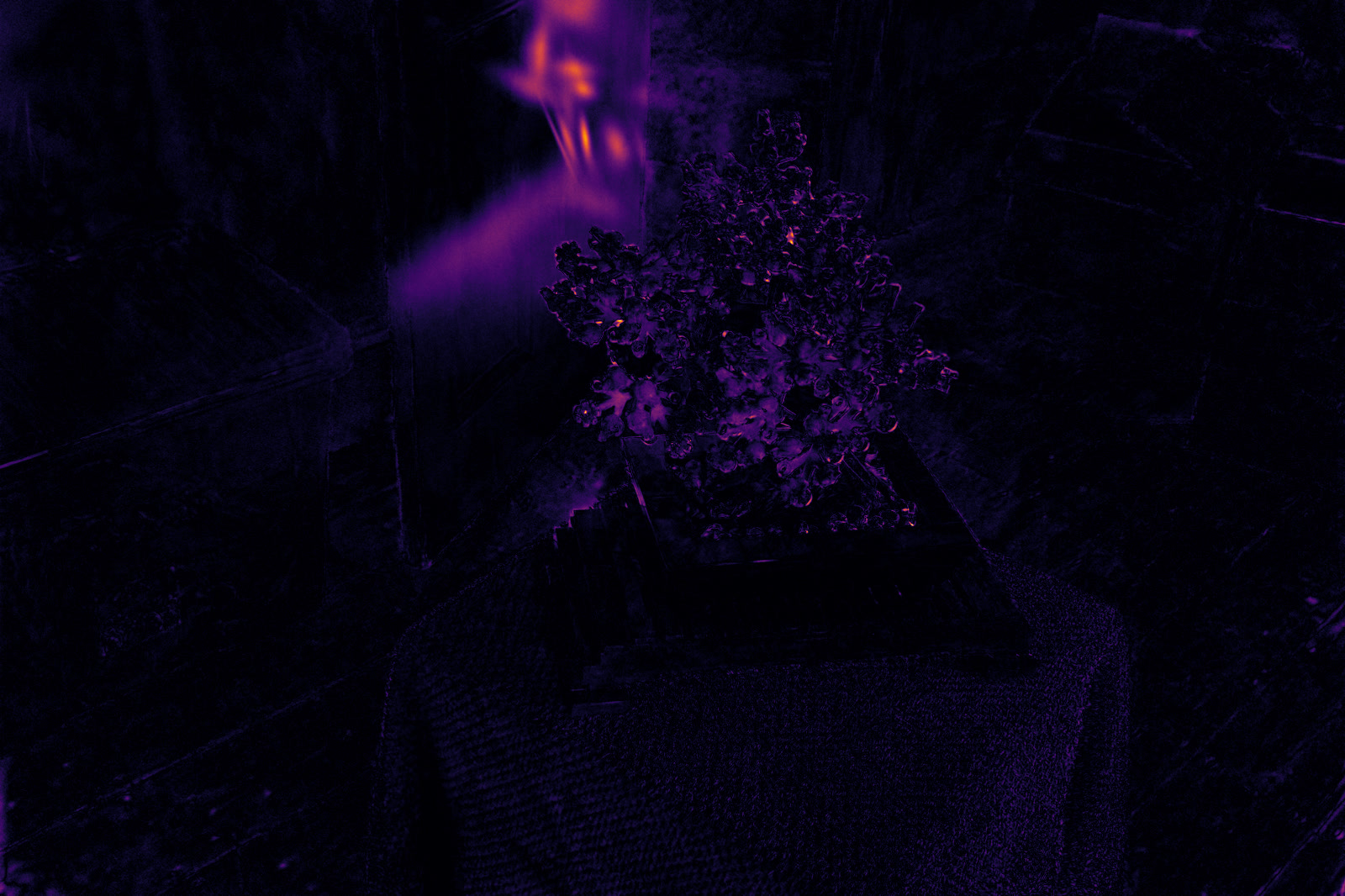}
\end{subfigure}

\begin{subfigure}{.245\textwidth}
      \centering
      \includegraphics[width=\textwidth]{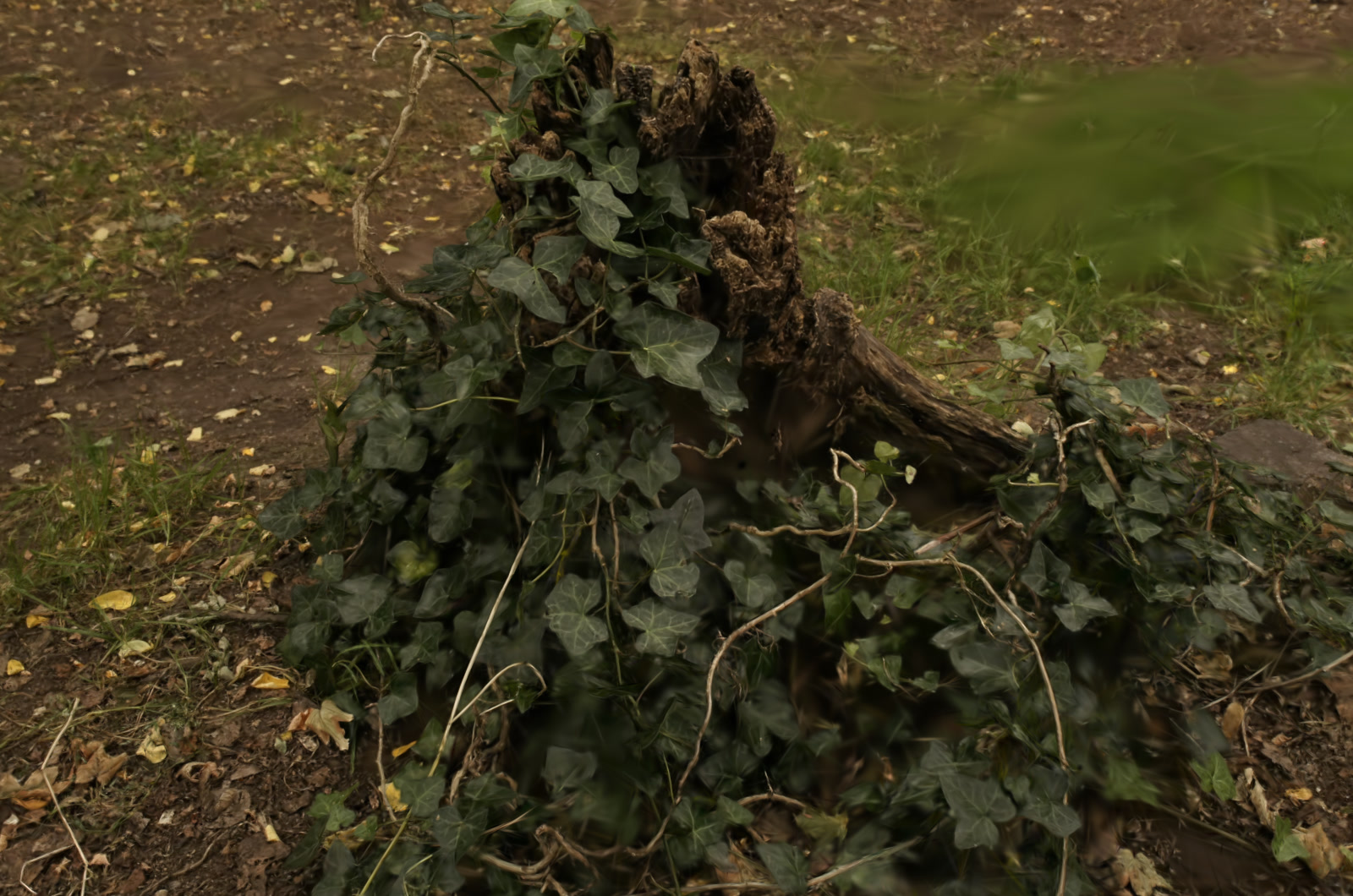}
  \end{subfigure}
\begin{subfigure}{.245\textwidth}
      \centering
      \includegraphics[width=\textwidth]{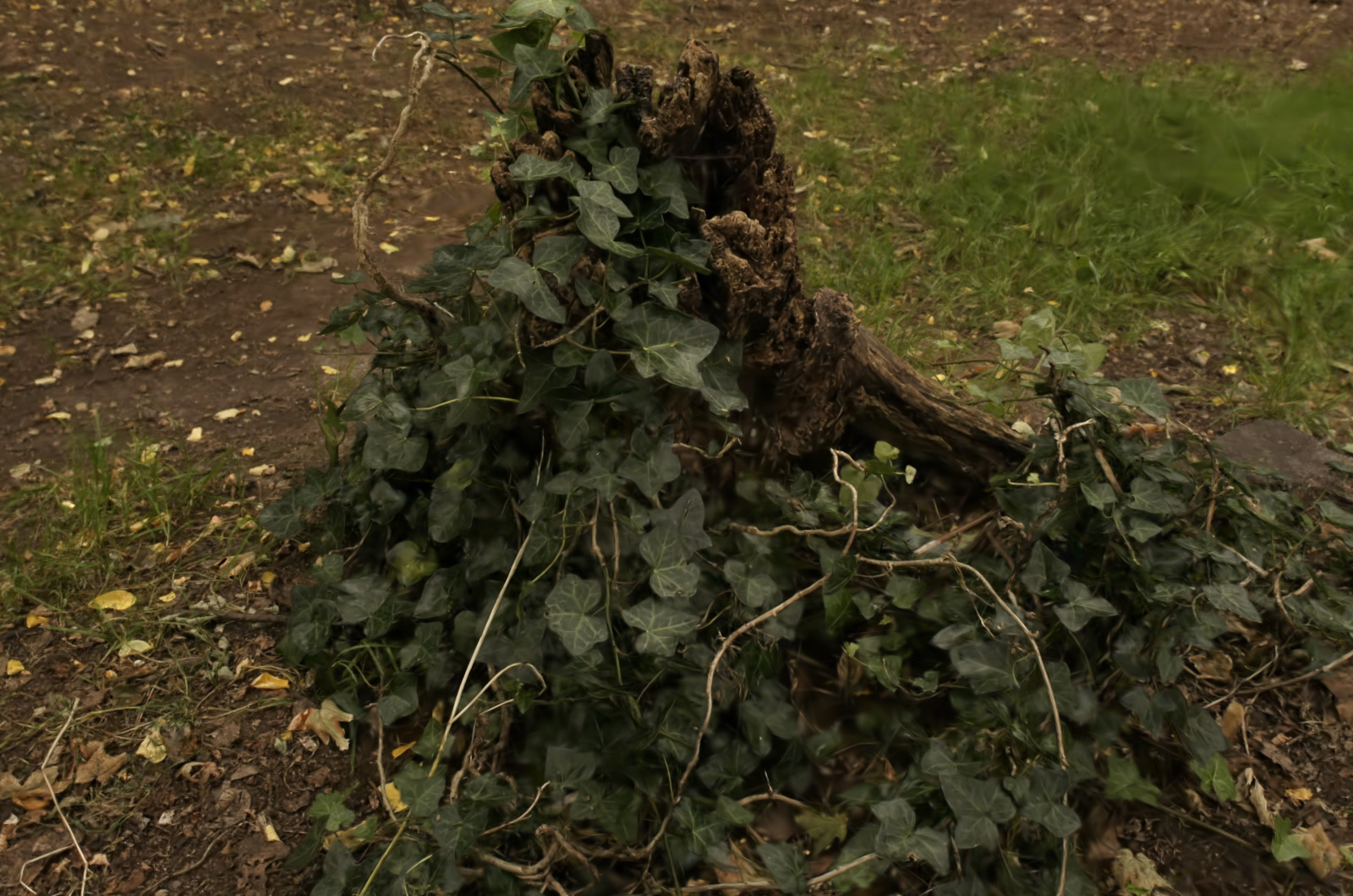}
  \end{subfigure}
\begin{subfigure}{.245\textwidth}
      \centering
      \includegraphics[width=\textwidth]{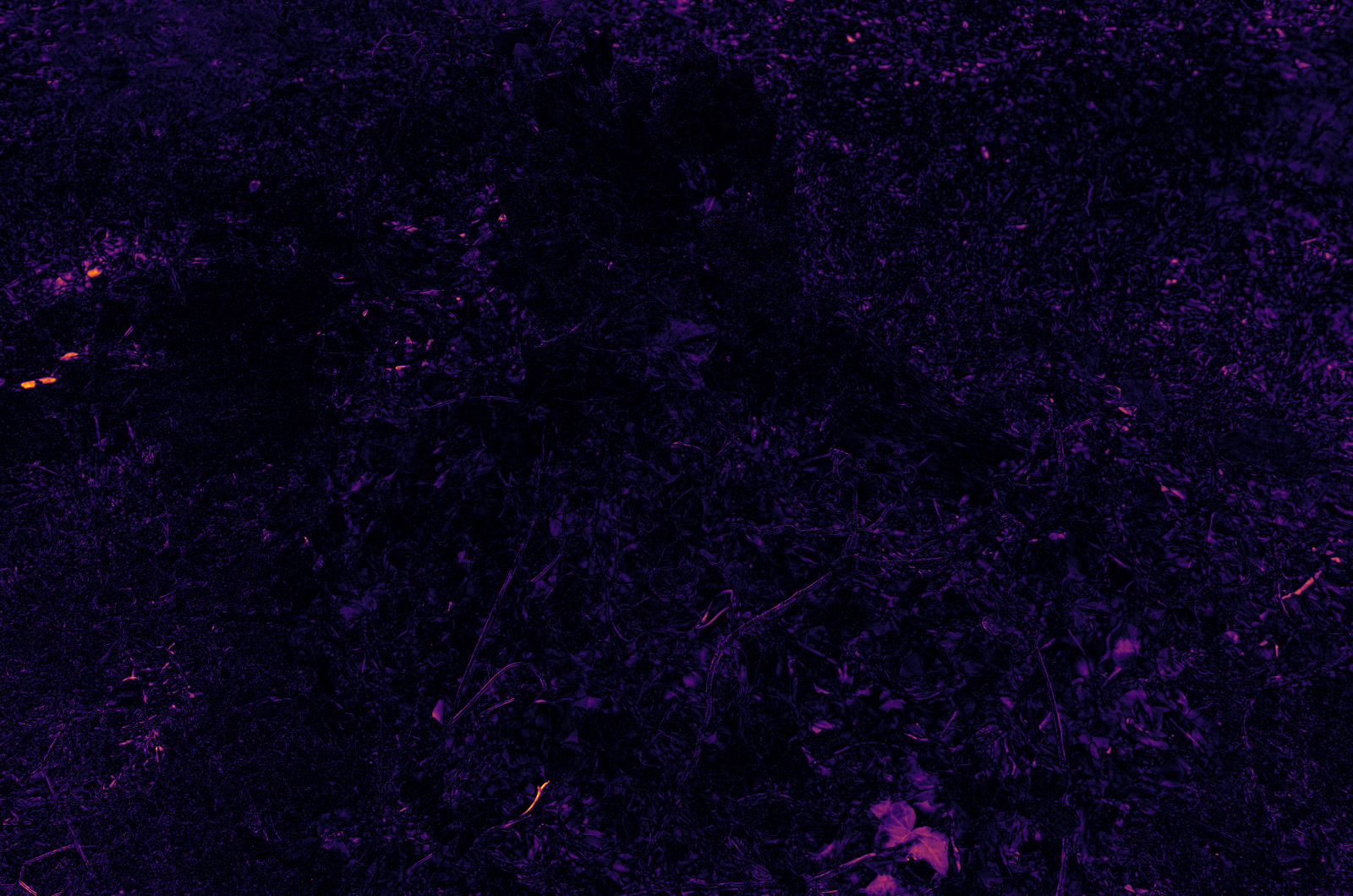}
  \end{subfigure}
\begin{subfigure}{.245\textwidth}
      \centering
      \includegraphics[width=\textwidth]{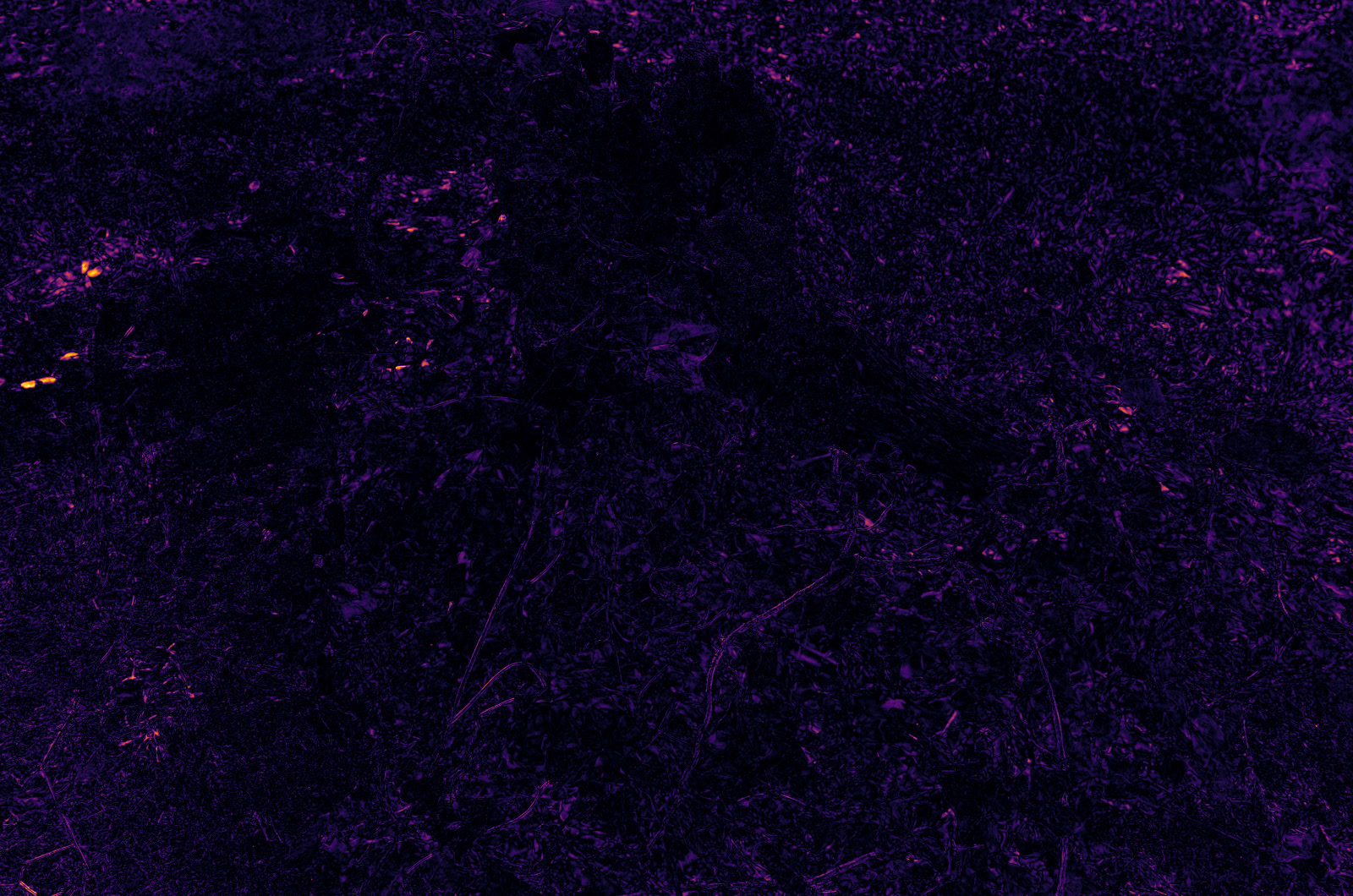}
\end{subfigure}

\begin{subfigure}{.245\textwidth}
      \centering
      \includegraphics[width=\textwidth]{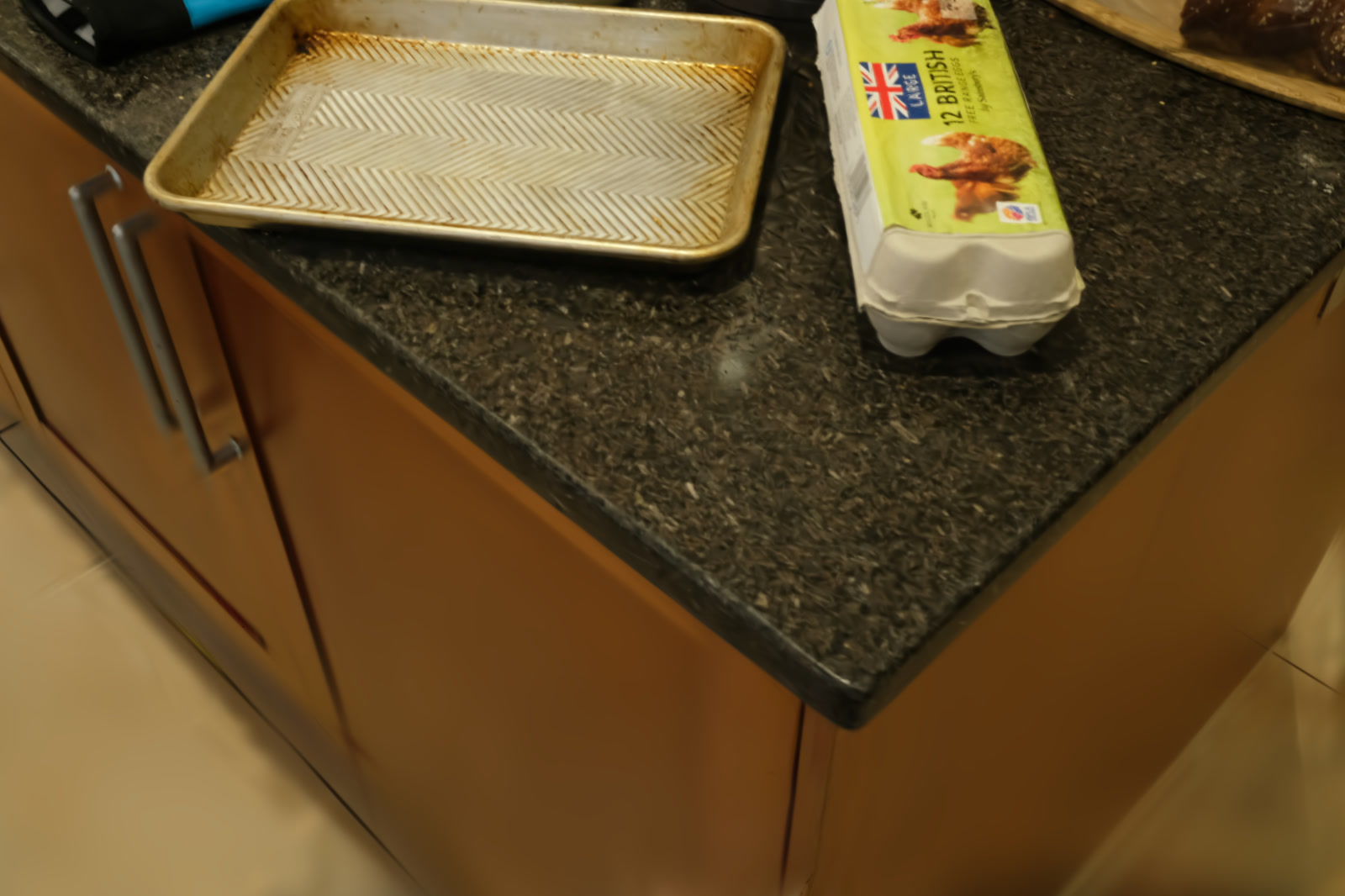}
      \caption*{COLMAP}
  \end{subfigure}
\begin{subfigure}{.245\textwidth}
      \centering
      \includegraphics[width=\textwidth]{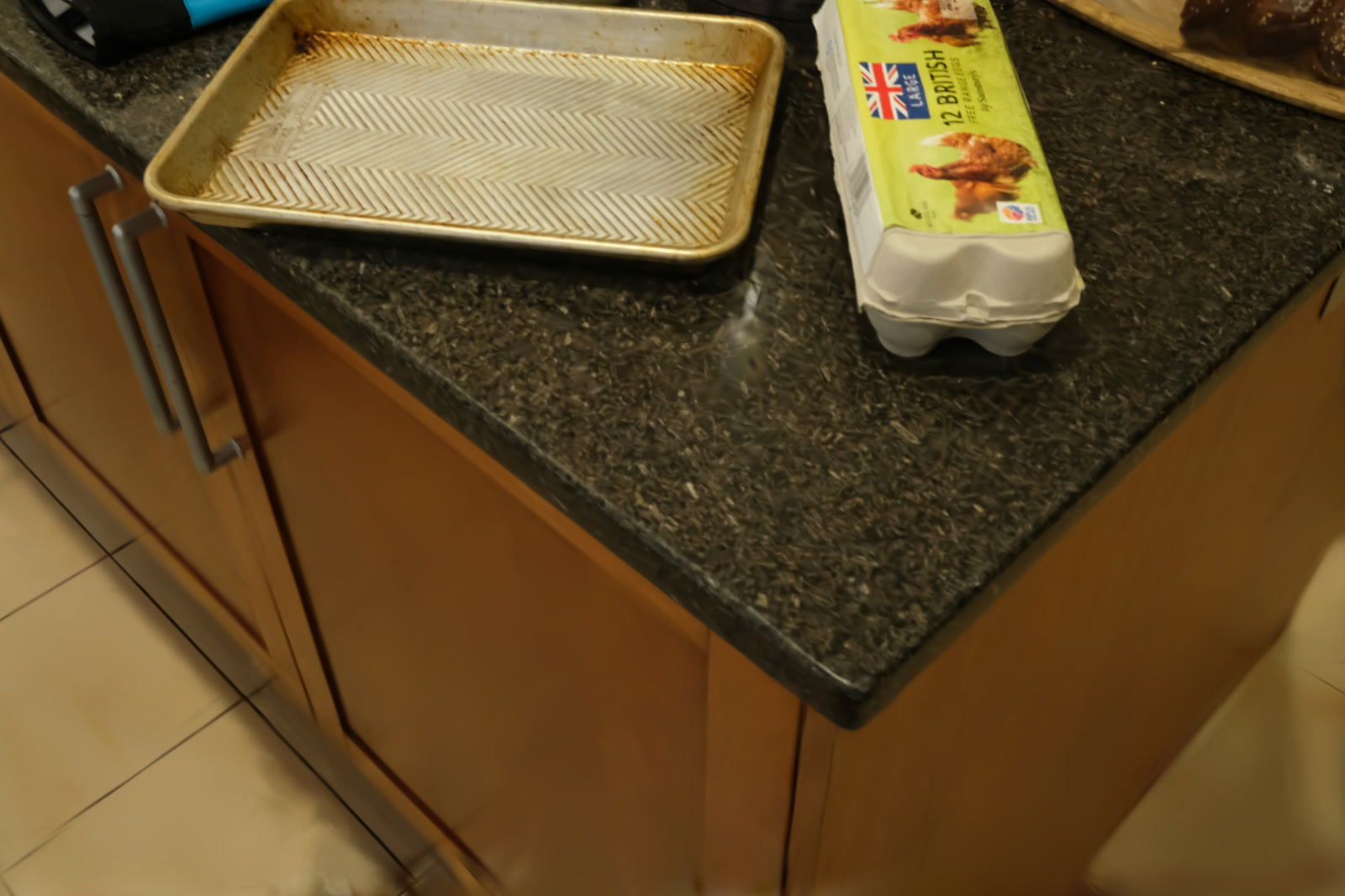}
      \caption*{Ours}
  \end{subfigure}
\begin{subfigure}{.245\textwidth}
      \centering
      \includegraphics[width=\textwidth]{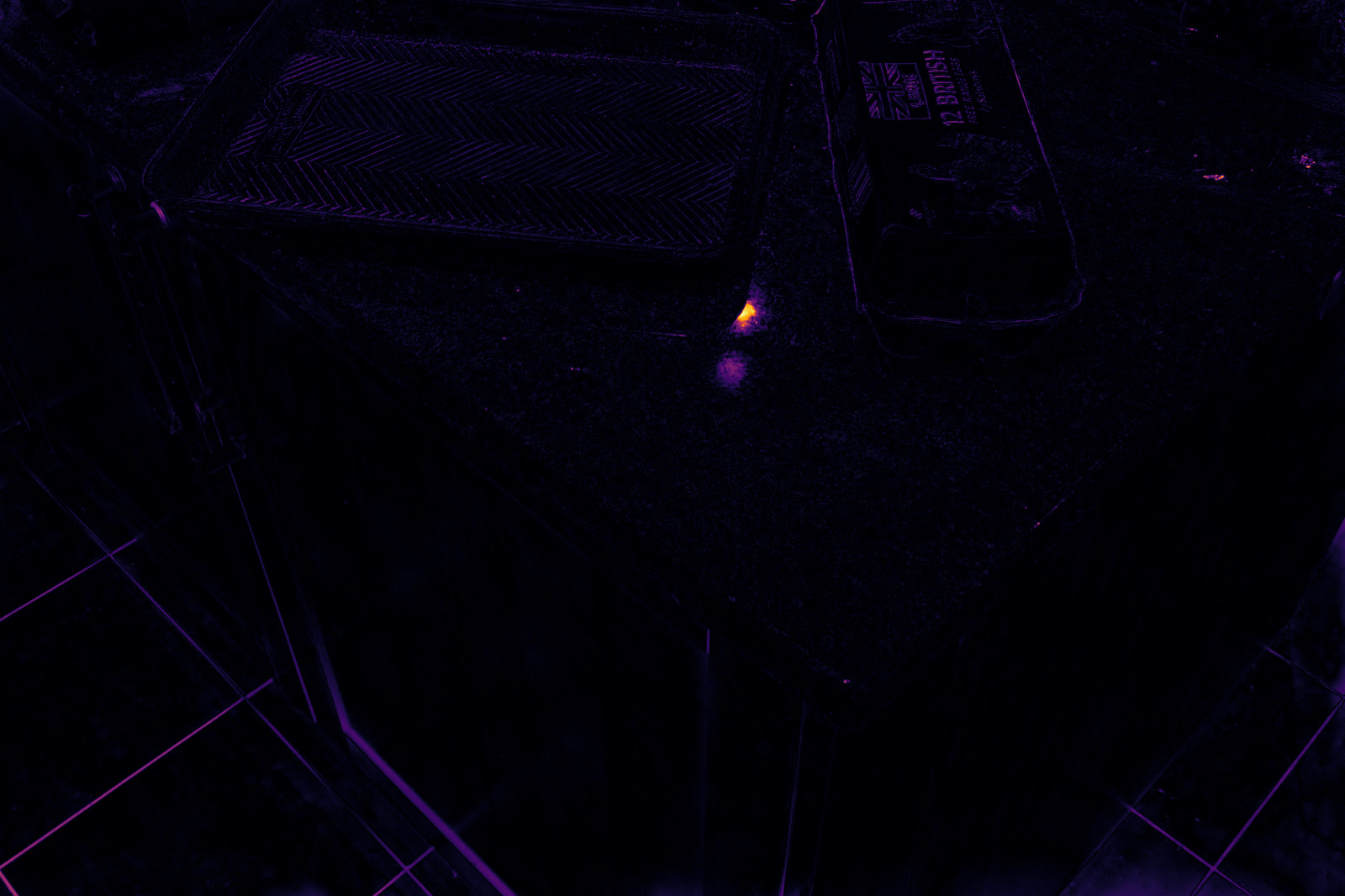}
      \caption*{COLMAP - Error Map}
  \end{subfigure}
\begin{subfigure}{.245\textwidth}
      \centering
      \includegraphics[width=\textwidth]{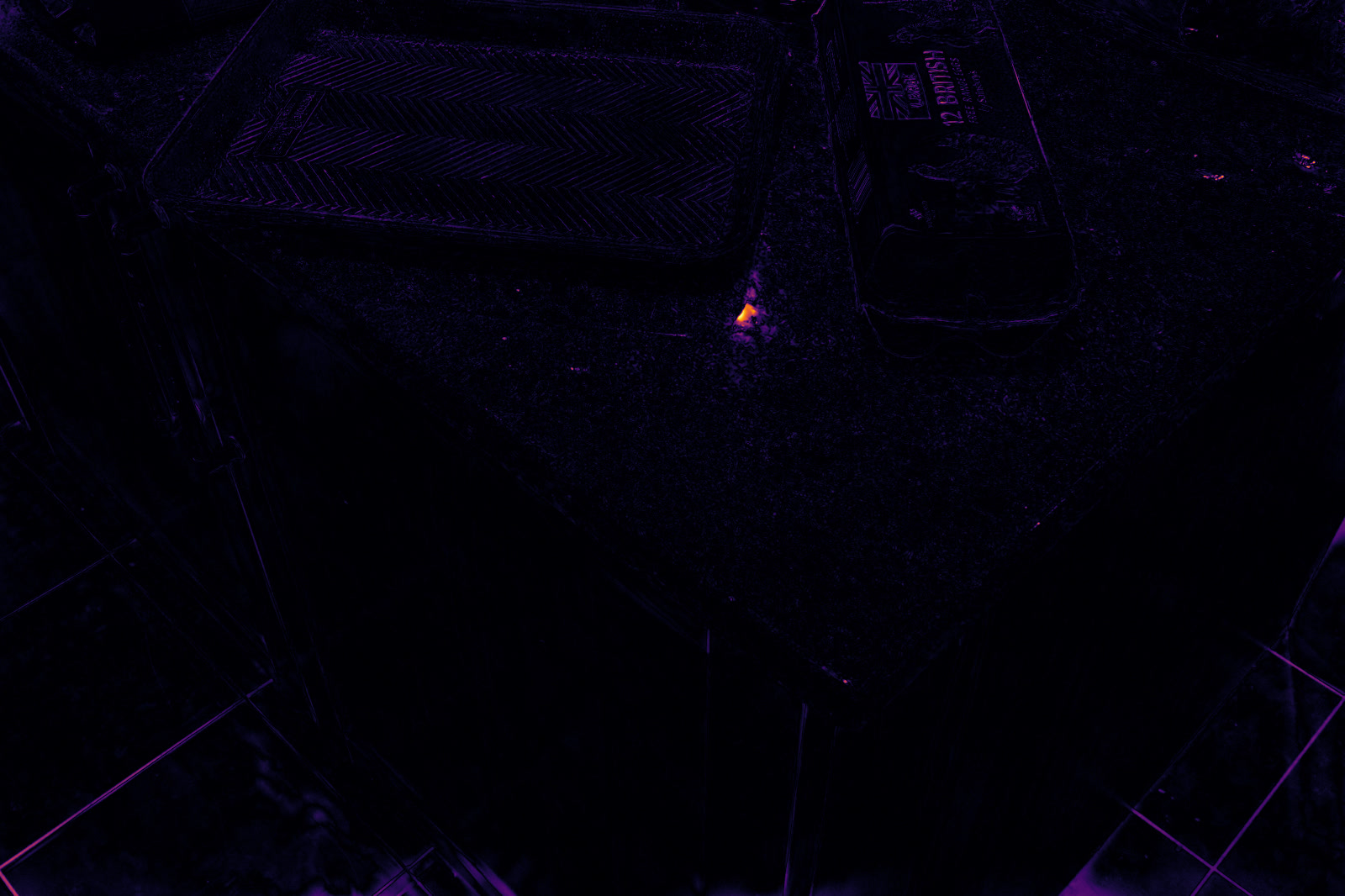}
      \caption*{Ours - Error Map}
\end{subfigure}
  \caption{Additional qualitative results from the OMMO dataset~\cite{lu2023large} and Mip-NeRF 360~\cite{barron2022mip}.
  Here we visualize only the result of training with COLMAP and our best performing model with NeRF initialization + depth supervision, to highlight the total contribution of the strategies we evaluated.
  Please zoom in to see details.}
  \label{fig:error_map}
\end{figure*}

\begin{figure*}[h!]
  \centering
  \begin{subfigure}{.495\textwidth}
      \centering
      \includegraphics[width=\textwidth]{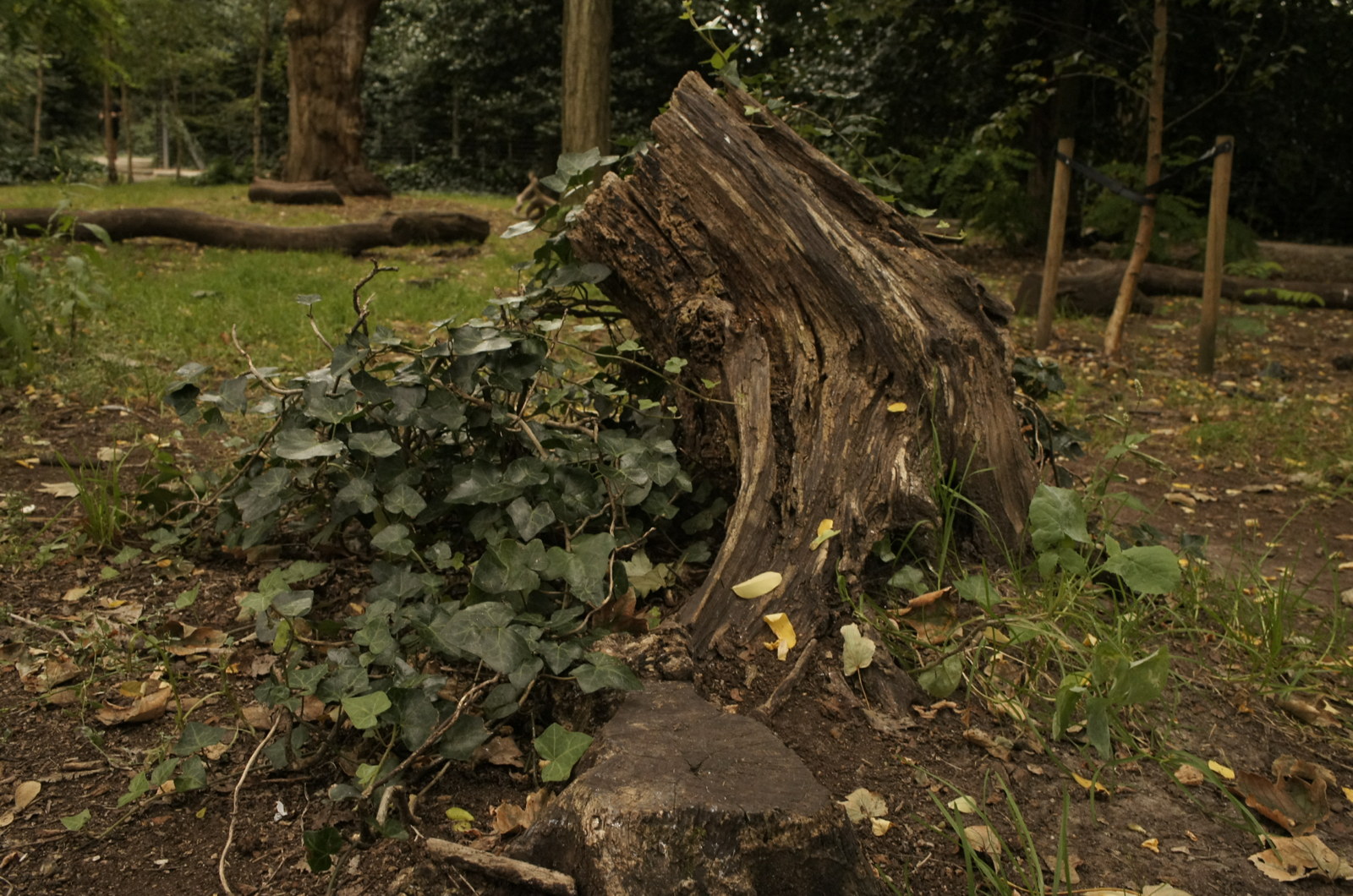}
      \caption*{Ground truth}
  \end{subfigure}
  \begin{subfigure}{.495\textwidth}
      \centering
      \includegraphics[width=\textwidth]{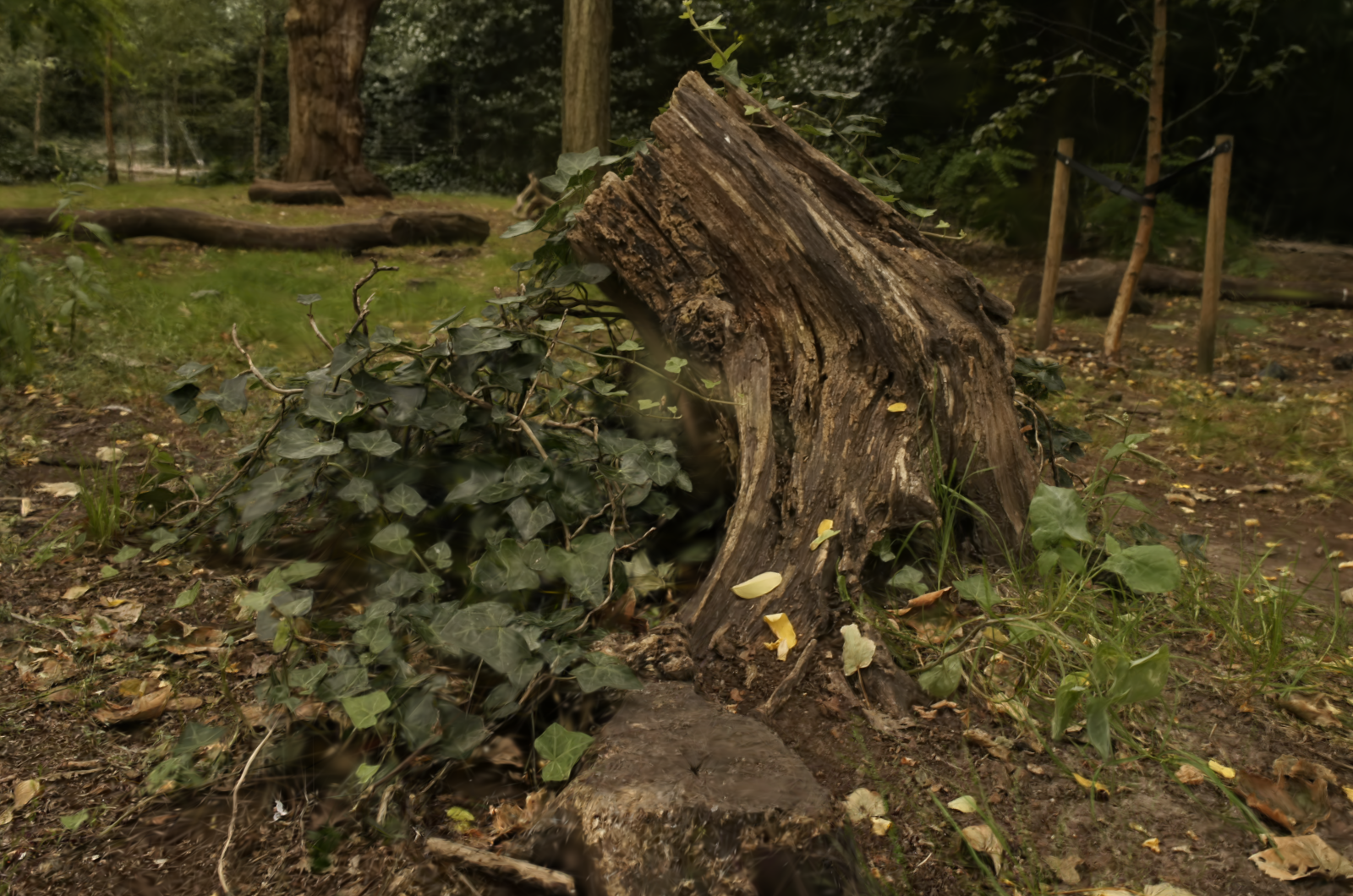}
      \caption*{COLMAP}
  \end{subfigure}
  \begin{subfigure}{.495\textwidth}
      \centering
      \includegraphics[width=\textwidth]{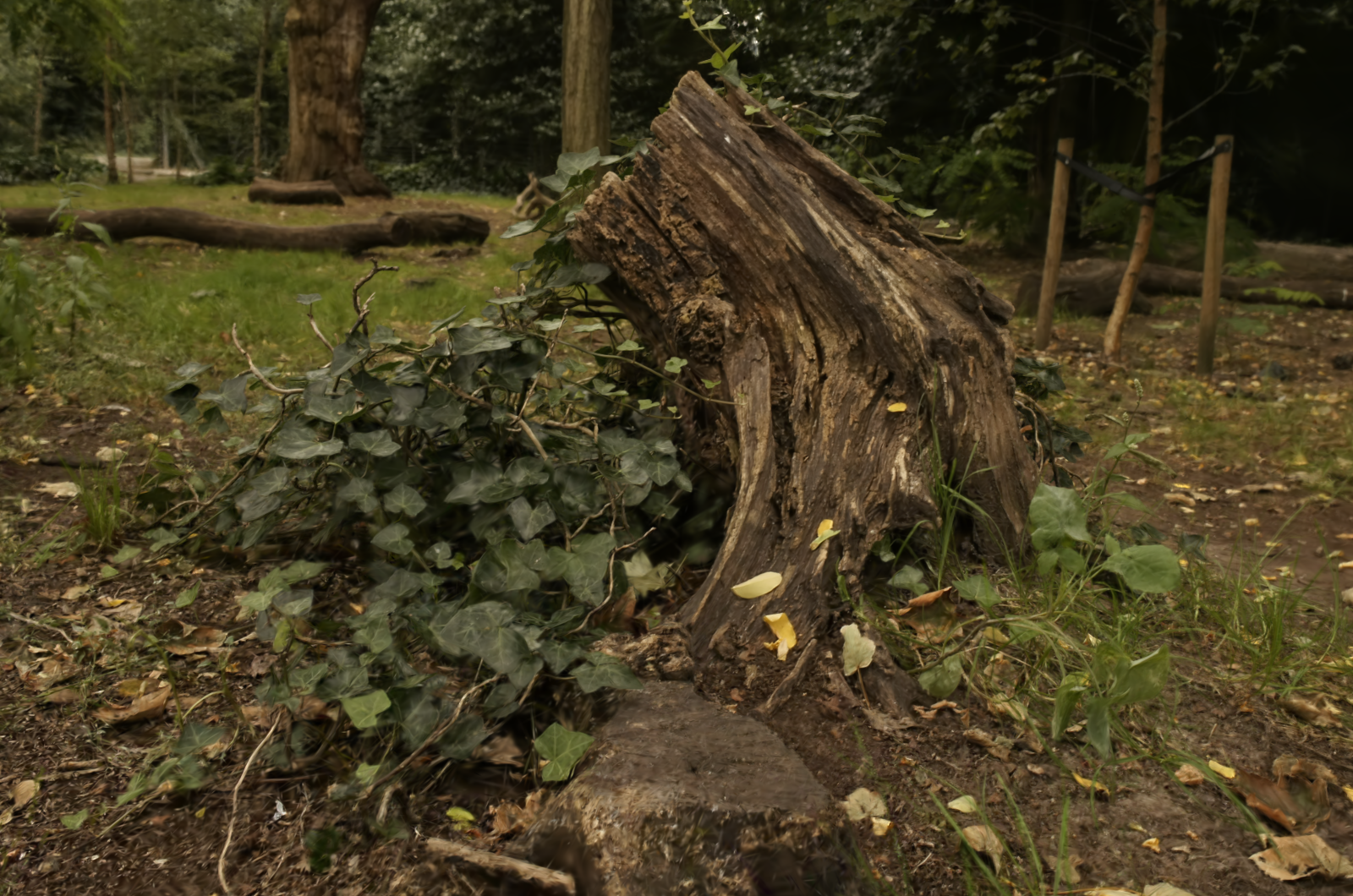}
      \caption*{NerfAcc}
  \end{subfigure}
    \begin{subfigure}{.495\textwidth}
      \centering
      \includegraphics[width=\textwidth]{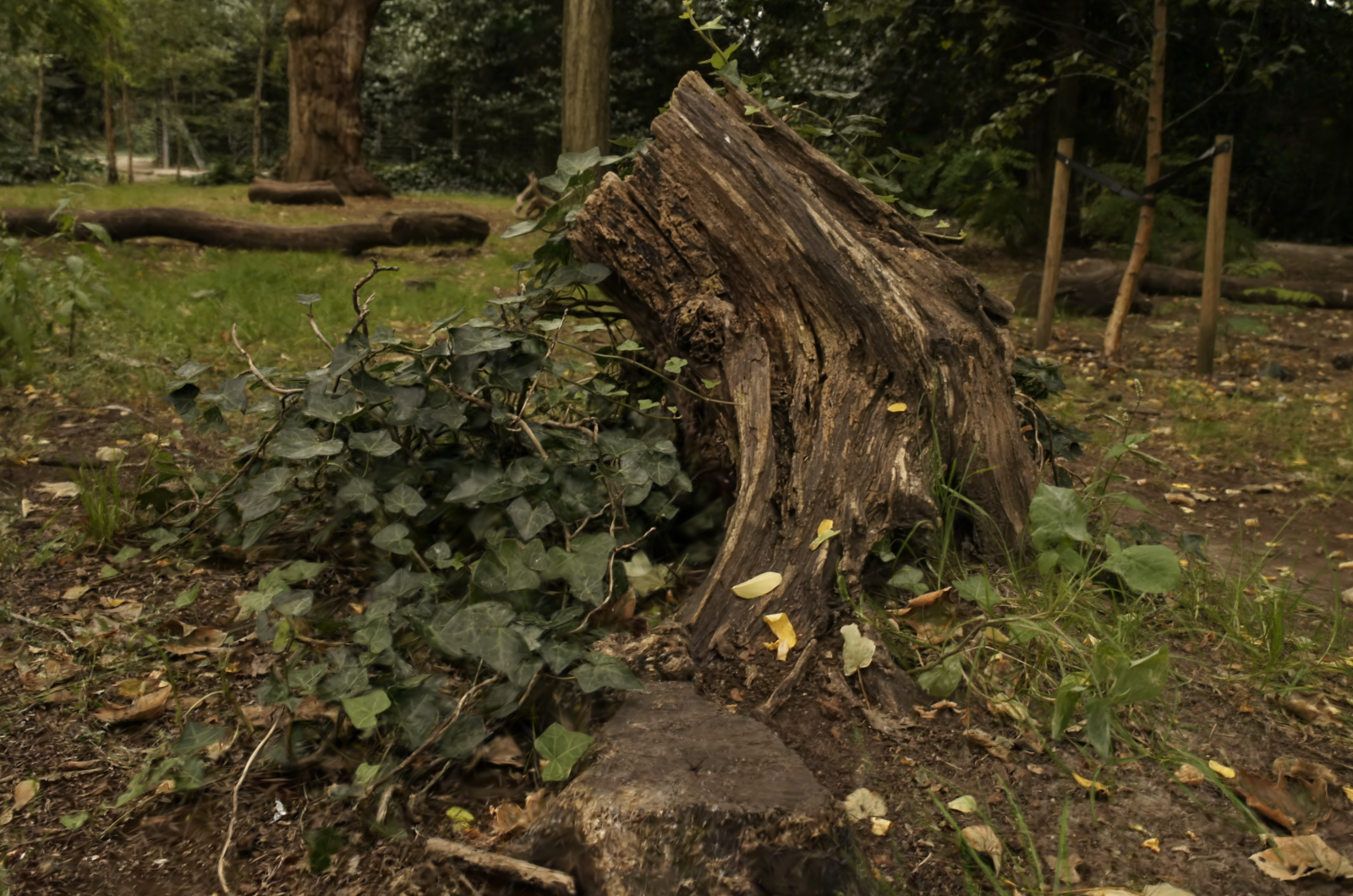}
      \caption*{NerfAcc + depth}
  \end{subfigure}
  \begin{subfigure}{.495\textwidth}
      \centering
      \includegraphics[width=\textwidth]{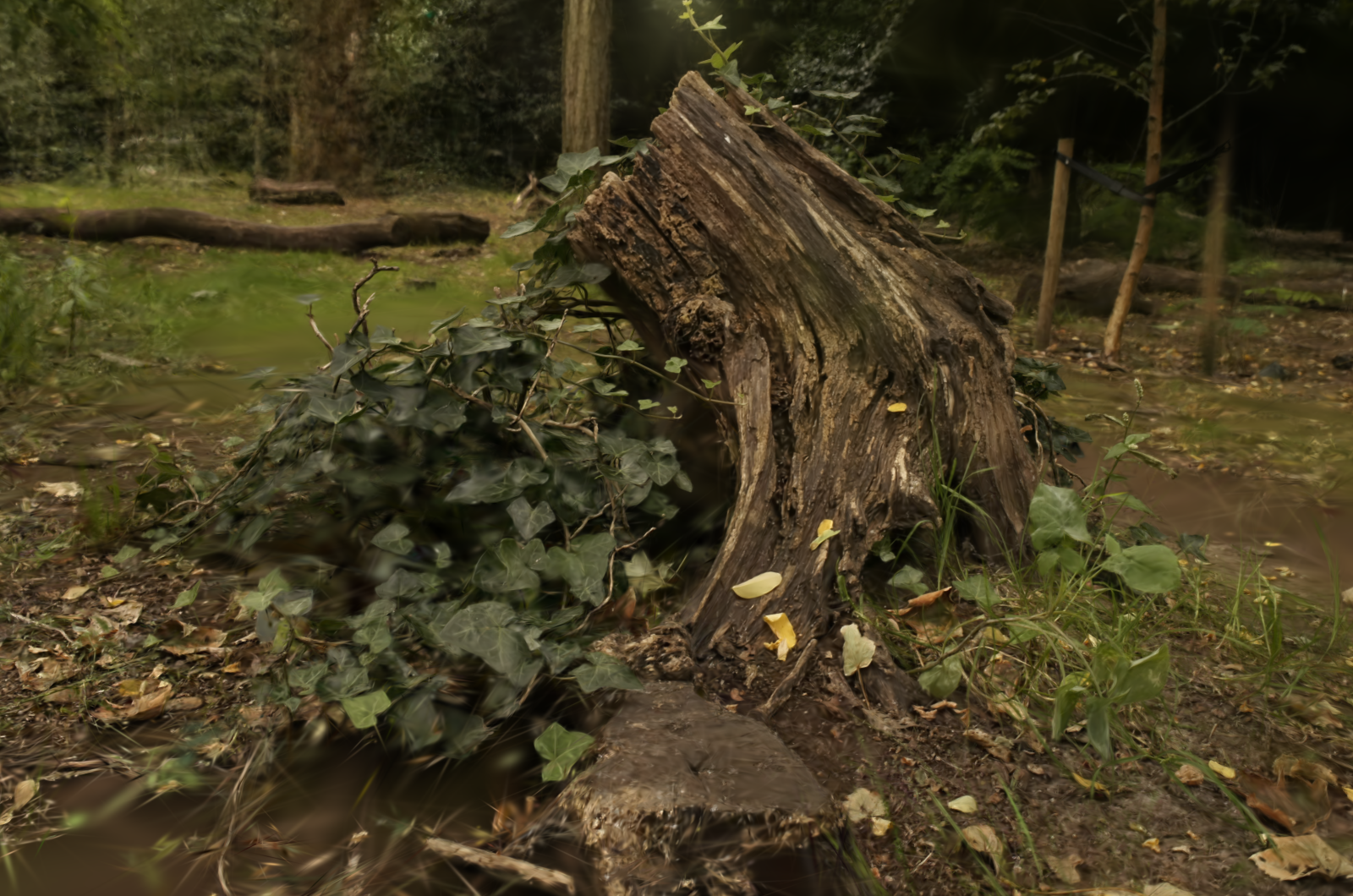}
      \caption*{Random}
  \end{subfigure}
    \begin{subfigure}{.495\textwidth}
      \centering
      \includegraphics[width=\textwidth]{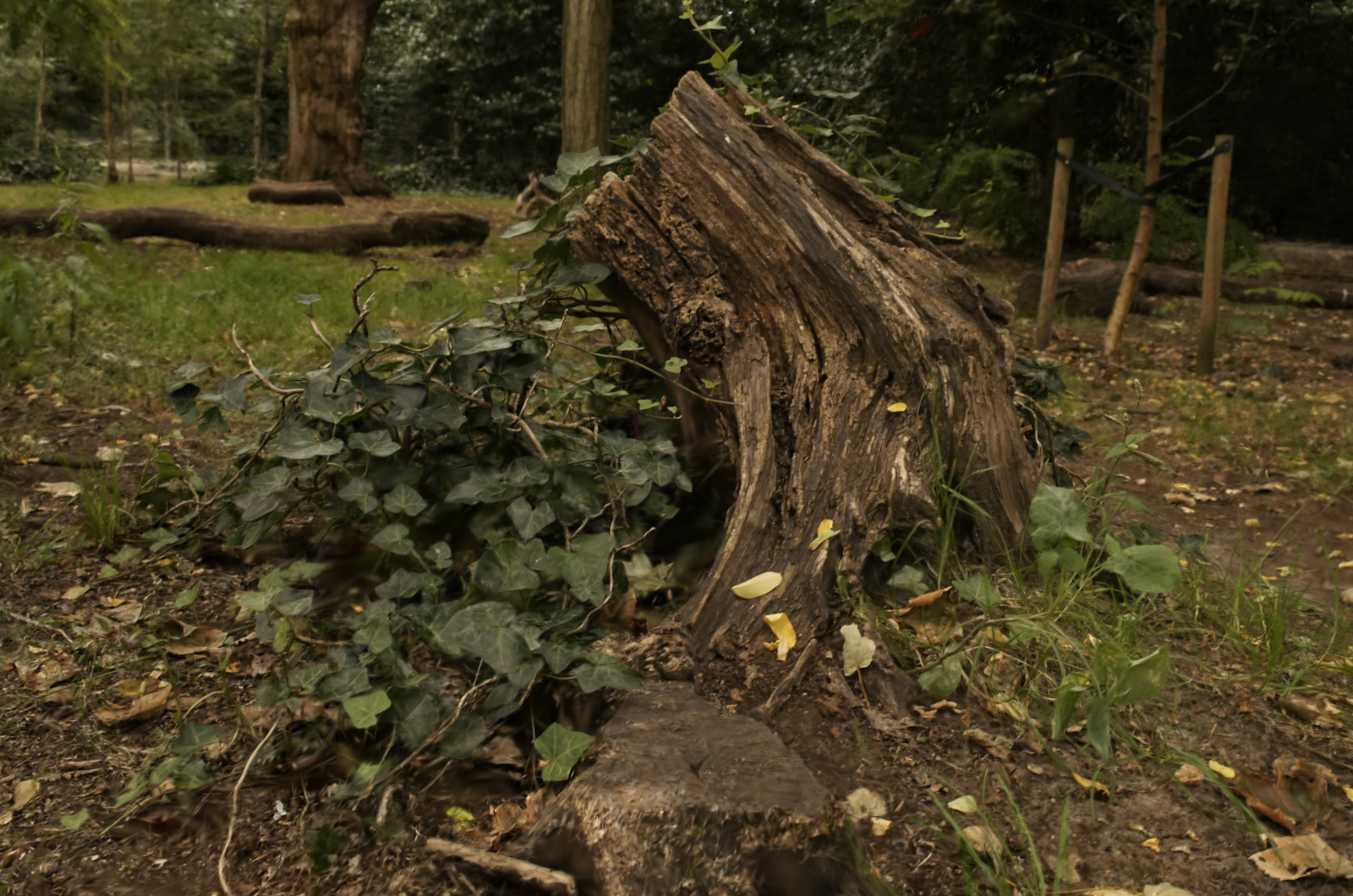}
      \caption*{Random + depth}
  \end{subfigure}
  \caption{Qualitative results on a scene from the Mip-NeRF 360 dataset~\cite{lu2023large}. We observe that the denser initialization from NeRF, and the structure guidance both lead to decreased loss of detail and thin geometry.
  Please zoom in to see details. Please also refer to the interactive comparison in the supplementary material.
  }
  \label{fig:stump}
\end{figure*}

\begin{figure*}[h!]
  \centering
  \begin{subfigure}{.495\textwidth}
      \centering
      \includegraphics[width=\textwidth]{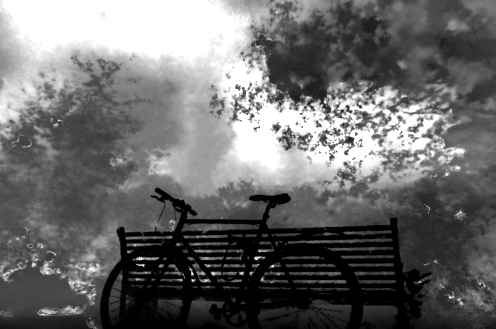}
  \end{subfigure}
  \begin{subfigure}{.495\textwidth}
      \centering
      \includegraphics[width=\textwidth]{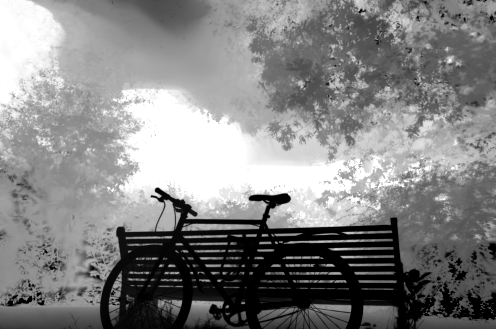}
  \end{subfigure}
  \begin{subfigure}{.495\textwidth}
      \centering
      \includegraphics[width=\textwidth]{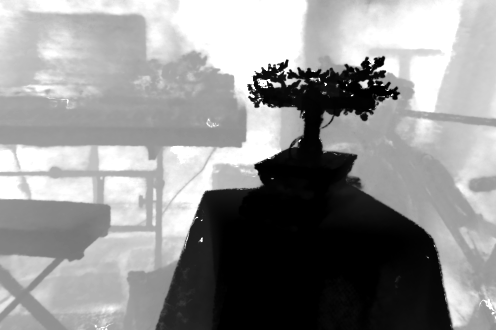}
  \end{subfigure}
    \begin{subfigure}{.495\textwidth}
      \centering
      \includegraphics[width=\textwidth]{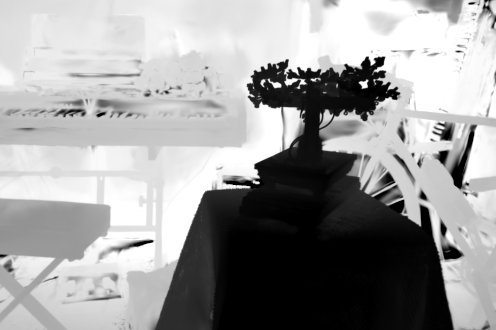}
  \end{subfigure}
  \begin{subfigure}{.495\textwidth}
      \centering
      \includegraphics[width=\textwidth]{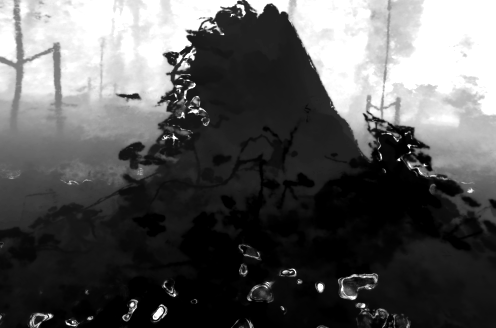}
      \caption*{NerfAcc}
  \end{subfigure}
    \begin{subfigure}{.495\textwidth}
      \centering
      \includegraphics[width=\textwidth]{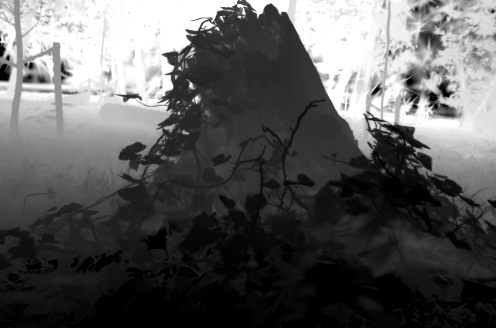}
      \caption*{Gaussian Splatting}
  \end{subfigure}
  \caption{Here we visualize the depth supervision from a NerfAcc model trained for 5000 iterations with the final depth rendered from the trained Gaussian Splatting model.
  The coarse structure of the NeRF depth is well-aligned with the final result, and allows Gaussian Splatting to more quickly refine the fine structures.}
  \label{fig:bicycle_depth}
\end{figure*}

\subsection{Results on additional data}

We repeat our experiments with NeRF-based initialization and depth distillation on two scenes from the Deep Blending dataset~\cite{hedman2018deep} which were used in the original Gaussian Splatting paper.
These results are shown in Table~\ref{tab:depth_loss_db}, and also show that better initialization and depth distillation improve over the results from COLMAP initialization.

We also attempted to train Gaussian Splatting models for the ``truck'' and ``train'' scenes from the Tanks and Temples dataset~\cite{knapitsch2017tanks}, but were unable to find an unbounded INGP model which trained well enough for them to provide structure guidance.
We did however observe that a better random initialization also outperformed the originally reported Gaussian Splatting results for these scenes, as we found with other datasets.

\subsection{Additional metrics}

We additionally report the perceptual quality metrics SSIM and LPIPS~\cite{zhang2018perceptual} for the Mip-NeRF 360~\cite{barron2022mip} and OMMO~\cite{lu2023large} datasets.
These results can be found in Tables~\ref{tab:ssim_360}, \ref{tab:ssim_ommo}, \ref{tab:lpips_360}, and \ref{tab:lpips_ommo}.

\subsection{Depth rendering with Gaussians}
To determine the depth of a Gaussian along a ray, we begin with the 3D Gaussian equation:
\begin{equation}
    G(x) = \textrm{exp}\left( -\tfrac{1}{2}x^T \Sigma^{-1} x \right),
\end{equation}
where, without loss of generality, we neglect the mean $\mu$ by assuming our coordinate system to be centered at $\mu$.
We also ignore any normalization or weighting of the Gaussian, which will not impact its depth.

We begin the derivation by substituting $o + td$ for $x$, where $o$ and $d$ are the origin and direction of the ray, and t is the depth:
\begin{equation}
    G(x) = \textrm{exp}\left( -\tfrac{1}{2}(o +td)^T \Sigma^{-1}(o +td) \right).
\end{equation}
By expanding the polynomial, we can rewrite this as:
\begin{equation}
    G(x) = \textrm{exp}\left( -\tfrac{1}{2}(o^T \Sigma^{-1} o + 2t d^T\Sigma^{-1} o + t^2 d^T \Sigma^{-1} d) \right).
\end{equation}
To find the depth with the maximum Gaussian value along the ray, we can factor out the scale of the polynomial solve the argmin:
\begin{align}
    d = \argmin_t &\frac{o^T \Sigma^{-1} o}{d^T \Sigma^{-1} d} + 2t \frac{d^T\Sigma^{-1} o}{d^T \Sigma^{-1} d} + t^2, \\
    d = &-\frac{d^T\Sigma^{-1} o}{d^T \Sigma^{-1} d}.
\end{align}

\subsection{Scheduling of depth loss weight}
We use the following schedule for the loss weight $\lambda$:
\begin{align}
    \lambda (i) = \lambda_{init} \times d ^ \frac{i}{S},
\label{eq:weight}
\end{align}
where $\lambda_{init}$ denotes initial weight, $d$ and $S$ represent decay rate and step, respectively and $i$ stands for iteration.

\begin{table*}
    \caption{%
    \textbf{Results on the Deep Blending dataset~\cite{hedman2018deep}.}
    We report metrics (PSNR / SSIM / LPIPS) on the two Deep Blending scenes reported in Gaussian Splatting.
    }%
    \centering
    \resizebox{\linewidth}{!}{
    \setlength{\tabcolsep}{4pt}
    \begin{tabular}{@{}lccR} \toprule
& \multicolumn{2}{c}{Scene} & \\
Initialization & DrJohnson & Playroom & Average \\ 
\midrule
    COLMAP (From~\cite{kerbl3Dgaussians}) & 28.77 / \textbf{0.8990} / 0.2440 & 30.04 / \textbf{0.9060} / 0.2410 & 29.41 / \textbf{0.9025} / 0.2425 \\
COLMAP (Re-run) & 29.03 / 0.8900 / 0.1763 & \cellcolor{color3}30.10 / 0.9000 / 0.1587 & 29.56 / 0.8950 / 0.1675 \\
\midrule
NerfAcc @ 5k & \cellcolor{color2}29.53 / 0.8952 / \textbf{0.1684} & \cellcolor{color1}30.66 / 0.9027 / \textbf{0.1524} & \cellcolor{color1}30.09	/ 0.8990 / \textbf{0.1604} \\
\midrule
$50^3$ Random + Depth Loss & \cellcolor{color3}29.38 / 0.8904 / 0.1816 & 30.06 / 0.8937 / 0.1615 & \cellcolor{color3}29.72 / 0.8921 / 0.1716 \\
NerfAcc @ 5k + Depth Loss & \cellcolor{color1}29.65 / 0.8965 / 0.1691 & \cellcolor{color2}30.33 / 0.8946 / 0.1569 & \cellcolor{color2}29.99 / 0.8956 / 0.1630 \\
    \bottomrule
    \end{tabular}
    }
    \label{tab:depth_loss_db}
\end{table*}

\begin{table*}
    \caption{%
    \textbf{SSIM on the Mip-NeRF 360 dataset~\cite{barron2022mip}.}
    We report SSIM metric on the Mip-NeRF 360 scenes reported in Gaussian Splatting.}
    \centering
    \resizebox{\linewidth}{!}{
    \setlength{\tabcolsep}{4pt}
    \begin{tabular}{@{}lcccccccR} \toprule
& \multicolumn{8}{c}{Scene} \\
Initialization & Garden & Bicycle & Stump & Counter & Kitchen & Bonsai & Room & Average \\ 
\midrule
COLMAP (From~\cite{kerbl3Dgaussians}) & \cellcolor{color1}0.8680 & \cellcolor{color2}0.7710 & \cellcolor{color2}0.7750 & \cellcolor{color3}0.9050 & 0.9220 & 0.9380 & 0.9140 & \cellcolor{color3}0.8704 \\
COLMAP (Re-run) & \cellcolor{color3}0.8598 & 0.7517 & \cellcolor{color3}0.7669 & \cellcolor{color1}0.9085 & \cellcolor{color3}0.9312 & \cellcolor{color2}0.9423 & \cellcolor{color2}0.9227 & 0.8690 \\
\midrule
NerfAcc @ 5k & \cellcolor{color3}0.8598 & \cellcolor{color1}0.7723 & \cellcolor{color1}0.7927 & \cellcolor{color3}0.9042 & \cellcolor{color2}0.9325 & \cellcolor{color1}0.9436 & \cellcolor{color1}0.9253 & \cellcolor{color1}0.8758 \\
\midrule
$50^3$ Random + Depth Loss & 0.8433 & 0.7010 & 0.7348 & 0.8863 & 0.9262 & 0.9385 & 0.9075 & 0.8482 \\
NerfAcc @ 5k + Depth Loss & \cellcolor{color2}0.8622 & \cellcolor{color3}0.7656 & \cellcolor{color1}0.7927 & 0.9026 & \cellcolor{color1}0.9334 & \cellcolor{color3}0.9415 & \cellcolor{color3}0.9220 & \cellcolor{color2}0.8743 \\
    \bottomrule
    \end{tabular}
    }
    \label{tab:ssim_360}
\end{table*}

\begin{table*}
    \caption{%
    \textbf{SSIM on the OMMO dataset~\cite{lu2023large}.}
    We report SSIM metric on the OMMO dataset containing large-scale scenes.
    }%
    \centering
    \resizebox{\linewidth}{!}{
    \setlength{\tabcolsep}{4pt}
    \begin{tabular}{@{}lcccccccR} \toprule
& \multicolumn{8}{c}{Scene} \\
Initialization & 03 & 05 & 06 & 10 & 13 & 14 & 15 & Average \\ 
\midrule
COLMAP & 0.8642 & \cellcolor{color2}0.8635 & 0.9018 & \cellcolor{color2}0.8845 & \cellcolor{color1}0.9446 & \cellcolor{color2}0.9393 & \cellcolor{color3}0.9304 & \cellcolor{color2}0.9040 \\
\midrule
NerfAcc @ 5k & \cellcolor{color3}0.8657 & \cellcolor{color3}0.8627 & \cellcolor{color3}0.9083 & 0.8646 & 0.9262 & \cellcolor{color3}0.9381 & \cellcolor{color2}0.9320 & \cellcolor{color3}0.8997 \\
\midrule
$50^3$ Random + Depth Loss & \cellcolor{color2}0.8755 & 0.8615 & \cellcolor{color2}0.9137 & \cellcolor{color1}0.8858 & \cellcolor{color3}0.9292 & 0.9358 & 0.9262 & \cellcolor{color2}0.9040 \\
NerfAcc @ 5k + Depth Loss & \cellcolor{color1}0.8849 & \cellcolor{color1}0.8642 & \cellcolor{color1}0.9220 & \cellcolor{color3}0.8810 & \cellcolor{color2}0.9344 & \cellcolor{color1}0.9408 & \cellcolor{color1}0.9362 & \cellcolor{color1}0.9091 \\
    \bottomrule
    \end{tabular}
    }
    \label{tab:ssim_ommo}
\end{table*}

\begin{table*}
    \caption{%
    \textbf{LPIPS on the Mip-NeRF 360 dataset~\cite{barron2022mip}.}
    We report LPIPS metric on the Mip-NeRF 360 scenes reported in Gaussian Splatting.
    }%
    \centering
    \resizebox{\linewidth}{!}{
    \setlength{\tabcolsep}{4pt}
    \begin{tabular}{@{}lcccccccR} \toprule
& \multicolumn{8}{c}{Scene} \\
Initialization & Garden & Bicycle & Stump & Counter & Kitchen & Bonsai & Room & Average \\ 
\midrule
COLMAP (From~\cite{kerbl3Dgaussians}) & 0.1030 & 0.2050 & 0.2100 & 0.2040 & 0.1290 & 0.2050 & 0.2200 & 0.1823 \\
COLMAP (Re-run) & \cellcolor{color1}0.0865 & \cellcolor{color3}0.1957 & \cellcolor{color3}0.1718 & \cellcolor{color1}0.1439 & \cellcolor{color3}0.0878 & \cellcolor{color3}0.1175 & \cellcolor{color2}0.1533 & \cellcolor{color3}0.1366 \\
\midrule
NerfAcc @ 5k & \cellcolor{color3}0.0869 & \cellcolor{color1}0.1722 & \cellcolor{color1}0.1466 & \cellcolor{color3}0.1458 & \cellcolor{color2}0.0845 & \cellcolor{color2}0.1131 & \cellcolor{color1}0.1473 & \cellcolor{color1}0.1281 \\
\midrule
$50^3$ Random + Depth Loss & 0.1057 & 0.2087 & \cellcolor{color3}0.1796 & 0.1764 & 0.0955 & 0.1203 & 0.1852 & 0.1531 \\
NerfAcc @ 5k + Depth Loss & \cellcolor{color2}0.0868 & \cellcolor{color2}0.1809 & \cellcolor{color1}0.1466 & \cellcolor{color2}0.1446 & \cellcolor{color1}0.0839 & \cellcolor{color1}0.1110 & \cellcolor{color3}0.1536 & \cellcolor{color2}0.1296 \\
    \bottomrule
    \end{tabular}
    }
    \label{tab:lpips_360}
\end{table*}

\begin{table*}
    \caption{%
    \textbf{LPIPS on the OMMO dataset~\cite{lu2023large}.}
    We report LPIPS metric on the OMMO dataset containing large-scale scenes.
    }%
    \centering
    \resizebox{\linewidth}{!}{
    \setlength{\tabcolsep}{4pt}
    \begin{tabular}{@{}lcccccccR} \toprule
& \multicolumn{8}{c}{Scene} \\
Initialization & 03 & 05 & 06 & 10 & 13 & 14 & 15 & Average \\ 
\midrule
COLMAP & 0.1681 & 0.1930 & 0.1586 & \cellcolor{color3}0.1678 & \cellcolor{color1}0.0982 & \cellcolor{color2}0.0755 & \cellcolor{color3}0.0807 & \cellcolor{color3}0.1346 \\
\midrule
NerfAcc @ 5k & \cellcolor{color3}0.1667 & \cellcolor{color3}0.1682 & \cellcolor{color3}0.1535 & 0.2015 & 0.1331 & \cellcolor{color3}0.0783 & \cellcolor{color2}0.0784 & 0.1399 \\
\midrule
$50^3$ Random + Depth Loss & \cellcolor{color2}0.1530 & \cellcolor{color2}0.1528 & \cellcolor{color2}0.1348 & \cellcolor{color1}0.1465 & \cellcolor{color3}0.1324 & 0.0830 & 0.0873 & \cellcolor{color2}0.1271 \\
NerfAcc @ 5k + Depth Loss & \cellcolor{color1}0.1390 & \cellcolor{color1}0.1368 & \cellcolor{color1}0.1181 & \cellcolor{color2}0.1601 & \cellcolor{color2}0.1207 & \cellcolor{color1}0.0742 & \cellcolor{color1}0.0709 & \cellcolor{color1}0.1171 \\
    \bottomrule
    \end{tabular}
    }
    \label{tab:lpips_ommo}
\end{table*}

\subsection{Broader impact}
\label{sec:impact}

We do not propose any new methods or technologies that would pose any novel risks to society which require mitigation.
However, our analysis focuses on the lowest-cost ways to train novel view synthesis models, and the results of this analysis could be relevant in determining the risks posed by such technologies in cases where low barrier to entry could increase the likelihood of misuse as part of synthetic or deceptive media.

\subsection{Dataset licenses}
\label{sec:licenses}

We use the following datasets:
\begin{itemize}
    \item Mip-NeRF 360~\cite{barron2022mip}: no license terms provided. Available at \url{https://jonbarron.info/mipnerf360/}.
    \item OMMO~\cite{lu2023large}: no license terms provided. Available at \url{https://ommo.luchongshan.com/}.
    \item Deep Blending~\cite{hedman2018deep}: no license terms provided. Available at \url{http://visual.cs.ucl.ac.uk/pubs/deepblending/}.
    \item TUM RGB-D~\cite{sturm12iros}: made available under the CC BY 4.0 license \url{https://creativecommons.org/licenses/by/4.0/}. Fetched using script provided by code release of MonoGS~\cite{monogs}.
    \item Replica~\cite{replica19arxiv}: made available under the following terms: \url{https://github.com/facebookresearch/Replica-Dataset/blob/main/LICENSE}. Fetched using script provided by code release of MonoGS~\cite{monogs}.
\end{itemize}

\clearpage

\end{document}